\PassOptionsToPackage{unicode}{hyperref}
\PassOptionsToPackage{hyphens}{url}
\PassOptionsToPackage{dvipsnames,svgnames,x11names}{xcolor}

\documentclass[
  12pt]{article}

\usepackage{enumitem}
\usepackage{amsmath,amssymb}
\usepackage{iftex}
\ifPDFTeX
  \usepackage[T1]{fontenc}
  \usepackage[utf8]{inputenc}
  \usepackage{textcomp}
\else
  \usepackage{unicode-math}
  \defaultfontfeatures{Scale=MatchLowercase}
  \defaultfontfeatures[\rmfamily]{Ligatures=TeX,Scale=1}
\fi
\usepackage{lmodern}
\IfFileExists{upquote.sty}{\usepackage{upquote}}{}
\IfFileExists{microtype.sty}{%
  \usepackage[]{microtype}
  \UseMicrotypeSet[protrusion]{basicmath}
}{}
\makeatletter
\@ifundefined{KOMAClassName}{%
  \IfFileExists{parskip.sty}{%
    \usepackage{parskip}
  }{%
    \setlength{\parindent}{0pt}
    \setlength{\parskip}{6pt plus 2pt minus 1pt}}
}{%
  \KOMAoptions{parskip=half}}
\makeatother
\usepackage{xcolor}
\makeatletter
\ifx\paragraph\undefined\else
  \let\oldparagraph\paragraph
  \renewcommand{\paragraph}{
    \@ifstar
      \xxxParagraphStar
      \xxxParagraphNoStar
  }
  \newcommand{\xxxParagraphStar}[1]{\oldparagraph*{#1}\mbox{}}
  \newcommand{\xxxParagraphNoStar}[1]{\oldparagraph{#1}\mbox{}}
\fi
\ifx\subparagraph\undefined\else
  \let\oldsubparagraph\subparagraph
  \renewcommand{\subparagraph}{
    \@ifstar
      \xxxSubParagraphStar
      \xxxSubParagraphNoStar
  }
  \newcommand{\xxxSubParagraphStar}[1]{\oldsubparagraph*{#1}\mbox{}}
  \newcommand{\xxxSubParagraphNoStar}[1]{\oldsubparagraph{#1}\mbox{}}
\fi
\makeatother

\usepackage{longtable,booktabs,array}
\usepackage{calc}
\usepackage{etoolbox}
\makeatletter
\patchcmd\longtable{\par}{\if@noskipsec\mbox{}\fi\par}{}{}
\makeatother
\IfFileExists{footnotehyper.sty}{\usepackage{footnotehyper}}{\usepackage{footnote}}
\makesavenoteenv{longtable}
\usepackage{graphicx}
\makeatletter
\def\maxwidth{\ifdim\Gin@nat@width>\linewidth\linewidth\else\Gin@nat@width\fi}
\def\maxheight{\ifdim\Gin@nat@height>\textheight\textheight\else\Gin@nat@height\fi}
\makeatother
\setkeys{Gin}{width=\maxwidth,height=\maxheight,keepaspectratio}
\makeatletter
\def\fps@figure{htbp}
\makeatother

\makeatletter
\@ifpackageloaded{caption}{}{\usepackage{caption}}
\AtBeginDocument{%
\ifdefined\contentsname
  \renewcommand*\contentsname{Table of contents}
\else
  \newcommand\contentsname{Table of contents}
\fi
\ifdefined\listfigurename
  \renewcommand*\listfigurename{List of Figures}
\else
  \newcommand\listfigurename{List of Figures}
\fi
\ifdefined\listtablename
  \renewcommand*\listtablename{List of Tables}
\else
  \newcommand\listtablename{List of Tables}
\fi
\ifdefined\figurename
  \renewcommand*\figurename{Figure}
\else
  \newcommand\figurename{Figure}
\fi
\ifdefined\tablename
  \renewcommand*\tablename{Table}
\else
  \newcommand\tablename{Table}
\fi
}
\@ifpackageloaded{float}{}{\usepackage{float}}
\floatstyle{ruled}
\@ifundefined{c@chapter}{\newfloat{codelisting}{h}{lop}}{\newfloat{codelisting}{h}{lop}[chapter]}
\floatname{codelisting}{Listing}

\makeatother
\makeatletter
\@ifpackageloaded{caption}{}{\usepackage{caption}}
\@ifpackageloaded{subcaption}{}{\usepackage{subcaption}}
\makeatother

\ifLuaTeX
  \usepackage{selnolig}
\fi
\usepackage[]{natbib}
\usepackage{bookmark}

\newcommand{\citationaudit}{1}

\IfFileExists{xurl.sty}{\usepackage{xurl}}{}
\hypersetup{
  pdftitle={Title},
  pdfauthor={Author 1; Author 2},
  pdfkeywords={3 to 6 keywords, that do not appear in the title},
  colorlinks=true,
  linkcolor={blue},
  filecolor={Maroon},
  citecolor={Blue},
  urlcolor={Blue},
  pdfcreator={LaTeX via pandoc}}

\newcommand{\anon}{1}

\usepackage{algorithm}
\usepackage{algorithmic}
\usepackage{amsthm}

\newtheorem{theorem}{Theorem}
\newtheorem{lemma}[theorem]{Lemma}
\newtheorem{proposition}[theorem]{Proposition}
\newtheorem{corollary}[theorem]{Corollary}
\theoremstyle{definition}
\newtheorem{definition}[theorem]{Definition}
\newtheorem{example}[theorem]{Example}
\theoremstyle{remark}
\newtheorem{remark}[theorem]{Remark}

\usepackage{comment}

\begin{document}

\def\spacingset#1{\renewcommand{\baselinestretch}%
{#1}\small\normalsize} \spacingset{1}

\newcommand{\yi}[1]{\textbf{\textcolor{purple}{[Yi: #1]}}}

\newcommand{\xiangyu}[1]{\textbf{\textcolor{red}{[Xiangyu: #1]}}}
\newcommand{\changcheng}[1]{\textbf{\textcolor{red}{[Changcheng: #1]}}}

\if1\anon
{
  \title{\bf Surrogate Losses for Fairness-Aware Classification: Fisher Consistency and Calibration}
  \author{Changcheng Song$^1$, Yi Yang$^2$, Xiangyu Chang$^3$\\
   Xiaoling Lu$^1$, Yufeng Liu$^4$\thanks{
    $^1$Renmin University of China, $^2$Arizona State University, $^3$Xi'an Jiaotong University, $^4$ University of Michigan.
    }\hspace{.2cm}\\
    }
  \maketitle
} \fi

\if0\anon
{
  \bigskip
  \bigskip
  \bigskip
  \begin{center}
    {\LARGE\bf Title}
\end{center}
  \medskip
} \fi

\bigskip
\begin{abstract}
Fairness-aware classification seeks to balance predictive accuracy and fairness across demographic groups.
Although surrogate losses facilitate the optimization of fairness-aware objectives, it remains unclear when minimizing these losses yields optimal decisions under the original fairness-aware risk.
This paper investigates the statistical validity of surrogate-based fairness-aware classification under demographic parity and equal opportunity, considering both attribute-blind and attribute-aware settings.
We introduce fairness-aware Fisher consistency and characterize it through conditions on the derivatives of the surrogate losses.
The characterization shows that consistency depends not only on the individual loss functions but also on their interaction within the fairness-regularized objective.
We further establish a calibration inequality that relates the excess fairness-aware risk to the excess surrogate risk, providing a quantitative connection between surrogate optimization and the original classification objective.
Numerical experiments illustrate how deviations from the consistency conditions affect fairness-aware excess risk and how the regularization parameter influences the well-posedness of surrogate optimization.
An application to consumer credit default prediction demonstrates that appropriately calibrated fairness regularization can substantially reduce demographic disparities while maintaining predictive performance.
\end{abstract}

\vfill

\newpage
\spacingset{1.8}

\section{Introduction}\label{sec:intro} 

Statistical classification models have become increasingly prevalent in high-stakes decision-making applications, including hiring screening \citep{raghavan-2020-mitigating}, criminal risk assessment \citep{angwin-2022-machine}, and healthcare resource allocation \citep{obermeyer-2019-dissecting}. 
While these models often achieve strong predictive performance, a growing body of empirical evidence \citep{Caton-2024-FairnessMachineLearning} has revealed that they may systematically disadvantage certain demographic groups defined by \textit{sensitive attributes} such as gender, race, or ethnicity. Such disparities can lead to unequal treatment across groups and may perpetuate or even amplify existing social inequalities when algorithmic decisions influence access to employment opportunities, judicial outcomes, or healthcare services. 
Consequently, developing classification models that balance predictive accuracy and fairness has become an important objective in the modern statistical learning community.

To address fairness concerns, prior work has proposed a range of group fairness notions for classification that formalize equity requirements across groups. 
Prominent examples include \emph{Demographic Parity (DP)} \citep{Dwork-2012-DP}, which requires equal selection rates across groups and \emph{Equal Opportunity (EO)} \citep{Hardt-2016-EqualityOfOpportunity}, which mandates equal true positive rates across groups.
These definitions provide measures of algorithmic fairness and are widely used to regulate classifier behavior and mitigate discriminatory outcomes.

Based on these fairness notions, researchers have developed a variety of fairness-aware classification approaches by incorporating fairness violations directly into the learning objective.
A common approach \citep{Menon-2018-CostFairnessBinary,Bechavod-2018-PenalizingUnfairnessBinary} is to formulate fairness-aware classification as a fairness-regularized risk minimization problem (See Eq.~\eqref{eq:fair-risk-01}), balancing predictive accuracy against fairness violations through a trade-off parameter.
Because both the classification error and fairness violations are typically defined through indicator functions over discrete events (See Eq.~\eqref{eq:cs-loss}), the corresponding optimization problem is generally non-convex and discontinuous.

From an optimization perspective, existing approaches to solve the fairness-regularized risk minimization problem fall into two methodological streams.
The first stream attempts to directly optimize the original fairness-aware problem based on the 0--1 loss and exact fairness violations, resulting in a discrete optimization problem. 
To handle the discrete optimization, mixed-integer programming approaches \citep{Ustun-2016-SupersparseLinearInteger, Yang-2025-TowardFairnessAware} are adopted for fairness-aware classification. 
Although the first methodological stream targets the original optimization problem, it typically suffers from substantial computational complexity and limited scalability on large-scale datasets \citep{Donini-2018-FairnessERM}.

The second stream takes a different approach by replacing the discontinuous components with convex and smooth surrogate functions.
In this paradigm, the 0--1 loss is substituted with surrogate losses such as the logistic, hinge, or exponential loss, while fairness violations are also approximated using surrogate functions \citep{Bechavod-2018-PenalizingUnfairnessBinary}. 
To fix notation, let $L$ and $\bar{L}$ denote the surrogate losses for the accuracy and fairness terms, with risks $\mathcal{R}_{L}(f)$ and $\mathcal{R}_{\bar{L}}(f)$. Surrogate-based methods then minimize
\begin{equation*}
\mathcal{R}_{\mathrm{fair}}^{L,\bar{L}}(f):=\mathcal{R}_{L}(f)-\lambda\cdot \mathcal{R}_{\bar{L}}(f),
\end{equation*}
where $\lambda\geq 0$ is the trade-off parameter. 
We formalize these definitions in Section~\ref{subsec:fair-surrogate} and denote $\mathcal{R}_{L}(f), \mathcal{R}_{\bar{L}}(f)$ and $\mathcal{R}_{\mathrm{fair}}^{L,\bar{L}}(f)$ as $L$-risk, $\bar{L}$-risk and $(L,\bar{L})$-risk with respect to $f$ respectively. 
This optimization problem can be efficiently optimized using standard gradient-based algorithms.
As a result, surrogate-based methods are computationally scalable and have become the dominant approach for fairness-aware classification \citep{Zafar-2019-FairnessConstraintsFlexible,He-2025-FairnessMachineLearning}.

The widespread adoption of surrogate-based methods naturally raises a fundamental question: \textit{when does minimizing a surrogate objective lead to a statistically valid solution for the original learning problem?}
In standard classification, this question is addressed through two foundational theoretical concepts. 
The first is \emph{Fisher consistency}, which examines whether minimizing the surrogate $L$-risk recovers the Bayes-optimal classifier for the underlying 0--1 loss \citep{Lin-2004-FisherConsistency, Yang-2021-PrivacyCostSensitive, Yang-2021-AngleCostSensitive}. 
The second is \emph{classification calibration}, which establishes quantitative relationships between the excess $L$-risk and the excess risk through \emph{calibrated inequalities} \citep{Bartlett-2006-ConvexityClassificationRisk}. 
However, much less is known about whether these guarantees continue to hold in fairness-aware classification, where recent finite-sample and distribution-free results focus on the fairness-accuracy trade-off \citep{Li-2023-FaiREE,Hou-2024-FiniteSampleFair} rather than on the consistency and calibration of surrogate fairness-aware risks.
Unlike standard classification, fairness-aware objectives simultaneously balance predictive accuracy and fairness violation through a fairness-regularized risk formulation.
This raises two fundamental questions:
\begin{enumerate}[label=\textbf{Q\arabic*.}]
\item \textbf{(Fairness-Aware Fisher Consistency):}
Under what conditions does the population minimizer of a $(L,\bar{L})$-risk induce the Bayes-optimal  classifier for the corresponding fairness-aware risk?

\item \textbf{(Fairness-Aware Calibration):}
Can we establish a calibration inequality that relates the excess $(L,\bar{L})$-risk to the excess fairness-aware risk?
\end{enumerate}

This paper addresses these questions by developing a statistical learning theory for surrogate-based fairness-aware classification.
To answer the first question, we introduce the notion of \emph{Fairness-Aware Fisher Consistency (FAFC)}.
We begin with a pointwise analysis of the fairness-aware conditional risk and identify a fairness-aware discriminant that determines the Bayes-optimal classifier for the original fairness-aware risk.
We then define FAFC for the fairness-aware surrogate loss pointwise. At every pair of conditional probabilities at which the fairness-aware discriminant does not vanish, the smallest conditional risk achievable by a wrong-sign prediction is strictly larger than the optimal conditional risk. 
We then establish necessary and sufficient conditions under which $L_{\mathrm{fair}}$ satisfies FAFC.
To answer the second question, we construct a fairness-aware calibration function by minimizing the pointwise wrong-sign excess $(L,\bar{L})$-risk over level sets of the fairness-aware discriminant.
By taking its convex envelope, we obtain a calibrated inequality that controls the excess fairness-aware risk in terms of the excess $(L,\bar{L})$-risk.
Consequently, under the proposed FAFC conditions, convergence of the excess $(L,\bar{L})$-risk to zero implies convergence of the excess fairness-aware risk to zero, thereby providing a formal justification for surrogate-based fairness-aware classification.

Our main contributions are summarized as follows:
\begin{itemize}
\item \textbf{A pointwise framework and FAFC characterization.}
We derive the fairness-aware conditional risk for DP and EO in both the attribute-blind and attribute-aware settings, and identify a fairness-aware discriminant that determines the Bayes-optimal classifier.
Based on it, we introduce FAFC for surrogate loss pairs $(L,\bar{L})$, and give necessary and sufficient conditions for FAFC in terms of the derivatives of the partial losses at the origin, which can be checked directly from the loss pair.

\item \textbf{A calibration inequality for fairness-aware risks.}
We construct a fairness-aware calibration function by minimizing the pointwise wrong-sign excess $(L,\bar{L})$-risk over level sets of the discriminant, and show that its convex envelope yields an inequality bounding the excess fairness-aware risk by the excess $(L,\bar{L})$-risk.
We further show that this envelope is invertible if and only if the pair is FAFC, so that FAFC is exactly the condition under which surrogate risk convergence implies convergence of the fairness-aware risk.

\item \textbf{Closed-form guarantees for widely used surrogates.}
We verify FAFC and derive explicit calibration functions and excess-risk bounds for fairness-aware logistic, hinge, and exponential losses, covering DP and EO in both feature settings, and we propose a general surrogate loss-based algorithm that can be used with any FAFC-compliant pair.
Simulations and a credit-default application illustrate the accuracy--fairness trade-off.
\end{itemize}

\section{Preliminaries}

We begin by introducing the notation and basic theory of cost-sensitive classification, since the fairness-aware objectives studied in this paper can be naturally represented and analyzed within the cost-sensitive learning framework~\citep{Menon-2018-CostFairnessBinary}.

\subsection{Cost-Sensitive Classification and Surrogate Loss}\label{subsec:prelims-cost-sensitive}

Let $(X,Y)$ be a random pair drawn from a distribution $D$ on $\mathcal{X}\times\{-1,1\}$, where $\mathcal{X}$ is the feature space and $Y\in\{-1,1\}$ is the ground-truth label. 
Let $\mathcal{F}$ denote the set of all measurable functions $f:\mathcal{X}\to\mathbb{R}$; each $f$ induces a classifier through the decision rule $\hat{Y}=\mathrm{sign}(f(X))$, with the convention $\mathrm{sign}(0)=-1$.

\begin{definition}[Cost-Sensitive Risk and Bayes-Optimal  Classifier]\label{def:cs-risk}
For a classifier $f\in\mathcal{F}$ and a distribution $D$, the \emph{cost-sensitive risk} with cost parameter $\alpha\in(0,1)$ is defined as
\begin{equation}\label{eq:cs-risk}
\mathcal{R}_{\alpha}(f;D): = (1-\alpha)\cdot\mathbb{P}(\hat{Y}=-1,Y=1) + \alpha\cdot\mathbb{P}(\hat{Y}=1,Y=-1),
\end{equation}
where $\hat{Y}=\mathrm{sign}(f(X))$ is the prediction label induced by $f$.
Then Bayes-optimal classifier is:
\begin{equation}\label{eq:bayes-optimal}
    \gamma(x)\in\arg\min_{f\in \mathcal{F}}\mathcal{R}_{\alpha}(f;D),
\end{equation}
where $\gamma(x)=\eta(x)-\alpha$ and $\eta(x):=\mathbb{P}(Y=1\mid X=x)$ is the posterior probability of the positive class. 
The cost parameter $\alpha$ controls the relative penalty on false negatives versus false positives; when $\alpha=0.5$, $\mathcal{R}_{\alpha}(f;D)$ reduces to the misclassification error rate $\mathbb{P}(\hat{Y}\neq Y)$.
\end{definition}
Equivalently, the risk can be written as the expectation of a pointwise cost-sensitive 0--1 loss \citep{Scott-2012-CalibratedAsymmetricSurrogate}.
Let us define
\begin{equation}\label{eq:cs-loss}
U_{\alpha}(y,t): = (1-\alpha)\cdot \mathbf{1}\{y=1\}\mathbf{1}\{t\le 0\} + \alpha\cdot \mathbf{1}\{y=-1\}\mathbf{1}\{t>0\},
\end{equation}
then $\mathcal{R}_{\alpha}(f;D)=\mathbb{E}_{(X,Y)\sim D}\left[U_{\alpha}(Y,f(X))\right]$. Here $\mathbf{1}\{\cdot\}$ denotes the indicator function.

The 0--1 loss $U_\alpha$ is discontinuous and non-convex, so direct minimization of $\mathcal{R}_{\alpha}(\cdot;D)$ from finite data is computationally intractable. 
The standard remedy in statistical learning is to replace $U_\alpha$ by a smooth, convex \emph{surrogate loss} $L:\{-1,1\}\times\mathbb{R}\to[0,\infty)$. 
Any surrogate loss admits the partial-loss decomposition
\begin{equation}\label{eq:surrogate-loss}
L(y,t):= \mathbf{1}\{y=1\}L_{1}(t) + \mathbf{1}\{y=-1\}L_{-1}(t),
\end{equation}
where $L_1(t):=L(1,t)$ and $L_{-1}(t):=L(-1,t)$ denote the partial losses for the positive and negative labels, respectively. 
Replacing $U_\alpha$ by $L$ in Eq.~\eqref{eq:cs-risk} yields the \emph{$L$-risk}
\begin{equation*}
\mathcal{R}_L(f;D): = \mathbb{E}_{(X,Y)\sim D}\left[L(Y,f(X))\right].
\end{equation*}
We denote the minimal risk by $\mathcal{R}_{\alpha}^*:=\inf_{f\in\mathcal{F}}\mathcal{R}_{\alpha}(f;D)$ and the minimal $L$-risk by $\mathcal{R}_L^*:=\inf_{f\in\mathcal{F}}\mathcal{R}_L(f;D)$. 

To validate that minimizing the $L$-risk yields a statistically valid solution for the original optimization problem, we require two theoretical guarantees: 
(1) a qualitative asymptotic guarantee that the surrogate recovers the Bayes-optimal  classifier, 
and (2) a quantitative bound that links the excess $L$-risk to the excess risk. 
We formalize these two properties in the remainder of this subsection.

\subsubsection{Fisher Consistency}

Fisher consistency is the fundamental asymptotic guarantee for surrogate optimization: the population minimizer of the $L$-risk must recover the Bayes-optimal classifier of the original risk \citep{Elkan-2001-FoundationsCostSensitiveLearning, Lin-2004-FisherConsistency}.

\begin{definition}[Fisher Consistency for Cost-Sensitive Classification]\label{def:fc-cs}
A surrogate loss $L$ is \emph{$\alpha$-Fisher consistent} for cost-sensitive classification with cost $\alpha\in(0,1)$ if, for every distribution $D$ and every minimizer $f^{*}\in\arg\min_{f\in\mathcal{F}}\mathcal{R}_L(f;D)$, its sign matches that of the cost-sensitive Bayes-optimal classifier:
\begin{equation*}
\mathrm{sign}(f^{*}(x)) = \mathrm{sign}(\gamma(x)) \quad \text{for } P_X\text{-a.e. } x\in\mathcal{X} \text{ with } \gamma(x)\neq0,
\end{equation*}
where $\gamma(x)=\eta(x)-\alpha$ is the cost-sensitive discriminant in Eq.~\eqref{eq:bayes-optimal}.
\end{definition}

Satisfying Fisher consistency guarantees that, given the population distribution, minimizing the surrogate loss yields the same classification decisions as the Bayes-optimal classifier almost surely.
However, this guarantee is stated at the level of population minimizers and is not directly checkable from the loss. The following result of \citet{Scott-2012-CalibratedAsymmetricSurrogate} shows that, for convex partial losses differentiable at $0$, Fisher consistency at cost $\alpha$ reduces to a local condition on the derivatives of $L_1$ and $L_{-1}$ at the origin, which can be read off from the loss alone.

\begin{lemma}\label{thm:cs-cc-prelim}
Let $L$ be a surrogate loss with convex partial losses $L_1,L_{-1}$ differentiable at $0$, and let $\alpha\in(0,1)$. Then $L$ is Fisher consistent at cost $\alpha$ if and only if
\begin{equation*}
L_{1}'(0)<0,\qquad L_{-1}'(0)>0,\qquad \alpha L_{1}'(0)+(1-\alpha)L_{-1}'(0)=0.
\end{equation*}
\end{lemma}

\subsubsection{Classification Calibration}

Beyond the asymptotic guarantee of Fisher consistency, classification calibration provides a quantitative bound relating the excess $L$-risk to the excess risk through a calibration function \citep{Bartlett-2006-ConvexityClassificationRisk,Scott-2012-CalibratedAsymmetricSurrogate}. 

\begin{lemma}[Classification Calibration for Cost-Sensitive Classification]\label{thm:cs-calibration-prelim}
Let $L$ be a non-negative surrogate loss and $\alpha\in(0,1)$.
\begin{enumerate}
    \item There exists a convex, non-decreasing calibration function $\psi_{L,\alpha}:[0,\infty)\to[0,\infty)$ with $\psi_{L,\alpha}(0)=0$ such that, for every measurable classifier $f\in\mathcal{F}$ and every distribution $D$,
    \begin{equation*}
    \psi_{L,\alpha}\left(\mathcal{R}_{\alpha}(f;D)-\mathcal{R}_{\alpha}^{*}\right) \le \mathcal{R}_{L}(f;D)-\mathcal{R}_{L}^{*}.
    \end{equation*}
    \item The calibration function $\psi_{L,\alpha}$ is invertible if and only if $L$ is Fisher consistent in the sense of Definition~\ref{def:fc-cs}.
\end{enumerate}
\end{lemma}

\subsection{Fairness-Aware Classification}\label{subsec:prelims-fairness-problem}

Fairness-aware classification extends the cost-sensitive setting of Section~\ref{subsec:prelims-cost-sensitive} by adding a binary sensitive attribute $A\in\{-1,1\}$ (e.g., gender or race) alongside the target label $Y$.
We consider both the attribute-aware setting, in which $A$ is available at training and prediction time, and the attribute-blind setting, in which $A$ is available at training time only \citep{Menon-2018-CostFairnessBinary}.
In either setting, the goal is to learn a classifier $f\in\mathcal{F}$ that balances accuracy with respect to $Y$ against group fairness with respect to $A$.

Group fairness is formalized as a statistical parity condition on the predicted label $\hat{Y}=\mathrm{sign}(f(X))$ across the two groups.
We focus on two of the most widely adopted criteria, \emph{Demographic Parity (DP)} and \emph{Equal Opportunity (EO)}.

\begin{definition}[DP \citep{Dwork-2012-DP}]\label{def:dp}
A classifier satisfies DP if its prediction $\hat{Y}$ is statistically independent of the sensitive attribute $A$:
\begin{equation*}
\mathbb{P}(\hat{Y}=1 \mid A=1) = \mathbb{P}(\hat{Y}=1 \mid A=-1).
\end{equation*}
This criterion demands identical positive prediction rates across the two groups, regardless of their ground-truth label distributions.
\end{definition}

\begin{definition}[EO \citep{Hardt-2016-EqualityOfOpportunity}]\label{def:eo}
A classifier satisfies EO if it achieves equal true positive rates across all groups:
\begin{equation*}
\mathbb{P}(\hat{Y}=1 \mid Y=1, A=1) = \mathbb{P}(\hat{Y}=1 \mid Y=1, A=-1).
\end{equation*}
EO relaxes the independence requirement of DP to conditional independence between $\hat{Y}$ and $A$ given $Y=1$.
\end{definition}

Exact parity in Definitions~\ref{def:dp} and~\ref{def:eo} is generally incompatible with accuracy and cannot be guaranteed on finite data \citep{Shen2023}.
In practice, fairness requirements are therefore relaxed to \emph{approximate fairness} and quantified by discrimination metrics \citep{Hardt-2016-EqualityOfOpportunity,kleinberg-2017-InherentTradeOffs}.
We use the two standard metrics, \emph{Mean Difference (MD)} and \emph{Mean Ratio (MR)}.

\begin{definition}[MD \citep{Calders-2010-ThreeNaiveBayes}]\label{def:md}
For a classifier $f$,
\begin{align*}
\mathrm{MD}_{\mathrm{DP}}(f) &= \mathbb{P}(\hat{Y}=1 \mid A=-1) - \mathbb{P}(\hat{Y}=1 \mid A=1),\\
\mathrm{MD}_{\mathrm{EO}}(f) &= \mathbb{P}(\hat{Y}=1 \mid Y=1, A=-1) - \mathbb{P}(\hat{Y}=1 \mid Y=1, A=1).
\end{align*}
\end{definition}

\begin{definition}[MR \citep{Feldman-2015-CertifyingRemovingDisparate}]\label{def:mr}
For a classifier $f$,
\begin{align*}
\mathrm{MR}_{\mathrm{DP}}(f) &= \frac{\mathbb{P}(\hat{Y}=1 \mid A=-1)}{\mathbb{P}(\hat{Y}=1 \mid A=1)},\\
\mathrm{MR}_{\mathrm{EO}}(f) &= \frac{\mathbb{P}(\hat{Y}=1 \mid Y=1, A=-1)}{\mathbb{P}(\hat{Y}=1 \mid Y=1, A=1)}.
\end{align*}
\end{definition}

The key observation of \citet{Menon-2018-CostFairnessBinary} is that approximate fairness measures can be written as a regularization term. 
Let $\bar{D}_{\mathrm{DP}}$ be the joint law of $(X,A)$ and $\bar{D}_{\mathrm{EO}}$ the conditional law of $(X,A)$ given $Y=1$.
For $\bar{\alpha}\in(0,1)$, define 
\begin{equation}\label{eq:fair-aux-risk}
\mathcal{R}_{\bar{\alpha}}(f;\bar{D})
:= (1-\bar{\alpha})\,\mathbb{P}(\hat{Y}=-1,A=1)
+ \bar{\alpha}\,\mathbb{P}(\hat{Y}=1,A=-1),
\end{equation}
where $\bar{D}$ is either $\bar{D}_{\mathrm{DP}}$ or $\bar{D}_{\mathrm{EO}}$.

The choice of $\bar{\alpha}$ in Eq.~\eqref{eq:fair-aux-risk} is dictated by whether MD or MR is used; Appendix~\ref{app:fair-penalized} records the four standard cases.
Motivated by this equivalence, we treat $\lambda$ as a trade-off parameter $\lambda\ge0$ chosen by the decision maker and study the \emph{fairness-regularized risk}
\begin{align}\label{eq:fair-risk-01}
\mathcal{R}_{\mathrm{fair}}^{\alpha,\bar{\alpha}}(f;D,\bar{D})
&:= \mathcal{R}_{\alpha}(f;D)-\lambda\cdot\mathcal{R}_{\bar{\alpha}}(f;\bar{D}) \nonumber\\
&= \mathbb{E}_{(X,Y)\sim D}\left[U_{\alpha}(Y,f(X))\right]
-\lambda\,\mathbb{E}_{(X,A)\sim\bar{D}}\left[U_{\bar{\alpha}}(A,f(X))\right],
\end{align}
where $\mathcal{R}_{\bar{\alpha}}$ is the auxiliary risk in Eq.~\eqref{eq:fair-aux-risk} and $U_{\bar{\alpha}}$ is defined as in Eq.~\eqref{eq:cs-loss} with $\alpha$ replaced by $\bar{\alpha}$.

The regularized form in Eq.~\eqref{eq:fair-risk-01} and the Bayes-optimal classifier below are due to \citet{Menon-2018-CostFairnessBinary}, \citet{Zeng-2026-BayesOptimalFair}, and \citet{Yang-2026-BayesOptimalFairClassification}, who derived them through the equivalence recorded in Lemma~\ref{lem:fair-penalized} (Appendix~\ref{app:fair-penalized}).
We adopt this formulation as our starting point.

To state the classifier, recall that $\eta(x)=\mathbb{P}(Y=1\mid X=x)$; let $\bar{\eta}_{\mathrm{DP}}(x):=\mathbb{P}(A=1\mid X=x)$, $\bar{\eta}_{\mathrm{EO}}(x):=\mathbb{P}(A=1\mid X=x,Y=1)$, $\pi_A:=\mathbb{P}(A=1)$, $\pi_{A\mid Y=1}:=\mathbb{P}(A=1\mid Y=1)$, and $\pi:=\mathbb{P}(Y=1)$.
In the attribute-aware setting, let $\eta(x,a):=\mathbb{P}(Y=1\mid X=x,A=a)$.

\begin{theorem}[Bayes-Optimal Fairness-Aware Classifiers]\label{thm:bayes-fair-01}
For both DP and EO, the Bayes-optimal classifier is $\hat{Y}=\mathrm{sign}(f^*(x))$ in the attribute-blind setting and $\hat{Y}=\mathrm{sign}(f^*(x,a))$ in the attribute-aware setting, where a minimizer $f^*\in\arg\min_{f\in\mathcal{F}}\mathcal{R}_{\mathrm{fair}}^{\alpha,\bar{\alpha}}(f;D,\bar{D})$ is given by the following forms:
\begin{align*}
\text{DP (attribute-blind):}\quad & f^{*}_{\mathrm{DP}}(x) = \eta(x) - \alpha - \lambda \left(\bar{\eta}_{\mathrm{DP}}(x)-\bar{\alpha}\right), \\
\text{DP (attribute-aware):}\quad & f^{*}_{\mathrm{DP}}(x,a) = \eta(x,a) - \alpha - \lambda \left( \frac{a+1}{2}-\bar{\alpha} \right), \\
\text{EO (attribute-blind):}\quad & f^{*}_{\mathrm{EO}}(x) = \eta(x) - \alpha - \lambda \pi^{-1}\eta(x) \left(\bar{\eta}_{\mathrm{EO}}(x)-\bar{\alpha}\right), \\
\text{EO (attribute-aware):}\quad & f^{*}_{\mathrm{EO}}(x,a) = \eta(x,a) - \alpha - \lambda \pi^{-1}\eta(x,a) \left( \frac{a+1}{2}-\bar{\alpha} \right),
\end{align*}
\end{theorem}

Theorem~\ref{thm:bayes-fair-01} shows that the fairness-aware Bayes classifier retains the cost-sensitive discriminant $\eta-\alpha$ of Eq.~\eqref{eq:bayes-optimal} and subtracts an instance-dependent adjustment.
For DP, the adjustment is $\lambda(\bar{\eta}_{\mathrm{DP}}-\bar{\alpha})$, the deviation of the sensitive-attribute posterior from its base rate; in the attribute-aware case the posterior is replaced by $(a+1)/2$.
For EO, the corresponding deviation is weighted by $\pi^{-1}\eta$, with $\eta$ evaluated at $x$ or $(x,a)$ according to the setting, reflecting that the fairness term is evaluated only on the positive class.
Setting $\lambda=0$ recovers the cost-sensitive classifier in Eq.~\eqref{eq:bayes-optimal}.

\subsection{Surrogate Losses for Fairness-Aware Classification}\label{subsec:fair-surrogate}

Although Eq.~\eqref{eq:fair-risk-01} admits the closed-form Bayes-optimal  classifier of Theorem~\ref{thm:bayes-fair-01}, the fairness-regularized risk is built on the discontinuous, non-convex 0--1 loss and is therefore intractable to minimize directly from finite data. 
Following the standard paradigm of Section~\ref{subsec:prelims-cost-sensitive}, practical fairness-aware methods replace the two cost-sensitive 0--1 losses by surrogate losses. 
Let $L:\{-1,1\}\times\mathbb{R}\to[0,\infty)$ be a surrogate loss for the accuracy term (with partial losses $L_1,L_{-1}$ as in Eq.~\eqref{eq:surrogate-loss}), and let $\bar{L}:\{-1,1\}\times\mathbb{R}\to\mathbb{R}$ be a surrogate loss for the fairness term (with partial losses $\bar{L}_1,\bar{L}_{-1}$). We call $(L,\bar{L})$ a \emph{fairness-aware surrogate loss pair}, and define the composite \emph{fairness-aware surrogate loss}
\begin{equation*}
L_{\mathrm{fair}}(y,a,t) := L(y,t) - \lambda\bar{L}(a,t),
\end{equation*}
and the corresponding \emph{$(L,\bar{L})$-risk}
\begin{align*}
    \mathcal{R}_{\mathrm{fair}}^{L,\bar{L}}(f;D,\bar{D}) &:= \mathcal{R}_{L}(f;D) - \lambda\cdot\mathcal{R}_{\bar{L}}(f;\bar{D})=\mathbb{E}[L_{\mathrm{fair}}(Y,A,f(X))].\\
&= \mathbb{E}_{(X,Y)\sim D}\left[L(Y,f(X))\right] - \lambda\mathbb{E}_{(X,A)\sim\bar{D}}\left[\bar{L}(A,f(X))\right]. 
\end{align*}
We denote the minimal values of the two risks by $\mathcal{R}_{\mathrm{fair}}^{\alpha,\bar{\alpha},*}:=\inf_{f\in\mathcal{F}}\mathcal{R}_{\mathrm{fair}}^{\alpha,\bar{\alpha}}(f;D,\bar{D})$ and $\mathcal{R}_{\mathrm{fair}}^{L,\bar{L},*}:=\inf_{f\in\mathcal{F}}\mathcal{R}_{\mathrm{fair}}^{L,\bar{L}}(f;D,\bar{D})$.

When the surrogate losses $L$ and $\bar{L}$ are convex and smooth, the $(L,\bar{L})$-risk can be minimized efficiently using standard gradient-based methods, which is why surrogate-based methods are the dominant approach in fairness-aware classification practice. 
However, the classical guarantees of Section~\ref{subsec:prelims-cost-sensitive} do not transfer directly: 
the fairness-aware objective is a \emph{difference} of two cost-sensitive risks, so neither the Bayes-optimal  classifier nor the calibration function of a single surrogate loss suffices to control the fairness-aware excess risk. 
This raises the central question of this paper: \emph{can we establish Fisher consistency and classification calibration in the fairness-aware classification in Eq.~\eqref{eq:fair-risk-01}?}
We answer this question in the following sections.

\section{Fisher Consistency for Fairness-Aware Classification}\label{sec:fc-fair}

This section adapts the concept of Fisher consistency to surrogate losses in fairness-aware classification defined in Section~\ref{subsec:fair-surrogate}.
Specifically, we begin with a pointwise analysis of the fairness-aware conditional risk, then define FAFC, and finally establish a necessary and sufficient condition for this property.

\subsection{Fairness-Aware Conditional Risk}\label{subsec:fc-pointwise}

To analyze the properties of the $(L,\bar{L})$-risk, we derive fairness-aware conditional risk results for the attribute-blind and attribute-aware settings individually, and unify both cases into a generalized framework. 
For clarity, we develop the attribute-blind setting in detail first and then show that the attribute-aware setting arises as a special case. 

\subsubsection{The Attribute-Blind Setting}

In the attribute-blind setting, the classifier $f$ makes decision solely based on the features $X$. For a fixed instance $x\in \mathcal{X}$, we will analyze the two fairness notions (i.e., DP and EO) discussed in Section \ref{subsec:prelims-fairness-problem}.

For DP, the fairness penalty is evaluated over the marginal distribution of $X$ and $A$. The fairness-aware conditional risk is: 
\begin{align*}
& \mathbb{E}_{(X,Y)\sim D|X=x} \left[ U_{\alpha}(Y, f(X))\right] - \lambda \mathbb{E}_{(X,A)\sim \bar{D}_{\mathrm{DP}}|X=x}\left[ U_{\bar{\alpha}}(A, f(X)) \right] \\
&= \left[ (1-\alpha)\eta(x) - \lambda(1-\bar{\alpha})\bar{\eta}_{\mathrm{DP}}(x) \right] \mathbf{1}\{f(x) \le 0\} \\
& + \left[ \alpha(1-\eta(x)) - \lambda\bar{\alpha}(1-\bar{\eta}_{\mathrm{DP}}(x)) \right] \mathbf{1}\{f(x) > 0\}.
\end{align*}

For EO, the fairness penalty is conditioned on the positive class $Y=1$. 
The fairness-aware conditional risk is: 
\begin{align*}
& \mathbb{E}_{(X,Y)\sim D|X=x} \left[ U_{\alpha}(Y, f(X)) \right] - \lambda \mathbb{E}_{(X,A) \sim \bar{D}_{\mathrm{EO}}|X=x} \left[ U_{\bar{\alpha}}(A, f(X)) \right] \\
&= \left[ (1-\alpha)\eta(x) - \lambda \pi^{-1} \eta(x) (1-\bar{\alpha})\bar{\eta}_{\mathrm{EO}}(x) \right] \mathbf{1}\{f(x) \le 0\} \\
& + \left[ \alpha(1-\eta(x)) - \lambda \pi^{-1} \eta(x) \bar{\alpha}(1-\bar{\eta}_{\mathrm{EO}}(x)) \right] \mathbf{1}\{f(x) > 0\}.
\end{align*}

We unify the above cases by defining a general fairness-aware conditional risk. 
Define that
\begin{equation}\label{eq:attributed-blind fairness aware conditional risk}
C_{\mathrm{fair}}^{\alpha,\bar{\alpha}}(\eta, \bar{\eta}, t) = A^{\alpha,\bar{\alpha}}(\eta,\bar{\eta}) \mathbf{1}\{t \le 0\} + B^{\alpha,\bar{\alpha}}(\eta,\bar{\eta}) \mathbf{1}\{t > 0\},
\end{equation}
where $\bar{\eta}$ denotes $\bar{\eta}_{\mathrm{DP}}$ or $\bar{\eta}_{\mathrm{EO}}$ depending on the context, and $A^{\alpha,\bar{\alpha}}(\eta,\bar{\eta})$ and $B^{\alpha,\bar{\alpha}}(\eta,\bar{\eta})$ are cost functions derived from the specific fairness notion. Since $A^{\alpha,\bar{\alpha}}(\eta,\bar{\eta})$ and $B^{\alpha,\bar{\alpha}}(\eta,\bar{\eta})$ depend on $\eta$ and $\bar{\eta}$, which is associated with every datapoint, we call them \emph{example-dependent fair cost functions.}

For DP, the example-dependent cost functions are 
\begin{align*}
A^{\alpha,\bar{\alpha}}_{\mathrm{DP}}(\eta ,\bar{\eta}_{\mathrm{DP}}) & = \eta(1-\alpha) - \lambda \bar{\eta}_{\mathrm{DP}}(1-\bar{\alpha}), \\
B^{\alpha,\bar{\alpha}}_{\mathrm{DP}}(\eta ,\bar{\eta}_{\mathrm{DP}}) & = (1-\eta)\alpha - \lambda (1-\bar{\eta}_{\mathrm{DP}})\bar{\alpha}
\end{align*}
and the conditional risk is defined as
\begin{equation*}
C^{\alpha,\bar{\alpha}}_{\mathrm{DP}}(\eta,\bar{\eta}_{\mathrm{DP}},t) := A^{\alpha,\bar{\alpha}}_{\mathrm{DP}}(\eta ,\bar{\eta}_{\mathrm{DP}})\mathbf{1}\left\{ t\leq 0 \right\}+B^{\alpha,\bar{\alpha}}_{\mathrm{DP}}(\eta ,\bar{\eta}_{\mathrm{DP}})\mathbf{1}\left\{ t>0 \right\}.
\end{equation*}

For EO, the example-dependent cost functions are
\begin{align*}
A^{\alpha,\bar{\alpha}}_{\mathrm{EO}}(\eta ,\bar{\eta}_{\mathrm{EO}}) & = \eta(1-\alpha) - \lambda \pi ^{-1}\eta \bar{\eta}_{\mathrm{EO}}(1-\bar{\alpha}), \\
B^{\alpha,\bar{\alpha}}_{\mathrm{EO}}(\eta ,\bar{\eta}_{\mathrm{EO}}) & = (1-\eta)\alpha - \lambda \pi ^{-1}\eta (1-\bar{\eta}_{\mathrm{EO}})\bar{\alpha}
\end{align*}
and the conditional risk is defined as
\begin{equation*}
C^{\alpha,\bar{\alpha}}_{\mathrm{EO}}(\eta,\bar{\eta}_{\mathrm{EO}},t) := A^{\alpha,\bar{\alpha}}_{\mathrm{EO}}(\eta ,\bar{\eta}_{\mathrm{EO}})\mathbf{1}\left\{ t\leq 0 \right\}+B^{\alpha,\bar{\alpha}}_{\mathrm{EO}}(\eta ,\bar{\eta}_{\mathrm{EO}})\mathbf{1}\left\{ t>0 \right\}.
\end{equation*}

Both the risk of DP and EO satisfies
\begin{equation*}
    \mathcal{R}_{\mathrm{fair}}^{\alpha,\bar{\alpha}}(f) = \mathbb{E}_{X}\left[ C^{\alpha,\bar{\alpha}}_{\mathrm{fair}}(\eta(X),\bar{\eta}(X),f(X)) \right] .
\end{equation*}

We define $C_{\mathrm{fair}}^{\alpha,\bar{\alpha},*}(\eta,\bar{\eta}):=\inf_{t\in \mathbb{R}}C_{\mathrm{fair}}^{\alpha,\bar{\alpha}}(\eta,\bar{\eta},t)$ to denote the optimal conditional risk. Consequently, $\mathcal{R}_{\mathrm{fair}}^{\alpha,\bar{\alpha},*}=\mathbb{E}_{X}\left[ C_{\mathrm{fair}}^{\alpha,\bar{\alpha},*}(\eta(X),\bar{\eta}(X)) \right]$. 

The same pointwise analysis applies to any surrogate loss pair. 
Recall that
\begin{equation*}
\mathcal{R}_{L}(f;D)=\mathbb{E}_{(X,Y)\sim D}\left[L(Y,f(X))\right], \qquad
\mathcal{R}_{\bar{L}}(f;\bar{D})=\mathbb{E}_{(X,A)\sim \bar{D}}\left[\bar{L}(A,f(X))\right],
\end{equation*}
where $L(y,t)=\mathbf 1\{y=1\}L_{1}(t)+\mathbf 1\{y=-1\}L_{-1}(t)$ and
$\bar{L}(a,t)=\mathbf 1\{a=1\}\bar{L}_{1}(t)+\mathbf 1\{a=-1\}\bar{L}_{-1}(t)$ by the partial-loss decomposition of Eq.~\eqref{eq:surrogate-loss}.

Conditioning on $X=x$, the accuracy term contributes $\eta(x)L_{1}(f(x))+(1-\eta(x))L_{-1}(f(x))$. 

For DP, the fairness term contributes
$\bar{\eta}_{\mathrm{DP}}(x)\bar{L}_{1}(f(x))+(1-\bar{\eta}_{\mathrm{DP}}(x))\bar{L}_{-1}(f(x))$, and hence we define
\begin{equation}\label{eq:dp-conditional-Lrisk}
C_{\mathrm{DP}}^{L,\bar{L}}(\eta,\bar{\eta}_{\mathrm{DP}},t)
:= \left[\eta L_{1}(t)+(1-\eta)L_{-1}(t)\right]
 - \lambda\left[\bar{\eta}_{\mathrm{DP}}\bar{L}_{1}(t)+(1-\bar{\eta}_{\mathrm{DP}})\bar{L}_{-1}(t)\right].
\end{equation}

For EO, the fairness term contributes $\pi^{-1}\eta(x)[\bar{\eta}_{\mathrm{EO}}(x)\bar{L}_{1}(f(x))+(1-\bar{\eta}_{\mathrm{EO}}(x))\bar{L}_{-1}(f(x))]$, and hence we define
\begin{equation}\label{eq:eo-conditional-Lrisk}
C_{\mathrm{EO}}^{L,\bar{L}}(\eta,\bar{\eta}_{\mathrm{EO}},t)
:= \left[\eta L_{1}(t)+(1-\eta)L_{-1}(t)\right]
 - \lambda\pi^{-1}\eta\left[\bar{\eta}_{\mathrm{EO}}\bar{L}_{1}(t)+(1-\bar{\eta}_{\mathrm{EO}})\bar{L}_{-1}(t)\right].
\end{equation}

Recall that $\bar{\eta}$ denote $\bar{\eta}_{\mathrm{DP}}$ or $\bar{\eta}_{\mathrm{EO}}$ according to the fairness notion. By the tower property,
\begin{equation*}
\mathcal{R}_{\mathrm{fair}}^{L,\bar{L}}(f)
= \mathbb{E}_{X}\left[C_{\mathrm{fair}}^{L,\bar{L}}(\eta(X),\bar{\eta}(X),f(X))\right].
\end{equation*}

Similarly, we define $C_{\mathrm{fair}}^{L,\bar{L},*}(\eta,\bar{\eta}):=\inf_{t\in \mathbb{R}}C_{\mathrm{fair}}^{L,\bar{L}}(\eta,\bar{\eta},t)$ to denote the optimal conditional $(L,\bar{L})$-risk. Since $\mathcal{F}$ denote the set of all measurable functions $f:\mathcal{X}\to\mathbb{R}$, consequently, 
\begin{equation*}
\mathcal{R}_{\mathrm{fair}}^{L,\bar{L},*}=\mathbb{E}_{X}\left[ C_{\mathrm{fair}}^{L,\bar{L},*}(\eta(X),\bar{\eta}(X)) \right].
\end{equation*}

\subsubsection{The Attribute-Aware Setting}\label{subsec:fc-aware}

In the attribute-aware setting, the classifier $f$ utilizes both the feature $X$ and the sensitive attribute $A$. For a fixed instance $(X=x, A=a)$, the posterior probability of the positive class becomes $\eta(x,a)=\mathbb{P}(Y=1|X=x,A=a)$. 
Since the true sensitive attribute is given, the inferred probabilities collapse to indicator functions: $\bar{\eta}_{\mathrm{DP}}(x,a) = \mathbb{P}(A=1|X=x,A=a) = \mathbf{1}\{a=1\}$ and $\bar{\eta}_{\mathrm{EO}}(x,a) = \mathbf{1}\{a=1\}$.

Consequently, the conditional risk no longer depends on the posterior probability $\bar{\eta}\in[0,1]$, but directly on the binary sensitive attribute $a \in \{-1, 1\}$. We define the attribute-aware conditional risk as 
\begin{equation}\label{eq:attributed-aware conditonal risk}
C_{\mathrm{fair}}^{\alpha,\bar{\alpha}}(\eta, a, t) = A^{\alpha,\bar{\alpha}}(\eta, a)\mathbf{1}\{t \leq 0\} + B^{\alpha,\bar{\alpha}}(\eta, a)\mathbf{1}\{t > 0\}.
\end{equation}

For DP, the cost functions $A$ and $B$ are simplified to
\begin{align}\label{eq:dp-aware-ab}
A^{\alpha,\bar{\alpha}}_{\mathrm{DP}}(\eta, a) &= \eta(1-\alpha) - \lambda (1-\bar{\alpha}) \mathbf{1}\{a=1\},\\
B^{\alpha,\bar{\alpha}}_{\mathrm{DP}}(\eta, a) &= (1-\eta)\alpha - \lambda \bar{\alpha} (1 - \mathbf{1}\{a=1\}). 
\end{align}

Similarly, for EO, the cost functions are
\begin{align*}
A^{\alpha,\bar{\alpha}}_{\mathrm{EO}}(\eta, a) &= \eta(1-\alpha) - \lambda \pi^{-1}\eta (1-\bar{\alpha}) \mathbf{1}\{a=1\},  \\
B^{\alpha,\bar{\alpha}}_{\mathrm{EO}}(\eta, a) &= (1-\eta)\alpha - \lambda \pi^{-1}\eta \bar{\alpha} (1 - \mathbf{1}\{a=1\}).
\end{align*}

Given an arbitrary surrogate loss pair $(L, \bar{L})$, the attribute-aware conditional $(L,\bar{L})$-risk is given by
\begin{align*}
C_{\mathrm{DP}}^{L,\bar{L}}(\eta, a, t) &= \left[ \eta L_1(t) + (1-\eta)L_{-1}(t) \right] - \lambda \left[ \mathbf{1}\{a=1\} \bar{L}_1(t) + (1 - \mathbf{1}\{a=1\}) \bar{L}_{-1}(t) \right], \\
C_{\mathrm{EO}}^{L,\bar{L}}(\eta, a, t) &= \left[ \eta L_1(t) + (1-\eta)L_{-1}(t) \right] - \lambda \pi^{-1}\eta \left[ \mathbf{1}\{a=1\} \bar{L}_1(t) + (1 - \mathbf{1}\{a=1\}) \bar{L}_{-1}(t) \right].
\end{align*}

\begin{remark}\label{rem:special}
Compared with \eqref{eq:attributed-blind fairness aware conditional risk} and \eqref{eq:attributed-aware conditonal risk}, the attribute-aware conditional risk is mathematically a special case of the attribute-blind conditional risk, where the continuous variable $\bar{\eta} \in [0,1]$ is restricted to the discrete set $\{0, 1\}$. 
\end{remark}

\subsection{Fairness-Aware Fisher Consistency} 

We now use the pointwise conditional risk to define FAFC. For any surrogate loss $L_{\mathrm{fair}}(y,a,t) = L(y,t) - \lambda \bar{L}(a,t)$, define 
the \emph{fairness-aware discriminant} as
\begin{equation}\label{eq:discriminant}
D^{\alpha,\bar{\alpha}}(\eta,\bar{\eta}) := A^{\alpha,\bar{\alpha}}(\eta,\bar{\eta})-B^{\alpha,\bar{\alpha}}(\eta,\bar{\eta}),
\end{equation}
and the \emph{excess conditional risk} as
\begin{equation*}
H_{L,\bar{L}}^{\alpha,\bar{\alpha}}(\eta,\bar{\eta}) := C_{\mathrm{fair}}^{L,\bar{L},-}(\eta,\bar{\eta}) - C_{\mathrm{fair}}^{L,\bar{L},*}(\eta,\bar{\eta})
\end{equation*}
for $\eta,\bar{\eta}\in[0,1]$, where $C_{\mathrm{fair}}^{L,\bar{L},-}(\eta, \bar{\eta}) := \inf_{\{t: t \cdot D^{\alpha,\bar{\alpha}}(\eta, \bar{\eta}) \le 0\}}C_{\mathrm{fair}}^{L,\bar{L}}(\eta, \bar{\eta}, t)$.

Before defining FAFC, we fix the domain on which the pointwise condition is imposed and clarify the role of the trade-off parameter.

\begin{remark}[Well-posedness and the admissible range of $\lambda$]\label{rem:lambda}
For a fixed loss pair $(L,\bar{L})$ and $\lambda\ge0$, let $\mathcal K$ be a domain on which the conditional risk is well posed:
\[
C_{\mathrm{fair}}^{L,\bar{L},*}(\eta,\bar{\eta})>-\infty
\quad\text{for every }(\eta,\bar{\eta})\in\mathcal K,
\]

The parameter $\lambda$ is admissible for a distribution $(D,\bar{D})$ if
\[
(\eta(X),\bar{\eta}(X))\in\left\{(\eta,\bar{\eta}):C_{\mathrm{fair}}^{L,\bar{L},*}(\eta,\bar{\eta})>-\infty\right\}
\quad\text{almost surely}.
\]
For a fixed distribution this is a condition on $\lambda$, not on the classifier. When $\bar L\ge0$, the maximal well-posed domain shrinks as $\lambda$ increases, so the admissible $\lambda$ form an interval starting at $0$. Its right endpoint is
\[
\lambda_{\max}:=\sup\left\{\lambda\ge0:\ (\eta(X),\bar{\eta}(X))\in\mathcal K\ \text{almost surely}\right\}.
\]
For the linear $\bar L$ in the examples, the same conclusion follows from the explicit linear inequalities, and this endpoint is computed there. If the condition fails, then $C_{\mathrm{fair}}^{L,\bar{L},*}=-\infty$ on a set of positive probability, so $\mathcal{R}_{\mathrm{fair}}^{L,\bar{L},*}=-\infty$ and the excess risk is undefined. All subsequent results are stated for admissible $\lambda$.
\end{remark}

On $\mathcal K$, $H_{L,\bar{L}}^{\alpha,\bar{\alpha}}(\eta,\bar{\eta})\geq0$, and it is finite by the definition of $\mathcal K$.

\begin{definition}[FAFC]\label{def:FAFC}
For a fixed $\lambda$, let $\mathcal K$ be a well-posed domain as in Remark~\ref{rem:lambda}. $L_{\mathrm{fair}}$ is $(\alpha,\bar{\alpha})$-FAFC on $\mathcal K$ with respect to $\alpha,\bar{\alpha}\in(0,1)$ if
\[
H_{L,\bar{L}}^{\alpha,\bar{\alpha}}(\eta,\bar{\eta})>0
\quad\text{for all }(\eta,\bar{\eta})\in\mathcal K\text{ with }D^{\alpha,\bar{\alpha}}(\eta,\bar{\eta})\neq0.
\]
\end{definition}

Intuitively, $L_{\mathrm{fair}}$ is $(\alpha,\bar{\alpha})$-FAFC on $\mathcal K$ if, at every point $(\eta,\bar{\eta})\in\mathcal K$ with $D^{\alpha,\bar{\alpha}}(\eta,\bar{\eta})\neq 0$, the value of $t=f(x)$ minimizing the conditional $(L,\bar{L})$-risk has the same sign as $D^{\alpha,\bar{\alpha}}(\eta,\bar{\eta})$. Corollary~\ref{cor:FAFC} shows that this pointwise requirement is inherited by every minimizer of the population risk whenever the support of $(\eta(X),\bar{\eta}(X))$ lies in $\mathcal K$.

\begin{corollary}\label{cor:FAFC}
Recall the definition of fairness-aware discriminant $D^{\alpha,\bar{\alpha}}$ in \eqref{eq:discriminant}. If $L_{\mathrm{fair}}$ is $(\alpha,\bar{\alpha})$-FAFC on $\mathcal K$, then for any minimizer $f^{*}\in\arg\min_{f\in\mathcal{F}}\mathcal{R}_{\mathrm{fair}}^{L,\bar{L}}(f;D,\bar{D})$, its sign matches the fairness-aware discriminant:
\begin{equation*}
\mathrm{sign}(f^{*}(x))=\mathrm{sign}\big(D^{\alpha,\bar{\alpha}}(\eta(x),\bar{\eta}(x))\big)
\quad\text{for }P_X\text{-a.e. }x\text{ with }D^{\alpha,\bar{\alpha}}(\eta(x),\bar{\eta}(x))\neq 0.
\end{equation*}
\end{corollary}

Note that Corollary~\ref{cor:FAFC} is one-directional for a fixed distribution: FAFC on $\mathcal K$ implies the population-level statement, but the population-level statement only constrains $(\eta,\bar{\eta})$ almost surely on the support. Example~\ref{ex:non-equiv} (See Appendix~\ref{app:pop-vs-pointwise}) illustrates this fixed-distribution gap. 

The following theorem provides a necessary and sufficient condition for FAFC on $\mathcal K$. The condition is stated purely in terms of the derivatives of the four partial losses $L_1,L_{-1},\bar{L}_1,\bar{L}_{-1}$ at $0$.

\begin{theorem}\label{thm:fair-fc-condition}
Let $\alpha, \bar{\alpha} \in (0, 1)$, $\lambda>0$, let $\mathcal K$ be a well-posed domain as in Remark~\ref{rem:lambda}, and let $L_{\mathrm{fair}}$ be a fairness-aware surrogate loss. Assume the partial losses $L_1, L_{-1}, \bar{L}_1, \bar{L}_{-1}$ are convex and differentiable at $0$. Set $r:=\lambda/\pi$, where $\pi=\mathbb{P}(Y=1)$.
Furthermore, suppose that, for every $(\eta,\bar{\eta})\in\mathcal K$, the map $t \mapsto C_{\mathrm{fair}}^{L,\bar{L}}(\eta,\bar{\eta},t)$ is strictly convex on $\mathbb{R}$, and that $\mathcal K$ contains a neighborhood of $(\alpha,\bar{\alpha})$.
Then, in the attribute-blind setting, $L_{\mathrm{fair}}$ is $(\alpha, \bar{\alpha})$-FAFC on $\mathcal K$ if and only if the following conditions hold for the corresponding criterion:
\begin{align*}
\text{(DP):}\quad
& \text{(D1)}\ L'_1(0)<L'_{-1}(0),\\
& \text{(D2)}\ L'_1(0)-L'_{-1}(0)=\bar L'_1(0)-\bar L'_{-1}(0),\\
& \text{(D3)}\ \alpha L'_1(0)+(1-\alpha)L'_{-1}(0)
=\lambda\bigl[\bar\alpha\bar L'_1(0)+(1-\bar\alpha)\bar L'_{-1}(0)\bigr];\\[4pt]
\text{(EO):}\quad
& \text{(E1)}\ \bar L'_1(0)-\bar L'_{-1}(0)<0,\\
& \text{(E2)}\ L'_1(0)-L'_{-1}(0)
=\bar L'_1(0)-\bar L'_{-1}(0)
+r\bigl[\bar\alpha\bar L'_1(0)+(1-\bar\alpha)\bar L'_{-1}(0)\bigr],\\
& \text{(E3)}\ \alpha L'_1(0)+(1-\alpha)L'_{-1}(0)
=r\alpha\bigl[\bar\alpha\bar L'_1(0)+(1-\bar\alpha)\bar L'_{-1}(0)\bigr].
\end{align*}
\end{theorem}

\begin{remark}[Attribute-aware setting]\label{rem:aware-fafc}
Theorem~\ref{thm:fair-fc-condition} is stated for the attribute-blind setting. Its sufficiency direction applies verbatim to the attribute-aware domain. For necessity, one additional nondegeneracy condition is required: for each \(a\in\{-1,1\}\), the section \(\mathcal K_a=\{\eta:(\eta,a)\in\mathcal K\}\) is an interval, and there exists an interior point \(\eta_a\) of the interval \(\mathcal K_a\) such that \(D^{\alpha,\bar\alpha}(\eta_a,a)=0\). This condition ensures that the decision boundary crosses each attribute edge, so that FAFC controls both sides of the boundary. Since \(D(\cdot,a)\) is affine, it fails only when the boundary misses an entire edge or meets it at an endpoint; it therefore restricts neither the loss class nor the data distribution and is a mild nondegeneracy condition. It holds, for instance, in the weighted-logistic example below. With this condition, the proof is the edgewise analogue of the attribute-blind proof.
\end{remark}

Theorem~\ref{thm:fair-fc-condition} provides the Fisher consistency property for the fairness-aware setting: whenever the fairness-regularized objective is well-posed, standard surrogate losses retain their consistency guarantees provided the loss pair satisfies the corresponding derivative conditions.

\begin{remark}[Connection to Classical Fisher Consistency]\label{rem:classical-cc}
Each condition in Theorem~\ref{thm:fair-fc-condition} traces back to the characterization of classical Fisher consistency (Theorem~\ref{thm:cs-cc-prelim}). We discuss the DP case; the EO case reduces to it when $\bar{\alpha}\bar{L}'_1(0)+(1-\bar{\alpha})\bar{L}'_{-1}(0)=0$, which holds in the examples below. In the degenerate case $\lambda=0$ (equivalently, as $\lambda\to0$ in the conditions above), (D3) is the classical decision-boundary condition $\alpha L_{1}'(0)+(1-\alpha)L_{-1}'(0)=0$.
(D1) is the classical sign requirement $L_{1}'(0)<0<L_{-1}'(0)$, written in the equivalent form $L_{1}'(0)<L_{-1}'(0)$, which suffices once (D3) holds.
(D2) has no classical counterpart: it is the fairness-specific requirement that the two surrogate losses share the same derivative gap at $0$.
\end{remark}

\section{Calibration Inequality for Fairness-Aware Classification}\label{sec:calibration}

Having characterized when the composite fairness-aware loss $L_{\mathrm{fair}}$ is FAFC, we now answer the next central question: establishing a quantitative bound between the excess fairness-aware risk and the excess $(L,\bar{L})$-risk under the attribute-blind setting.
Then, the attribute-aware setting will be discussed in Remark \ref{rem:ext}.

Throughout this section, $\mathcal K\subseteq[0,1]^2$ is a compact, connected subset of a well-posed domain as in Remark~\ref{rem:lambda} that contains $(\alpha,\bar{\alpha})$, and every distribution considered satisfies $(\eta(X),\bar{\eta}(X))\in\mathcal K$ almost surely. Define the maximum discriminant value as $B_{\max}:=\max_{(\eta,\bar{\eta})\in \mathcal K}\left| D^{\alpha,\bar{\alpha}}(\eta,\bar{\eta}) \right|$. For any $\epsilon\in[0,B_{\max}]$, the calibration function $\nu_{L,\bar{L}}^{\alpha, \bar{\alpha}}:[0,B_{\max}]\to[0,\infty]$ is defined as 
\begin{equation*}
    \nu_{L,\bar{L}}^{\alpha, \bar{\alpha}}(\epsilon): = \inf_{(\eta,\bar{\eta})\in\mathcal K:\left| D^{\left( \alpha,\bar{\alpha} \right)}(\eta,\bar{\eta}) \right| = \epsilon} H_{L,\bar{L}}^{\alpha,\bar{\alpha}}(\eta,\bar{\eta}).
\end{equation*}

If the level set $\left\{ (\eta,\bar{\eta})\in \mathcal K:\left| D^{\alpha,\bar{\alpha}}(\eta,\bar{\eta}) \right|=\epsilon \right\}$ is empty for some $\epsilon \in (0,B_{\max}]$, we define $\nu_{L,\bar{L}}^{\alpha,\bar{\alpha}}(\epsilon)=\infty$. 

Since the convexity of $\nu_{L,\bar{L}}^{\alpha, \bar{\alpha}}$ cannot be guaranteed in general, we require its convex envelope to utilize Jensen's inequality for the excess risk bound. Concretely, we define 
$$\psi_{L,\bar{L}}^{\alpha, \bar{\alpha}}:=(\nu ^{\alpha,\bar{\alpha}}_{L,\bar{L}})^{**},\ \text{for all}\ \epsilon\in[0,B_{\max}],$$ 
where $g^{**}$ denotes the Fenchel-Legendre biconjugate of $g$. Equivalently, $\psi_{L,\bar{L}}^{\alpha,\bar{\alpha}}$ is the largest convex, lower semi-continuous function majorized by $\nu_{L,\bar{L}}^{\alpha,\bar{\alpha}}$, and it can be characterized via the relation $\mathrm{Epi}\, g ^{**} = \overline{\mathrm{co\, Epi}\, g}$, where $\mathrm{Epi}\,g=\left\{ (r,s):g(r)\leq s \right\}$ denotes the epigraph of $g$; ``co" stands for the convex hull of a set, and the overbar signifies the closure of the set. 

With $\nu_{L,\bar{L}}^{\alpha, \bar{\alpha}}$ and its convex envelope $\psi_{L,\bar{L}}^{\alpha, \bar{\alpha}}$ defined, we now state the theorem relating the fairness-aware excess risk to the excess $(L,\bar{L})$-risk.

\begin{theorem}\label{thm:fair-bound}
Let $\alpha,\bar{\alpha}\in(0,1)$, let $L_{\mathrm{fair}}=L-\lambda \bar{L}$ be a fairness-aware loss, and let $\mathcal K$ be as above.
\begin{enumerate}
    \item For any $f\in \mathcal{F}$ and distribution $D,\bar{D}$ with $(\eta(X),\bar{\eta}(X))\in\mathcal K$ almost surely, we have 
    \[
    \psi_{L,\bar{L}}^{\alpha, \bar{\alpha}} \left( \mathcal{R}^{\alpha,\bar{\alpha}}_{\mathrm{fair}}(f) - \mathcal{R}_{\mathrm{fair}}^{\alpha,\bar{\alpha},*} \right) \leq \mathcal{R}_{\mathrm{fair}}^{L,\bar{L}}(f) - \mathcal{R}_{\mathrm{fair}}^{L,\bar{L},*}.
    \]
    \item Furthermore, if the map $t\mapsto C_{\mathrm{fair}}^{L,\bar{L}}(\eta,\bar{\eta},t)$ is convex on $\mathbb{R}$, then $\psi_{L,\bar{L}}^{\alpha, \bar{\alpha}}$ is invertible if and only if $L_{\mathrm{fair}}$ is $(\alpha,\bar{\alpha})$-FAFC on $\mathcal K$.
\end{enumerate}
\end{theorem}

\begin{remark}[Tighter Bounds for Attribute-Aware Classification]\label{rem:ext}
While the calibration inequality holds for both settings, an intrinsic relationship exists between their respective calibration functions. 
For the attribute-aware setting we take $\mathcal K\subseteq[0,1]\times\{-1,1\}$ and let
\begin{equation*}
\nu_{L,\bar{L}}^{\alpha,\bar{\alpha},\mathrm{aw}}(\epsilon) := \inf_{\substack{(\eta,a)\in\mathcal K\\ \left| D^{\alpha,\bar{\alpha}}(\eta,\mathbf{1}\{a=1\}) \right|=\epsilon}} H_{L,\bar{L}}^{\alpha,\bar{\alpha}}(\eta,\mathbf{1}\{a=1\}),
\end{equation*}
Taking this domain to be the intersection of the attribute-blind $\mathcal K$ with the two edges, the attribute-aware level sets are subsets of the attribute-blind ones, so minimizing over them yields a value no smaller, and hence
$\nu_{L,\bar{L}}^{\alpha,\bar{\alpha},\mathrm{aw}}(\epsilon)\ge\nu_{L,\bar{L}}^{\alpha,\bar{\alpha}}(\epsilon)$.
Because the Fenchel-Legendre biconjugate is monotone with respect to pointwise order, we conclude that $\psi_{L,\bar{L}}^{\alpha,\bar{\alpha},\mathrm{aw}}(\epsilon)\ge\psi_{L,\bar{L}}^{\alpha,\bar{\alpha}}(\epsilon)$.
This reveals a theoretical insight: the calibration bound for the attribute-aware setting is no looser. If $L_{\mathrm{fair}}$ is $(\alpha,\bar{\alpha})$-FAFC in the attribute-blind setting, it is automatically $(\alpha,\bar{\alpha})$-FAFC in the attribute-aware setting.
\end{remark}

\section{Application to Fairness-Aware Classification}\label{sec:application}

This section applies the framework of Sections \ref{sec:fc-fair} and \ref{sec:calibration} to concrete classifiers. To illustrate the FAFC conditions in a concrete model, we first revisit the fairness-aware logistic regression proposed by \citet{Zafar-2019-FairnessConstraintsFlexible} and verify FAFC on the relevant domain for its composite loss $L_{\mathrm{fair}}$, including the well-posedness range of $\lambda$ and a closed-form calibration function. We then present a general surrogate loss-based fairness-aware classification algorithm that can be used with any surrogate loss pair $(L,\bar{L})$ and trade-off $\lambda$ for which $L_{\mathrm{fair}}$ is $(\alpha,\bar{\alpha})$-FAFC on a domain containing the support.
\subsection{Fairness-Aware Logistic Regression}\label{subsec:logistic}

We begin with fairness-aware logistic regression under DP in the attribute-blind setting---the configuration studied by \citet{Zafar-2019-FairnessConstraintsFlexible}.
Let the accuracy surrogate be $L(y,t)=\log(1+e^{-yt})$ and the fairness surrogate be $\bar{L}(a,t)=\left(\pi_A - \frac{a+1}{2}\right)t$, where $\pi_A=\mathbb{P}(A=1)$.
The fairness-aware surrogate loss is
\begin{equation}\label{eq:fair-logistic-loss}
    L_{\mathrm{fair}}(y,a,t) = L(y,t)-\lambda\bar{L}(a,t)=\log(1+e^{ -yt }) - \lambda\left(\pi_A - \frac{a+1}{2}\right)t.
\end{equation}
This pair fits the framework of Section~\ref{subsec:fair-surrogate}. The partial losses are $L_1(t)=\log(1+e^{-t})$, $L_{-1}(t)=\log(1+e^{t})$, $\bar{L}_1(t)=(\pi_A-1)t$, and $\bar{L}_{-1}(t)=\pi_A t$. The following proposition proves that the logistic composite loss $L_{\mathrm{fair}}$ is FAFC on the domain $\mathcal K$ defined below.

\begin{proposition}\label{prop:zafar-logistic-fafc}
Let $(L,\bar{L})$ be the logistic surrogate loss pair defined in Eq.~\eqref{eq:fair-logistic-loss}, and let
\[
\mathcal K:=\bigl\{(\eta,\bar{\eta})\in[0,1]^2:\ 0\le \eta-\lambda(\bar{\eta}-\pi_A)\le1\bigr\}.
\]
If $0<\lambda<\lambda_{\max}$, with $\lambda_{\max}$ given in \eqref{eq:lambda-max}, then $(\eta(X),\bar{\eta}(X))\in\mathcal K$ almost surely, and $L_{\mathrm{fair}}$ is $(0.5,\pi_A)$-FAFC on $\mathcal K$ for the DP criterion in the attribute-blind setting.
\end{proposition}

\begin{proof}
We verify (D1)--(D3) of Theorem~\ref{thm:fair-fc-condition}. (D1) holds since $L'_1(0)=-\frac12<L'_{-1}(0)=\frac12$. (D2) holds since $L'_1(0)-L'_{-1}(0)=-1=\bar{L}'_1(0)-\bar{L}'_{-1}(0)$. 

For (D3), with $\alpha=0.5$,
\[
(1-\alpha)L'_{-1}(0)+\alpha L'_1(0)=\frac12\cdot\frac12+\frac12\cdot(-\frac12)=0,
\]
and with $\bar{\alpha}=\pi_A$,
\[
(1-\bar{\alpha})\bar{L}'_{-1}(0)+\bar{\alpha}\bar{L}'_1(0)=(1-\pi_A)\pi_A+\pi_A(\pi_A-1)=0.
\]

Hence (D3)
\[
(1-\alpha)L'_{-1}(0)+\alpha L'_1(0) = \lambda \bigl[(1-\bar{\alpha})\bar{L}'_{-1}(0)+\bar{\alpha}\bar{L}'_1(0)\bigr],
\]
holds.
On $\mathcal K$ the conditional risk is strictly convex and bounded below: strict convexity is shown in the next paragraph for all $(\eta,\bar\eta)\in[0,1]^2$, and boundedness below follows from $0\le\eta-\lambda(\bar\eta-\pi_A)\le1$ and the tail expansion below. Hence the sufficiency direction of Theorem~\ref{thm:fair-fc-condition} gives that $L_{\mathrm{fair}}$ is $(0.5,\pi_A)$-FAFC on $\mathcal K$. The inclusion of the support in $\mathcal K$ is verified in the next paragraph.
\end{proof}

It remains to ensure that the fairness-aware risk is well-posed (Remark~\ref{rem:lambda}). Substituting the surrogate loss pair into the conditional $(L,\bar L)$-risk gives
\begin{align*}
    C_{\mathrm{fair}}^{L,\bar{L}}(\eta,\bar{\eta},t) & = \eta\log(1+e^{ -t }) + (1-\eta)\log(1+e^{ t }) - \lambda\bigl[ \bar{\eta}(\pi_A-1)t+(1-\bar{\eta})\pi_A t \bigr]  \\
    & = \eta\log(1+e^{ -t }) + (1-\eta)\log(1+e^{ t }) + \lambda(\bar{\eta}-\pi_A)t .
\end{align*}

The conditional risk is strictly convex in $t$, since the second derivatives of $\log(1+e^{-t})$ and $\log(1+e^{t})$ both equal $e^{t}/(1+e^{t})^{2}$,
\begin{equation*}
    \frac{\partial^{2}}{\partial t^{2}}C_{\mathrm{fair}}^{L,\bar L}(\eta,\bar\eta,t)
    = \frac{e^{t}}{(1+e^{t})^{2}} > 0
\end{equation*}
for all $(\eta,\bar\eta)\in[0,1]^2$ and $t\in\mathbb{R}$.

With $S:=\eta-\lambda(\bar\eta-\pi_A)$, the conditional risk can be written as
\begin{equation*}
    C_{\mathrm{fair}}^{L,\bar L}(\eta,\bar\eta,t)=\log(1+e^{t})-S t.
\end{equation*}
Since $\log(1+e^{t})=t+o(t)$ as $t\to+\infty$ and $\log(1+e^{t})=o(t)$ as $t\to-\infty$, the tails satisfy
\begin{equation*}
    C_{\mathrm{fair}}^{L,\bar L}(\eta,\bar\eta,t)=
    \begin{cases}
        (1-S)\,t+o(t), & t\to+\infty,\\
        -S\,t+o(t), & t\to-\infty,
    \end{cases}
\end{equation*}
so the conditional risk is bounded below if and only if $0\le S\le1$. Because $\sigma(\mathbb{R})=(0,1)$ and the conditional risk is strictly convex, the pointwise minimizer is uniquely characterized by $\sigma(t^{*})=S$ precisely when $0<S<1$. Consequently, we require $0<S(x)<1$ for $P_X$-a.e. $x$. For $\lambda>0$, this is ensured by $\lambda<\lambda_{\max}$, where
\begin{equation}\label{eq:lambda-max}
    \lambda_{\max}:=\min\Bigg\{ \inf_{\substack{x\in\mathrm{supp}(X):\\ \bar{\eta}(x)>\pi_A}}\frac{\eta(x)}{\bar{\eta}(x)-\pi_A},\quad \inf_{\substack{x\in\mathrm{supp}(X):\\ \bar{\eta}(x)<\pi_A}}\frac{1-\eta(x)}{\pi_A-\bar{\eta}(x)} \Bigg\},
\end{equation}
and the infimum over an empty set is understood as $+\infty$. Under $\lambda\in(0,\lambda_{\max})$, the conditional risk is strictly convex with a unique pointwise minimizer characterized by $\sigma(t^{*})=S\in(0,1)$, hence $\mathrm{sign}(t^{*})=\mathrm{sign}(D^{\alpha,\bar{\alpha}})$ and the support of $(\eta(X),\bar{\eta}(X))$ lies in the interior of $\mathcal K$. Both bounds in \eqref{eq:lambda-max} are support-dependent: the first degenerates as $\eta(x)\to0$ on $\{\bar{\eta}(x)>\pi_A\}$ and the second degenerates as $\eta(x)\to1$ on $\{\bar{\eta}(x)<\pi_A\}$, so no bound on $\lambda$ that is independent of the support can be given.

Having verified FAFC on $\mathcal K$ for $L_{\mathrm{fair}}$, we now derive the calibration function and excess-risk bound for this example.

\begin{proposition}\label{prop:zafar-logistic-calibration}
Suppose $0<\lambda<\lambda_{\max}$, and let $\mathcal K$ be the domain of Proposition~\ref{prop:zafar-logistic-fafc}. Then, on $\mathcal K$, for every $\epsilon\in[0,0.5)$ the calibration function of the pair $(L,\bar{L})$ is
\begin{equation*}
    \nu_{L,\bar{L}}^{\alpha,\bar{\alpha}}(\epsilon) = (0.5+\epsilon)\ln(1+2\epsilon) + (0.5-\epsilon)\ln(1-2\epsilon).
\end{equation*}

Its convex envelope coincides with it, $\psi_{L,\bar{L}}^{\alpha,\bar{\alpha}}=\nu_{L,\bar{L}}^{\alpha,\bar{\alpha}}$, the quadratic lower bound $\psi_{L,\bar{L}}^{\alpha,\bar{\alpha}}(\epsilon)\ge 2\epsilon^{2}$ holds, and consequently, for every $f\in\mathcal{F}$,
\begin{equation}\label{eq:zafar-bound}
    \mathcal{R}_{\mathrm{fair}}^{\alpha,\bar{\alpha}}(f) - \mathcal{R}_{\mathrm{fair}}^{\alpha,\bar{\alpha},*}
    \;\le\;
    \min\Bigg\{ \frac{1}{2},\; \sqrt{\frac{1}{2}\Bigl(\mathcal{R}_{\mathrm{fair}}^{L,\bar{L}}(f)-\mathcal{R}_{\mathrm{fair}}^{L,\bar{L},*}\Bigr)} \Bigg\}.
\end{equation}
\end{proposition}

\begin{proof}

Since $C_{\mathrm{fair}}^{L,\bar L}(\eta,\bar\eta,t)=\log(1+e^{t})-St$ and $S=\eta-\lambda(\bar{\eta}-\pi_A)$, then we have the derivative of the conditional risk is
\begin{equation*}
    \frac{\partial}{\partial t}C_{\mathrm{fair}}^{L,\bar L}(\eta,\bar\eta,t) = \sigma(t)-S,
\end{equation*}
where $\sigma(t)=1/(1+e^{-t})$. Setting the derivative to zero yields $\sigma(t^{*}) =S$, and $D^{\alpha,\bar{\alpha}}(\eta,\bar{\eta})=\eta-0.5-\lambda(\bar{\eta}-\pi_A)$. Hence $D^{\alpha,\bar{\alpha}}=S-0.5$. 
So, we can evaluate the conditional risk at the unique minimizer gives
\begin{equation*}
    C_{\mathrm{fair}}^{L,\bar{L},*} (\eta,\bar{\eta}) = -S\log S-(1-S)\log(1-S),
\end{equation*}
and the wrong-sign infimum is attained at $t=0$:
\begin{equation*}
    C_{\mathrm{fair}}^{L,\bar{L},-} (\eta,\bar{\eta}) = C_{\mathrm{fair}}^{L,\bar{L}}(\eta,\bar{\eta},0 ) = \log 2.
\end{equation*}

Hence the wrong-sign excess conditional $(L,\bar{L})$-risk is
\begin{equation*}
    H_{L,\bar{L}}^{\alpha,\bar{\alpha}}(\eta,\bar{\eta}) = S \log S + (1-S) \log(1-S) + \log 2 .
\end{equation*}
Note that $H_{L,\bar{L}}^{\alpha,\bar{\alpha}}$ and $D^{\alpha,\bar{\alpha}}$ depend on $\eta$ and $\bar{\eta}$ solely through $S$.

To find $\nu_{L,\bar{L}}^{\alpha,\bar{\alpha}}(\epsilon)$, we minimize $H_{L,\bar{L}}^{\alpha,\bar{\alpha}}$ over the set $\{(\eta, \bar{\eta}) : |S-0.5| = \epsilon\}$. This constraint implies $S = 0.5 \pm \epsilon$. Since $H$ depends on $S$ only through the symmetric quantity $S \log S + (1-S)\log(1-S) + \log 2$, we have $H_{L,\bar{L}}^{\alpha,\bar{\alpha}}(0.5+\epsilon) = H_{L,\bar{L}}^{\alpha,\bar{\alpha}}(0.5-\epsilon)$. Thus, the minimization yields the exact calibration function:
\begin{equation*}
    \nu_{L,\bar{L}}^{\alpha,\bar{\alpha}}(\epsilon) = (0.5+\epsilon)\ln(1+2\epsilon) + (0.5-\epsilon)\ln(1-2\epsilon).
\end{equation*}

$\nu_{L,\bar{L}}^{\alpha,\bar{\alpha}}$ is convex on $[0,0.5)$, since its second derivative equals $\nu''=4/(1-4\epsilon^2)>0$. So $\psi_{L,\bar{L}}^{\alpha,\bar{\alpha}}(\epsilon)=\nu_{L,\bar{L}}^{\alpha,\bar{\alpha}}(\epsilon)$.

Finally, noting that $H_{L,\bar{L}}^{\alpha,\bar{\alpha}}$ is the KL divergence between the Bernoulli distributions with parameters $S$ and $0.5$, and applying Pinsker's inequality $D_{\mathrm{KL}}(p\|q) \ge 2(p-q)^2$, we obtain the quadratic lower bound
\begin{equation*}
    \psi_{L,\bar{L}}^{\alpha,\bar{\alpha}}(\epsilon) = \nu_{L,\bar{L}}^{\alpha,\bar{\alpha}}(\epsilon) \ge 2( (0.5+\epsilon) - 0.5 )^2 = 2\epsilon^2.
\end{equation*}

Combining this lower bound with the calibration inequality of Theorem~\ref{thm:fair-bound} (which applies because the support lies in $\mathcal K$) yields $2\bigl(\mathcal{R}_{\mathrm{fair}}^{\alpha,\bar{\alpha}}(f)-\mathcal{R}_{\mathrm{fair}}^{\alpha,\bar{\alpha},*}\bigr)^{2}\le \mathcal{R}_{\mathrm{fair}}^{L,\bar{L}}(f)-\mathcal{R}_{\mathrm{fair}}^{L,\bar{L},*}$, i.e.\ $\mathcal{R}_{\mathrm{fair}}^{\alpha,\bar{\alpha}}(f)-\mathcal{R}_{\mathrm{fair}}^{\alpha,\bar{\alpha},*}\le\sqrt{\frac12\bigl(\mathcal{R}_{\mathrm{fair}}^{L,\bar{L}}(f)-\mathcal{R}_{\mathrm{fair}}^{L,\bar{L},*}\bigr)}$. 

By Lemma~\ref{lem:excess-risk}, since $\lvert D^{\alpha,\bar{\alpha}}\rvert=\lvert S-\frac12\rvert<\frac12$, the trivial bound $\mathcal{R}_{\mathrm{fair}}^{\alpha,\bar{\alpha}}(f)-\mathcal{R}_{\mathrm{fair}}^{\alpha,\bar{\alpha},*}\le\frac12$ holds as well, and taking the minimum of the two gives \eqref{eq:zafar-bound}. 
\end{proof}

Up to now, we have focused on the DP criterion and the attribute-blind setting of fairness-aware logistic regression; the EO criterion, the attribute-aware setting, and further examples of FAFC composite losses are deferred (e.g., hinge and exponential loss) to Appendix~\ref{app:other-examples}.

\subsection{A General Surrogate Loss-Based Fairness-Aware Classification Algorithm}\label{subsec:algorithm}

The fairness-aware risk in Eq.~\eqref{eq:fair-risk-01} is built from discontinuous 0--1 losses and is not amenable to gradient-based learning from finite samples.
The surrogate formulation in Section~\ref{subsec:fair-surrogate} instead minimizes the population $(L,\bar{L})$-risk $\mathcal{R}_{\mathrm{fair}}^{L,\bar{L}}(f;D,\bar{D})$.
When $L_{\mathrm{fair}}$ is $(\alpha,\bar{\alpha})$-FAFC on $\mathcal K$, every population minimizer aligns with the fairness-aware discriminant (Corollary~\ref{cor:FAFC}); when in addition the calibration inequality of Theorem~\ref{thm:fair-bound} applies, controlling the excess $(L,\bar{L})$-risk controls the excess fairness-aware risk.
Algorithm~\ref{alg:general-fair} implements the same principle empirically: it replaces the population objective by its sample average, parameterizes the score as $f_\theta$, and minimizes the resulting smooth (or piecewise smooth) empirical $(L,\bar{L})$-risk, optionally with a penalty $\Omega(\theta)$ for stability.
The DP and EO criteria enter only through the construction of the fairness term in Step~2, mirroring the population definitions in Section~\ref{subsec:prelims-fairness-problem}; any pair $(L,\bar{L})$ and trade-off $\lambda$ for which $L_{\mathrm{fair}}$ is FAFC can be plugged in at Step~1.
Step~3 then calls a standard unconstrained optimizer (e.g., L-BFGS when the objective is differentiable).

\begin{algorithm}[htb!]

\caption{General Surrogate Loss-Based Fairness-Aware Classification Algorithm}
\label{alg:general-fair}
\begin{algorithmic}
\begin{small}
    \STATE \textbf{Input}: 
    \begin{enumerate}
        \item Cost parameters $\alpha, \bar{\alpha} \in (0,1)$;
        \item Dataset $D = \{(x_i, a_i, y_i)\}_{i=1}^n$, where $x_i$ denotes the feature, $a_i$ the protected attribute, and $y_i$ the label;
        \item Trade-off parameter $\lambda \geq 0$;
        \item Fairness criterion ($\text{DP}$ or $\text{EO}$).
    \end{enumerate}
    \STATE \textbf{Step 1 (Select the fairness-aware surrogate loss pair)}: Select a pair of fairness-aware surrogate loss functions $(L, \bar{L})$ such that, for the given trade-off $\lambda$, $L_{\mathrm{fair}}$ is $(\alpha, \bar{\alpha})$-FAFC on a domain $\mathcal K$ containing the support of $(\eta(X),\bar{\eta}(X))$.
    \STATE \textbf{Step 2 (Construct the empirical $(L,\bar{L})$-risk objective)}: 
    \begin{enumerate}
        \item For the $\text{DP}$ fairness criterion, construct the empirical risk as:
        $$
        \hat{\mathcal{R}}_{\mathrm{DP}}^{L,\bar{L}}(f_\theta) = \frac{1}{n}\sum_{i=1}^n L(y_i, f_\theta(x_i)) - \frac{\lambda}{n}\sum_{i=1}^n \bar{L}(a_i, f_\theta(x_i));
        $$
        \item For the $\text{EO}$ fairness criterion, first estimate the positive class prior $\hat{\pi} = \frac{1}{n}\sum_{i=1}^n \mathbf{1}\{y_i=1\}$, then construct the empirical risk as:
        $$
        \hat{\mathcal{R}}_{\mathrm{EO}}^{L,\bar{L}}(f_\theta) = \frac{1}{n}\sum_{i=1}^n L(y_i, f_\theta(x_i)) - \frac{\lambda}{\hat{\pi} n}\sum_{i: y_i=1} \bar{L}(a_i, f_\theta(x_i)).
        $$
    \end{enumerate}
    \STATE \textbf{Step 3 (Solve the unconstrained optimization problem)}: Solve the following problem using an optimization algorithm (e.g., L-BFGS):
    $$
    \theta^* = \arg\min_{\theta} \hat{\mathcal{R}}_{\mathrm{fair}}^{L,\bar{L}}(f_\theta) + \Omega(\theta),
    $$
    where $\Omega(\theta)$ is an optional regularization term (e.g., $L_2$ regularization).
    \STATE \textbf{Output}: Decision function $\hat{f}^*(x) = f_{\theta^*}(x)$ and classification rule $\hat{Y} = \mathrm{sign}(\hat{f}^*(x))$.

    \end{small}
\end{algorithmic}
\end{algorithm}

Section~\ref{sec:experiment} turns to numerical experiments that show the empirical performance of Algorithm~\ref{alg:general-fair}.

\section{Numerical Experiments}\label{sec:experiment}

This section aims to validate the theoretical properties and the practical performance of Algorithm~\ref{alg:general-fair},
through simulation experiments and a real-data study on credit card default prediction. 

\subsection{Simulation}

Two distinct data-generating processes are constructed to correspond to the DP and EO fairness criteria, respectively. The first setup adopts the fairness-aware logistic loss under the attribute-blind setting for DP (Section~\ref{subsec:logistic}), while the second setup employs the fairness-aware exponential loss under the attribute-aware setting for EO (Appendix~\ref{app:exponential}). 
\subsubsection{Fairness-Aware Logistic Loss}

We generate binary class labels $Y\in\{-1,1\}$ uniformly at random and assign a two-dimensional feature vector per label sampled from two Gaussian distributions:
\begin{equation*}
\begin{aligned}
X\mid Y=1 &\sim N\left(\begin{pmatrix}0\\3\end{pmatrix},\begin{pmatrix}4&0\\0&6\end{pmatrix}\right),\\
X\mid Y=-1 &\sim N\left(\begin{pmatrix}0\\-3\end{pmatrix},\begin{pmatrix}4&0\\0&6\end{pmatrix}\right).
\end{aligned}
\end{equation*}

Then, we restrict the population to the box $\mathcal{B}=[-6,6]^{2}$ by conditioning the Gaussians on $X\in\mathcal{B}$, using the conditional distribution over $\mathcal{B}$ rather than truncated samples. This bounded support is required because the fairness surrogate loss $\bar{L}$ is linear in the score, and unbounded support would make the conditional risk unbounded below, collapsing $\lambda_{\max}$ in \eqref{eq:lambda-max} to zero. Over $\mathcal{B}$, the well-posedness threshold \eqref{eq:lambda-max} satisfies $\lambda_{\max}=1.1808$. 
The sensitive attribute is drawn from a Bernoulli distribution,
\begin{equation*}
A \mid X=x \sim \text{Binomial}\big(1,\, \sigma(w^\top x + b)\big),
\end{equation*}
where $\sigma(u)=1/(1+e^{-u})$ denotes the logistic function and $ w=(-0.25,0.6),b=0.1$. The resulting marginal probability is $\pi_A=\mathbb{P}(A=1)=\mathbb{E}[\sigma(w^{\top}X+b)]=0.5139$. We set $\alpha=0.5$, $\bar\alpha=\pi_A$ and $\lambda=0.6$.

Let $(L,\bar{L})$ be the logistic surrogate loss pair defined in Eq.~\eqref{eq:fair-logistic-loss}. We minimize the empirical $(L,\bar{L})$-risk of Algorithm~\ref{alg:general-fair} for the DP criterion over the linear class
\begin{equation*}
\mathcal{F}=\{f_\theta(x)=\theta_0+\theta_1\eta(x)+\theta_2\bar{\eta}_{\mathrm{DP}}(x):\theta_0,\theta_1,\theta_2\in\mathbb{R}\}.
\end{equation*}

For $f_\theta\in \mathcal{F}$, the corresponding empirical $(L,\bar{L})$-risk is
\begin{equation*}
\hat{\mathcal{R}}_{\mathrm{fair}}^{L,\bar{L}}(f_\theta)
=\frac{1}{2n}\sum_{i=1}^{n}\log\!\left(1+e^{-y_i f_\theta(x_i)}\right)
+\frac{\lambda}{2n}\sum_{i=1}^{n}
\left(\frac{a_i+1}{2}-\pi_A\right)f_\theta(x_i)
+\rho\lVert\theta\rVert_2^{2}.
\end{equation*}
We solve it with L-BFGS using a ridge penalty $\rho=10^{-4}$.

\begin{figure}[ht]
\centering
\includegraphics[width=\textwidth]{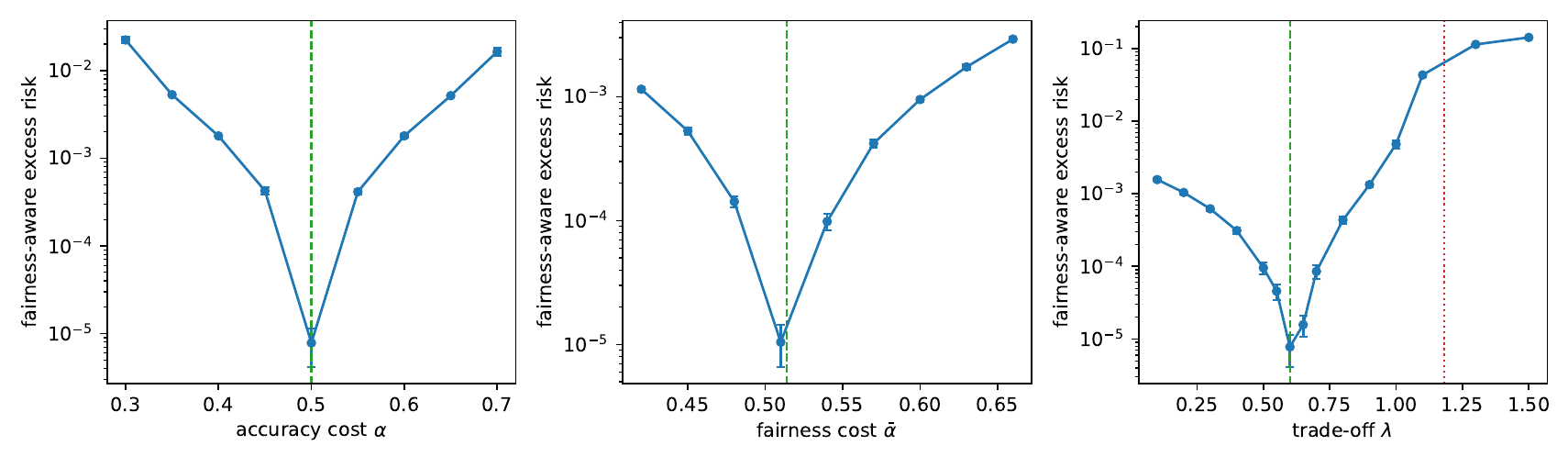}
\caption{Sensitivity sweep over hyperparameters under the DP target.}
\label{fig:sweeps-dp}
\end{figure}

Figure~\ref{fig:sweeps-dp} reports one‑at‑a‑time hyperparameter sweeps for the DP setting. We fix two parameters to the FAFC reference configuration $(\alpha,\bar{\alpha},\lambda)=(1/2,\pi_A,0.6)$ and vary the third: accuracy cost parameter $\alpha$ (left), fairness cost parameter $\bar{\alpha}$ (middle), and trade‑off parameter $\lambda$ (right). Fairness‑aware excess risk is evaluated at sample size $n=20000$. The vertical axis uses a log scale, with error bars representing $\pm 2$ standard errors over $20$ independent replications. Dashed lines mark the FAFC reference values, and the dotted line in the right panel marks the well‑posedness threshold $\lambda_{\max}=1.1808$.

Each curve attains its minimum precisely at the FAFC reference point, and excess risk grows monotonically as parameters deviate away from this optimum. This observation illustrates that satisfying all three FAFC conditions is necessary for low excess risk. For the $\lambda$ sweep, excess risk stays small below $\lambda_{\max}$ but rises steeply once the regularization strength exceeds this well‑posedness bound.

\subsubsection{Fairness-Aware Exponential Loss}

We simulate synthetic data under the following data‑generating process. Binary class labels $Y\in\{-1,1\}$ are sampled uniformly at random. The binary sensitive attribute $A\in\{-1,1\}$ is also drawn uniformly. Conditional on the sensitive‑attribute value $A=a$, the feature vector $X$ follows a zero‑mean Gaussian distribution with group‑dependent variance:
\begin{equation*}
X\mid A=a \sim N\!\left(0,\sigma_a^{2}\right).
\end{equation*}

We set group‑specific variances $\sigma_{1}=2$ and $\sigma_{-1}=1$. The difference in variances across groups creates inherent disparities in true‑positive rates, making this setup suitable for studying equal‑opportunity fairness.

The positive-class posterior is a logistic function of the feature,
\begin{equation*}
\mathbb{P}(Y=1\mid X=x,A=a)=\sigma(x),
\end{equation*}
and $Y\mid X,A\sim\text{Binomial}\big(1,\sigma(X)\big)$. The design has unbounded support because the exponential accuracy loss has exponential tails on both sides (Appendix~\ref{app:exponential}), so the conditional risk is bounded below for every $\lambda$ and no well-posedness threshold restricts the trade-off.

The classifier observes $(X,A)$, so the sensitive-attribute posterior collapses to $\bar{\eta}=\mathbf{1}\{a=1\}$. We set $\alpha=0.5$, $\bar{\alpha}=0.5$ and $\lambda=0.3$.

Let $(L,\bar{L})$ be the exponential surrogate loss pair defined in Eq.~\eqref{eq:exp-partials}. We minimize the empirical $(L,\bar{L})$-risk of Algorithm~\ref{alg:general-fair} for the EO criterion over the group-wise linear class
\begin{equation*}
\mathcal{F}=\bigl\{f_\theta(x,a)=\theta_{0,a}+\theta_{1,a}\eta(x,a):
\theta_{0,a},\theta_{1,a}\in\mathbb{R},\ a\in\{-1,1\}\bigr\}.
\end{equation*}
For $f_\theta\in\mathcal{F}$, the corresponding empirical $(L,\bar{L})$-risk is
\begin{equation*}
\hat{\mathcal{R}}_{\mathrm{fair}}^{L,\bar{L}}(f_\theta)
=\frac{1}{2n}\sum_{i=1}^{n}e^{-y_i f_\theta(x_i,a_i)}
+\frac{\lambda}{2\hat{\pi}n}\sum_{i:\,y_i=1}a_i\,f_\theta(x_i,a_i)
+\rho\lVert\theta\rVert_2^{2},
\end{equation*}
where $\hat{\pi}=\frac{1}{n}\sum_{i=1}^{n}\mathbf{1}\{y_i=1\}$. We solve it with L-BFGS using
a ridge penalty $\rho=10^{-4}$.

\begin{figure}[ht]
\centering
\includegraphics[width=\textwidth]{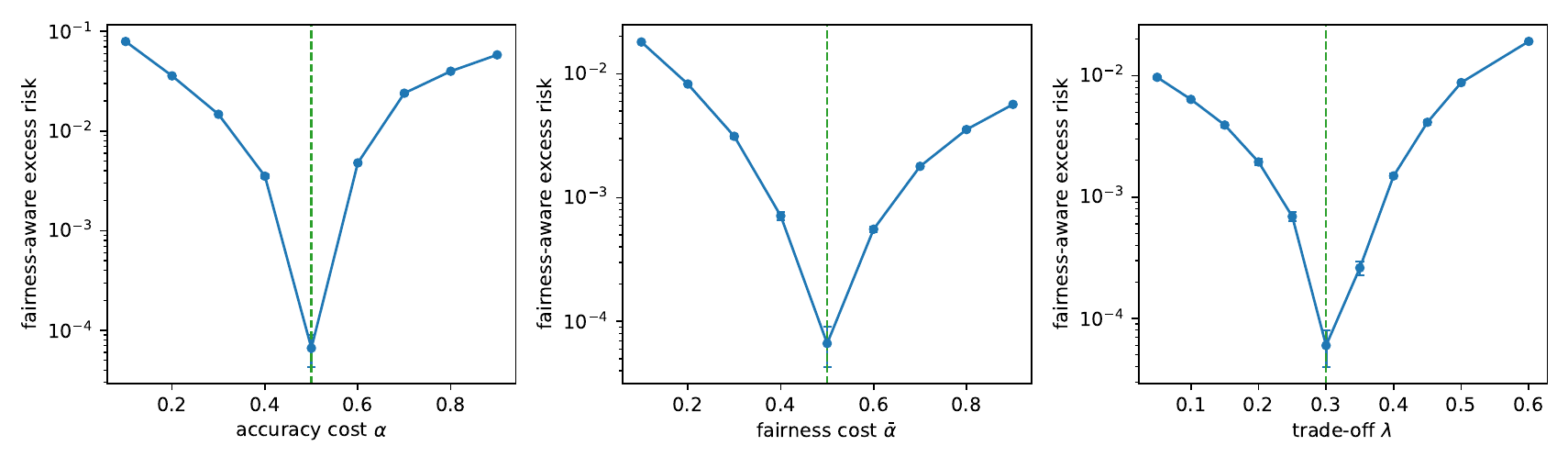}
\caption{Sensitivity sweep over hyperparameters under the EO target.}
\label{fig:sweeps-eo}
\end{figure}

Figure~\ref{fig:sweeps-eo} reports one-at-a-time hyperparameter sweeps for the EO setting, in the same format as Figure~\ref{fig:sweeps-dp}. Two parameters are fixed at the FAFC reference configuration $(\alpha,\bar{\alpha},\lambda)=(0.5,0.5,0.3)$ and the third is varied: accuracy cost $\alpha$ (left), fairness cost $\bar{\alpha}$ (middle), and trade-off $\lambda$ (right), with the fairness-aware excess risk evaluated at $n=20000$ (log vertical axis, $\pm2$ standard errors over $20$ replications). Each curve attains its minimum at the FAFC reference value and increases monotonically away from it. 

\subsection{A Real Case Study: Personal Credit Default Prediction}

Identifying borrowers at risk of credit default is important for retail lenders to manage portfolio risk. Misclassifying a high-risk applicant who will default incurs substantially larger financial losses than rejecting a creditworthy low-risk borrower, which naturally introduces asymmetric misclassification costs in credit scoring \citep{verbraken-2014-development,Elkan-2001-FoundationsCostSensitiveLearning}. 
Beyond cost imbalance, standard risk models trained solely for prediction accuracy often encode demographic disparities and yield unequal outcomes across protected groups. This makes consumer credit default prediction a setting where cost-sensitive classification and fairness regularization are both practically indispensable. Our proposed fairness-aware surrogate loss framework is particularly valuable here, as it unifies asymmetric error weighting and fairness regularization under a theoretically grounded classification procedure. 

The Give Me Some Credit (GMC) dataset\footnote{\url{https://www.kaggle.com/c/GiveMeSomeCredit}} encompasses 150,000 anonymized consumer credit records; the separate official competition test set has no available ground‑truth labels and is not used here. It includes features such as revolving utilization of unsecured credit lines, borrower age, counts of past-due episodes across 30–59, 60–89 and 90+ day delinquency tiers, debt-to-income ratio, monthly income, number of open credit accounts, number of real estate loans, and number of dependents. The objective is to predict whether a borrower will experience serious delinquency, defined as 90 or more days past due, within a two-year window. 

After preprocessing, we retain all 150,000 records. The binary label $Y=1$ indicates serious delinquency (90 or more days past due within two years) and $Y=-1$ otherwise, with $\mathbb{P}(Y=1)=0.06684$ empirically estimated. 
We select borrower age as the sensitive attribute $A$. Age is a formally protected ground for non-discrimination under the EU Charter of Fundamental Rights.\footnote{Charter of Fundamental Rights of the European Union, \url{https://eur-lex.europa.eu/legal-content/EN/TXT/?uri=CELEX:12012P/TXT}.} We encode age as a binary variable $A$: $A=-1$ for applicants younger than 25 years and $A=1$ otherwise, giving $\mathbb{P}(A=1)=0.9862$. The 25-year threshold is chosen for two reasons. First, it follows the standard binarization proposed by \citet{Kamiran-2009-ClassifyingWithoutDiscriminating} and widely adopted in the fair credit scoring literature. Second, this cutoff aligns with the United Nations official statistical definition of youth (aged 15–24), which corresponds to the demographic group that typically has limited credit history and faces systematically differential treatment in lending decisions. 

We adopt the EO criterion under the attribute-blind setting: age is only used to compute the fairness penalty during training and is excluded from the feature matrix at prediction time. We evaluate with five folds stratified jointly on the four $(Y,A)$ cells, training on $120{,}000$ records and testing on the remaining $30{,}000$ in each fold.

We first train a cost-sensitive SVM. In credit-scoring applications, failing to detect a delinquent borrower generates substantially greater financial losses than denying credit to a low-risk applicant. We set $\alpha=0.2$ to reflect this asymmetry: false negatives receive weight $1-\alpha=0.8$ and false positives weight $\alpha=0.2$, corresponding to a $4:1$ cost ratio between false negatives and false positives. The trained classifier achieves an overall false negative rate $\mathbb{P}(\hat{Y}=-1\mid Y=1)=0.6358$. For comparison, the standard SVM ($\alpha=1/2$) yields an even worse false negative rate of $1.0$. Under this setup, the asymmetric cost weighting is a necessary prerequisite to obtain a non-degenerate classifier. Hence all subsequent experiments use $\alpha=0.2$.

Cost-sensitive classification alone, however, fails to align classification performance across demographic groups. Without fairness regularization, $|\mathrm{MD}_{\mathrm{EO}}|=0.1273$. This observation motivates incorporating our fairness regularizer. 

\begin{figure}[htbp]
  \centering
  \begin{subfigure}{0.48\textwidth}
    \centering
    \includegraphics[width=\linewidth]{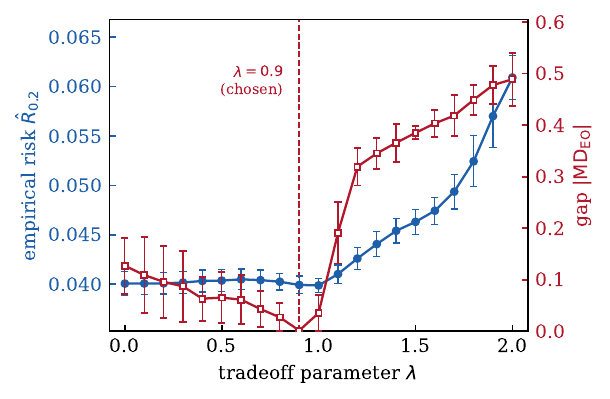}
    \caption{SVM (hinge loss)}
    \label{fig:gmc-tradeoff-svm}
  \end{subfigure}
  \hfill
  \begin{subfigure}{0.48\textwidth}
    \centering
    \includegraphics[width=\linewidth]{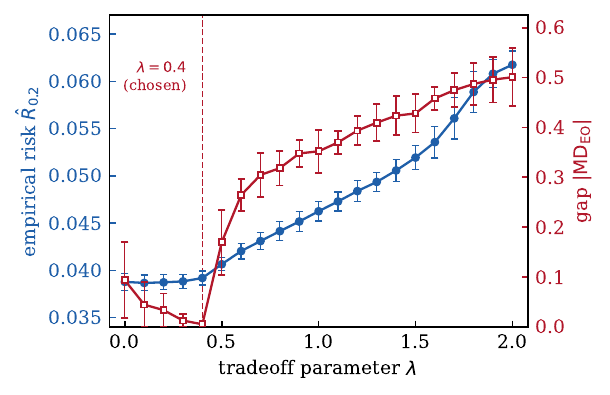}
    \caption{Logistic regression}
    \label{fig:gmc-tradeoff-lr}
  \end{subfigure}
  \caption{Cost-sensitive risk and equal-opportunity disparity across the $\lambda$ sweep for the GMC credit dataset.}
  \label{fig:gmc-tradeoff-overall}
\end{figure}

Figure~\ref{fig:gmc-tradeoff-svm} plots the empirical cost-sensitive risk $\hat{R}_{0.2}$ (blue, left axis) and equal-opportunity gap $\mathrm{MD}_{\mathrm{EO}}$ (red, right axis) as functions of $\lambda$, with $\alpha=0.2$ and $\bar{\alpha}=0.9784$ held fixed. Here $\bar{\alpha}=\hat{\mathbb{P}}(A=1\mid Y=1)=0.9784$ is estimated on the training folds of each split. All plotted values are aggregated from five-fold out-of-fold predictions, with error bars representing ±2 standard errors estimated across the cross-validation folds.

The absolute equal-opportunity gap $|\mathrm{MD}_{\mathrm{EO}}|$ exhibits a U-shaped trajectory as $\lambda$ increases: it starts at $0.127$ for the pure cost-sensitive SVM ($\lambda=0$), decreases to a minimum close to zero near $\lambda=1$, and then rises to $0.489$ at $\lambda=2$. The cost-sensitive risk remains nearly constant when $\lambda\in [0,1.1]$. When $\lambda>1.1$, however, both curves rise together, and should not be read as a continuation of the trade-off. The fairness term is linear in the score, so its pull grows in proportion to $\lambda$, whereas the accuracy surrogate term is a hinge loss: its subgradient is bounded and vanishes once the margin is met, so it can resist the penalty only up to a fixed level. Once the penalty exceeds that local restoring force, the minimizer is no longer pinned by the accuracy term and the fitted score drifts in the direction the penalty favors, which here is the direction that raises the scores of the group coded $A=-1$. Both the empirical risk and the disparity then increase with $\lambda$, and the pair leaves the range covered by \eqref{eq:svm-wellposed}.

At $\lambda=0.9$, the equal-opportunity gap is nearly eliminated: $|\mathrm{MD}_{\mathrm{EO}}|=0.0015$, compared with $0.1273$ for the cost-sensitive baseline. This is the optimal tradeoff parameter $\lambda$, where the cost-sensitive empirical risk $\hat{R}_{0.2}$ remains essentially unchanged, moving only from $0.0401$ to $0.0399$. The vertical dashed line in Figure~\ref{fig:gmc-tradeoff-svm} marks this point.

To verify that the above findings are not specific to the hinge-loss SVM, we repeat the analysis using a fairness-aware logistic regression classifier under the same equal-opportunity criterion, with $\alpha=0.2$ and $\bar{\alpha}=0.9784$ held fixed as in the SVM setup.

Figure~\ref{fig:gmc-tradeoff-lr} presents the resulting trade-off curves. The blue trace shows the empirical cost-sensitive risk $\hat{R}_{0.2}$ on the left axis, and the red trace shows the equal-opportunity disparity $\mathrm{MD}_{\mathrm{EO}}$ on the right axis. All points are computed from five-fold out-of-fold predictions, consistent with SVM.

The curve follows the same qualitative pattern observed for the SVM. For $\lambda\in[0,0.4]$, the empirical cost-sensitive risk remains nearly flat, while the EO gap narrows monotonically toward zero. When $\lambda>0.4$, however, the fairness penalty begins to dominate the objective, and the cost-sensitive risk rises noticeably with further increases in $\lambda$. The disparity crosses zero around $\lambda=0.4$. At this point we obtain $|\mathrm{MD}_{\mathrm{EO}}|=0.0051$, with the cost-sensitive risk still close to its baseline value. This confirms that equal-opportunity fairness can be achieved at negligible cost to the primary risk objective.

\section{Conclusion}

In this paper, we develop a statistical learning theory framework for surrogate-based fairness-aware classification. Within this framework, we introduce FAFC and the associated calibration inequality, covering both the DP and EO criteria under both the attribute-blind and attribute-aware settings.

First, we establish a necessary and sufficient condition for FAFC, stated purely in terms of the partial losses at the origin. Building on FAFC, we derive a calibration inequality showing that the excess fairness-aware risk is controlled by the excess $(L,\bar{L})$-risk through an explicit calibration function. We then instantiate the general framework with concrete surrogate loss pairs and provide a general algorithm for fairness-aware classification.

Turning to empirical validation, our simulations corroborate the theoretical predictions: the excess fairness-aware risk is minimized at the FAFC reference configuration, grows monotonically as the parameters deviate from it, and rises sharply once $\lambda$ exceeds the well-posedness threshold. A real-data study on the Give Me Some Credit credit default dataset further bears out the practical relevance of these results: the equal-opportunity disparity can be driven to nearly zero at negligible cost to the cost-sensitive risk.

Finally, we note a key limitation of the present theory: it presumes the fairness-regularized risk to be well posed. Beyond this threshold the conditional risk need not be bounded below, in which case no minimizer exists and neither the FAFC notion nor the calibration inequality applies.

\newpage
\ifnum\citationaudit=1
  \nocite{*}%
\fi
\bibliography{bibliography.bib}

\newpage
\section{Appendix}

\subsection{Choice of \texorpdfstring{$\bar{\alpha}$}{alpha}: Regularization Form of MD and MR}\label{app:fair-penalized}
The following lemma records the regularization form of the four standard fairness constraints and, in particular, the value of $\bar{\alpha}$ induced by each constraint. It is due to \citet{Menon-2018-CostFairnessBinary}.

\begin{lemma}[Regularization form of MD and MR \citep{Menon-2018-CostFairnessBinary}]\label{lem:fair-penalized}
For any distribution of $(X,A,Y)$ and any $\alpha,\tau\in(0,1)$, each of the four cases below specifies a constraint and a pair $(\bar{\alpha},\bar{D})$ such that the problem $\min_{f\in\mathcal{F}}\mathcal{R}_{\alpha}(f;D)$ subject to that constraint is equivalent, for a suitable $\lambda>0$, to the penalized problem
\begin{equation*}
\min_{f\in\mathcal{F}}\ \mathcal{R}_{\alpha}(f;D)-\lambda\,\mathcal{R}_{\bar{\alpha}}(f;\bar{D}).
\end{equation*}
\begin{enumerate}[label=(\roman*),leftmargin=*,itemsep=2pt,topsep=2pt]
\item (DP, MD): $\mathrm{MD}_{\mathrm{DP}}(f)\ge\tau$, with $\bar{\alpha}=\pi_A$ and $\bar{D}=\bar{D}_{\mathrm{DP}}$;
\item (DP, MR): $\mathrm{MR}_{\mathrm{DP}}(f)\ge\tau$, with $\bar{\alpha}=\pi_A/[\pi_A+(1-\pi_A)\tau]$ and $\bar{D}=\bar{D}_{\mathrm{DP}}$;
\item (EO, MD): $\mathrm{MD}_{\mathrm{EO}}(f)\ge\tau$, with $\bar{\alpha}=\pi_{A\mid Y=1}$ and $\bar{D}=\bar{D}_{\mathrm{EO}}$;
\item (EO, MR): $\mathrm{MR}_{\mathrm{EO}}(f)\ge\tau$, with $\bar{\alpha}=\pi_{A\mid Y=1}/[\pi_{A\mid Y=1}+(1-\pi_{A\mid Y=1})\tau]$ and $\bar{D}=\bar{D}_{\mathrm{EO}}$.
\end{enumerate}
\end{lemma}

\subsection{Proof of Corollary \ref{cor:FAFC}}\label{app:cor-14}
\begin{proof}
Let $f^{*}$ be a minimizer. Since $C_{\mathrm{fair}}^{L,\bar{L}}(\eta(x),\bar{\eta}(x),f^{*}(x))\ge C_{\mathrm{fair}}^{L,\bar{L},*}(\eta(x),\bar{\eta}(x))$ for every $x$, we have
\begin{equation*}
\mathbb{E}\left[C_{\mathrm{fair}}^{L,\bar{L}}(\eta(X),\bar{\eta}(X),f^{*}(X))\right]
=\mathcal{R}_{\mathrm{fair}}^{L,\bar{L}}(f^{*};D,\bar{D})
=\mathcal{R}_{\mathrm{fair}}^{L,\bar{L},*}
=\mathbb{E}\left[C_{\mathrm{fair}}^{L,\bar{L},*}(\eta(X),\bar{\eta}(X))\right],
\end{equation*}
so $C_{\mathrm{fair}}^{L,\bar{L}}(\eta(x),\bar{\eta}(x),f^{*}(x))=C_{\mathrm{fair}}^{L,\bar{L},*}(\eta(x),\bar{\eta}(x))$ for $P_X$-a.e. $x$.

Let $x$ be such that $(\eta(x),\bar{\eta}(x))\in\mathcal K$, $D^{\alpha,\bar{\alpha}}(\eta(x),\bar{\eta}(x))\neq0$, and the last equality holds. Suppose that $f^{*}(x)$ has the wrong sign, i.e., $f^{*}(x)D^{\alpha,\bar{\alpha}}(\eta(x),\bar{\eta}(x))\le0$. Then
\begin{align*}
C^{L,\bar{L}}_{\mathrm{fair}}(\eta(x),\bar{\eta}(x),f^{*}(x))
&\ge C^{L,\bar{L},-}_{\mathrm{fair}}(\eta(x),\bar{\eta}(x)) \\
&= C^{L,\bar{L},*}_{\mathrm{fair}}(\eta(x),\bar{\eta}(x))+H^{\alpha,\bar{\alpha}}_{L,\bar{L}}(\eta(x),\bar{\eta}(x)) \\
&> C^{L,\bar{L},*}_{\mathrm{fair}}(\eta(x),\bar{\eta}(x)),
\end{align*}
where the strict inequality uses FAFC on $\mathcal K$. This contradicts $C_{\mathrm{fair}}^{L,\bar{L}}(\eta(x),\bar{\eta}(x),f^{*}(x))=C_{\mathrm{fair}}^{L,\bar{L},*}(\eta(x),\bar{\eta}(x))$. Hence $f^{*}(x)D^{\alpha,\bar{\alpha}}(\eta(x),\bar{\eta}(x))>0$ for $P_X$-a.e. $x$ with $D^{\alpha,\bar{\alpha}}(\eta(x),\bar{\eta}(x))\neq0$.
\end{proof}

\subsection{Proof of Theorem \ref{thm:fair-fc-condition}}
\begin{proof}

\emph{Attribute-blind setting.}
For this setting, write
\[
\Phi(\eta,\bar{\eta})
=\left.\frac{\partial}{\partial t}C_{\mathrm{fair}}^{L,\bar L}(\eta,\bar{\eta},t)\right|_{t=0}.
\]
The quantities \(\Phi_{\mathrm{DP}}\) and \(\Phi_{\mathrm{EO}}\) below are the DP and EO versions of this derivative.

We use the following fact twice. Suppose that \(\Phi\) is continuous on \(\mathcal K\), that \(D^{\alpha,\bar{\alpha}}(\alpha,\bar{\alpha})=0\), and that \(\partial_\eta D^{\alpha,\bar{\alpha}}(\alpha,\bar{\alpha})\neq0\). Then \(\Phi=0\) on the part of the decision boundary \(\{D^{\alpha,\bar{\alpha}}=0\}\) near \((\alpha,\bar{\alpha})\). Indeed, near \((\alpha,\bar{\alpha})\) the boundary separates points with \(D^{\alpha,\bar{\alpha}}>0\) from points with \(D^{\alpha,\bar{\alpha}}<0\); at such points FAFC gives \(\Phi<0\) and \(\Phi>0\), respectively, and continuity gives \(\Phi=0\) on the boundary. In particular, \(\Phi(\alpha,\bar{\alpha})=0\).

\emph{DP.} Let 
\begin{align*}
\Phi_{\mathrm{DP}}(\eta,\bar{\eta}) &= \frac{ \partial  }{ \partial t }C_{\mathrm{DP}}^{L,\bar{L}}(\eta,\bar{\eta},t)\bigg|_{t=0} \\
& = \left[ \eta L'_{1}(0)+(1-\eta)L_{-1}'(0) \right] - \lambda \left[ \bar{\eta}\bar{L}'_{1}(0)+(1-\bar{\eta})\bar{L}_{-1}'(0) \right] \\
& =\bigr[L'_{-1}(0)-\lambda\bar{L}_{-1}'(0) \bigr]+\bigl[L'_{1}(0)-L'_{-1}(0)\bigr]\eta+\Bigl[\lambda \bar{L}_1'(0) -\lambda\bar{L}'_{-1}(0)\Bigr]\bar{\eta}.
\end{align*}

Since $\Phi_{\mathrm{DP}}(\eta,\bar{\eta})$ is affine to $(\eta,\bar{\eta})$, it is continuous on $\mathcal K$.

Since $t\mapsto C_{\mathrm{DP}}^{L,\bar{L}}(\eta,\bar{\eta},t)$ is strictly convex on $\mathcal K$ (with boundedness below guaranteed by the definition of $\mathcal K$), Lemma~\ref{lem:sign} shows that $L_{\mathrm{fair}}$ is $(\alpha,\bar{\alpha})$-FAFC on $\mathcal K$ if and only if
\begin{equation*}
\Phi_{\mathrm{DP}}(\eta, \bar{\eta}) 
\begin{cases} 
< 0 & \text{if } D^{\alpha,\bar{\alpha}}(\eta, \bar{\eta}) > 0, \\ 
> 0 & \text{if } D^{\alpha,\bar{\alpha}}(\eta, \bar{\eta}) < 0, \\
= 0 & \text{if } D^{\alpha,\bar{\alpha}}(\eta, \bar{\eta}) = 0.
\end{cases}
\end{equation*}

\textit{Necessity ($\implies$).} Assume $L_{\mathrm{fair}}$ is $(\alpha,\bar{\alpha})$-FAFC on $\mathcal K$ for this fixed $\lambda$.

By the continuity fact (applied to \(D_{\mathrm{DP}}^{\alpha,\bar{\alpha}}\), whose \(\eta\)-derivative is \(1\)), \(\Phi_{\mathrm{DP}}(\alpha,\bar{\alpha})=0\). 

Since 
\begin{equation*}
    \Phi_{\mathrm{DP}}(\alpha,\bar{\alpha}) = \left[ \alpha L'_{1}(0)+(1-\alpha)L_{-1}'(0) \right] - \lambda \left[ \bar{\alpha}\bar{L}'_{1}(0)+(1-\bar{\alpha})\bar{L}_{-1}'(0) \right] =0,
\end{equation*}
we obtain
\begin{equation*}
    \alpha L'_{1}(0) + (1-\alpha)L'_{-1}(0) = \lambda\left[ \bar{\alpha} \bar{L}'_{1}(0) + (1-\bar{\alpha})\bar{L}'_{-1}(0) \right].
\end{equation*}
This is exactly condition (D3).

By using condition (D3), $\Phi_{\mathrm{DP}}$ simplifies to 
\begin{align*} 
\Phi_{\mathrm{DP}}(\eta,\bar{\eta}) & = \left[ \eta L'_{1}(0)+(1-\eta)L_{-1}'(0) \right] -\lambda \left[ \bar{\eta}\bar{L}'_{1}(0)+(1-\bar{\eta})\bar{L}_{-1}'(0) \right] - \\ & \left( \alpha L'_{1}(0) + (1-\alpha)L'_{-1}(0) - \lambda\left[ \bar{\alpha} \bar{L}'_{1}(0) + (1-\bar{\alpha})\bar{L}'_{-1}(0) \right] \right)\\
& = (\eta-\alpha)(L'_{1}(0)-L'_{-1}(0)) - \lambda (\bar{\eta}-\bar{\alpha})(\bar{L}'_{1}(0)-\bar{L}'_{-1}(0)).
\end{align*}

Consider any point on the decision boundary where $D_{\mathrm{DP}}^{\alpha,\bar{\alpha}}(\eta,\bar{\eta})=0$, i.e., $\eta-\alpha=\lambda(\bar{\eta}-\bar{\alpha})$. By the continuity fact, $\Phi_{\mathrm{DP}}=0$ there as well, hence
\begin{equation*}
    \lambda (\bar{\eta}-\bar{\alpha}) \left[ (L_{1}'(0)-L_{-1}'(0)) - (\bar{L}_{1}'(0)-\bar{L}_{-1}'(0)) \right] = 0.
\end{equation*}
The decision boundary contains two distinct points with $\bar{\eta}\neq\bar{\alpha}$ (the line $\eta=\alpha+\lambda(\bar{\eta}-\bar{\alpha})$ lies in $\mathcal K$ for all $\bar{\eta}$ in a neighborhood of $\bar{\alpha}$), and the fixed value of $\lambda$ is nonzero, so $L_{1}'(0)-L_{-1}'(0)=\bar{L}'_{1}(0)-\bar{L}'_{-1}(0)$ must hold. This is exactly condition (D2).

By condition (D2), we continue simplifying $\Phi_{\mathrm{DP}}$:
\begin{equation*}
    \Phi_{\mathrm{DP}}(\eta,\bar{\eta}) = (L'_{1}(0)-L'_{-1}(0)) \cdot \left[ (\eta-\alpha)-\lambda(\bar{\eta}-\bar{\alpha}) \right] = (L'_{1}(0)-L'_{-1}(0)) \cdot D_{\mathrm{DP}}^{\alpha,\bar{\alpha}}(\eta,\bar{\eta}).
\end{equation*}
Since FAFC on $\mathcal K$ requires $\Phi_{\mathrm{DP}}$ and $D_{\mathrm{DP}}^{\alpha,\bar{\alpha}}$ to have strictly opposite signs, $L'_{1}(0)-L'_{-1}(0)$ must be strictly negative. Thus, $L'_{1}(0)-L'_{-1}(0)<0$. This is exactly condition (D1). 

\textit{Sufficiency ($\impliedby$).} Assume conditions (D1)--(D3) hold. 

For the fixed $\lambda$, using condition (D3), we have
\begin{equation*}
    \Phi_{\mathrm{DP}}(\eta,\bar{\eta}) = (\eta - \alpha)\left(L'_1(0) - L'_{-1}(0)\right) - \lambda(\bar{\eta} - \bar{\alpha})\left(\bar{L}'_1(0) - \bar{L}'_{-1}(0)\right).
\end{equation*}
Using condition (D2), we have
\begin{equation*}
    \Phi_{\mathrm{DP}}(\eta,\bar{\eta}) = (L_{1}'(0)-L_{-1}'(0)) \left[ (\eta-\alpha)-\lambda(\bar{\eta}-\bar{\alpha}) \right]
\end{equation*}
Since condition (D1) ensures $L_{1}'(0)-L_{-1}'(0)<0$, $\Phi_{\mathrm{DP}}(\eta,\bar{\eta})$ has precisely the opposite sign of $D_{\mathrm{DP}}^{\alpha,\bar{\alpha}}(\eta,\bar{\eta})$. Thus, $L_{\mathrm{fair}}$ is $(\alpha,\bar{\alpha})$-FAFC on $\mathcal K$.

\emph{EO.} Let $r:=\lambda/\pi$. Then
\begin{align*}
\Phi_{\mathrm{EO}}(\eta,\bar{\eta}) &= \frac{ \partial  }{ \partial t }C_{\mathrm{EO}}^{L,\bar{L}}(\eta,\bar{\eta},t)\bigg|_{t=0} \\
& = \left[ \eta L'_{1}(0)+(1-\eta)L_{-1}'(0) \right]
-r\eta \left[ \bar{\eta}\bar{L}'_{1}(0)+(1-\bar{\eta})\bar{L}_{-1}'(0) \right],
\end{align*}
and $D_{\mathrm{EO}}^{\alpha,\bar{\alpha}}(\eta,\bar{\eta})=(\eta-\alpha)-r\eta(\bar{\eta}-\bar{\alpha})$.

Since $t\mapsto C_{\mathrm{EO}}^{L,\bar{L}}(\eta,\bar{\eta},t)$ is strictly convex on $\mathcal K$ (with boundedness below guaranteed by the definition of $\mathcal K$), Lemma~\ref{lem:sign} shows that $L_{\mathrm{fair}}$ is $(\alpha,\bar{\alpha})$-FAFC on $\mathcal K$ if and only if
\begin{equation*}
\Phi_{\mathrm{EO}}(\eta, \bar{\eta}) 
\begin{cases} 
< 0 & \text{if } D^{\alpha,\bar{\alpha}}(\eta, \bar{\eta}) > 0, \\ 
> 0 & \text{if } D^{\alpha,\bar{\alpha}}(\eta, \bar{\eta}) < 0, \\
= 0 & \text{if } D^{\alpha,\bar{\alpha}}(\eta, \bar{\eta}) = 0.
\end{cases}
\end{equation*}

\textit{Necessity ($\implies$).} Assume $L_{\mathrm{fair}}$ is $(\alpha,\bar{\alpha})$-FAFC on $\mathcal K$ for this fixed $\lambda$.

By the continuity fact (applied to \(D_{\mathrm{EO}}^{\alpha,\bar{\alpha}}\), whose \(\eta\)-derivative at \((\alpha,\bar{\alpha})\) is \(1\)), \(\Phi_{\mathrm{EO}}(\alpha,\bar{\alpha})=0\). That is,
\[
\alpha L'_{1}(0)+(1-\alpha)L'_{-1}(0)
=r\alpha\left[ \bar{\alpha}\bar{L}'_{1}(0)+(1-\bar{\alpha})\bar{L}'_{-1}(0) \right],
\]
which is (E3).

For general $(\eta,\bar{\eta})$, subtracting $\Phi_{\mathrm{EO}}(\alpha,\bar{\alpha})$ gives
\begin{align*}
&\Phi_{\mathrm{EO}}(\eta,\bar{\eta})-\Phi_{\mathrm{EO}}(\alpha,\bar{\alpha})\\
&\quad=(\eta-\alpha)\left[L'_1(0)-L'_{-1}(0)\right]
-r(\eta-\alpha)\left[\bar{\alpha}\bar{L}'_1(0)+(1-\bar{\alpha})\bar{L}'_{-1}(0)\right]\\
&\qquad-r\eta(\bar{\eta}-\bar{\alpha})\left[\bar{L}'_1(0)-\bar{L}'_{-1}(0)\right].
\end{align*}

On the decision boundary $D_{\mathrm{EO}}^{\alpha,\bar{\alpha}}(\eta,\bar{\eta})=0$, i.e., $\eta-\alpha=r\eta(\bar{\eta}-\bar{\alpha})$, the continuity fact gives $\Phi_{\mathrm{EO}}=0$. Substituting this into the last display yields
\[
r\eta(\bar{\eta}-\bar{\alpha})\left[
L'_1(0)-L'_{-1}(0)
-r\left(\bar{\alpha}\bar{L}'_1(0)+(1-\bar{\alpha})\bar{L}'_{-1}(0)\right)
-\left(\bar{L}'_1(0)-\bar{L}'_{-1}(0)\right)\right]=0.
\]
The boundary contains points with $\bar{\eta}\neq\bar{\alpha}$ arbitrarily close to $(\alpha,\bar{\alpha})$ (the curve $\eta=\alpha/[1-r(\bar{\eta}-\bar{\alpha})]$ lies in $\mathcal K$ for all $\bar{\eta}$ in a neighborhood of $\bar{\alpha}$), and $r\eta>0$ there, so
\[
L'_1(0)-L'_{-1}(0)
=\bar{L}'_1(0)-\bar{L}'_{-1}(0)
+r\left[\bar{\alpha}\bar{L}'_1(0)+(1-\bar{\alpha})\bar{L}'_{-1}(0)\right].
\]
This is (E2).

Using (E2) and $\Phi_{\mathrm{EO}}(\alpha,\bar{\alpha})=0$, we obtain
\[
\Phi_{\mathrm{EO}}(\eta,\bar{\eta})
=\left[\bar{L}'_1(0)-\bar{L}'_{-1}(0)\right]
\left[(\eta-\alpha)-r\eta(\bar{\eta}-\bar{\alpha})\right]
=\left[\bar{L}'_1(0)-\bar{L}'_{-1}(0)\right]D_{\mathrm{EO}}^{\alpha,\bar{\alpha}}(\eta,\bar{\eta}).
\]
Since FAFC on $\mathcal K$ requires $\Phi_{\mathrm{EO}}$ and $D_{\mathrm{EO}}^{\alpha,\bar{\alpha}}$ to have strictly opposite signs,
\[
\bar{L}'_1(0)-\bar{L}'_{-1}(0)<0.
\]
This is (E1).

\textit{Sufficiency ($\impliedby$).} Assume (E1)--(E3) hold. By (E2) and (E3), the last display holds for every $(\eta,\bar{\eta})$. Since (E1) gives $\bar{L}'_1(0)-\bar{L}'_{-1}(0)<0$, $\Phi_{\mathrm{EO}}$ has precisely the opposite sign of $D_{\mathrm{EO}}^{\alpha,\bar{\alpha}}$, and Lemma~\ref{lem:sign} yields that $L_{\mathrm{fair}}$ is $(\alpha,\bar{\alpha})$-FAFC on $\mathcal K$.

\emph{Attribute-aware setting.}
The sufficiency direction is the same as in the attribute-blind case: it is pointwise and needs no additional assumption. Only the necessity direction requires the condition in Remark~\ref{rem:aware-fafc}. That condition is used below to ensure that each boundary point \(\eta_a\) has points of \(\mathcal K_a\) on both sides, so that FAFC and continuity give \(\Phi(\eta_a,a)=0\).
Assume this condition. Write
\[
\Phi(\eta,a)=\left.\frac{\partial}{\partial t}C_{\mathrm{fair}}^{L,\bar L}(\eta,a,t)\right|_{t=0}.
\]
On the edge \(a=1\), let \(\eta_+\) be the point where \(D(\eta,1)=0\); on \(a=-1\), let \(\eta_-\) be the corresponding point. Since \(D(\cdot,a)\) changes sign at these points, FAFC and continuity give
\[
\Phi(\eta_+,1)=0,\qquad \Phi(\eta_-,-1)=0.
\]

For DP, \(D(\eta,1)=0\) gives \(\eta_+=\alpha+\lambda(1-\bar\alpha)\), and \(D(\eta,-1)=0\) gives \(\eta_-=\alpha-\lambda\bar\alpha\). Hence
\[
0=\eta_+L_1'+(1-\eta_+)L_{-1}'-\lambda\bar L_1',
\]
\[
0=\eta_-L_1'+(1-\eta_-)L_{-1}'-\lambda\bar L_{-1}'.
\]
Subtracting yields \(L_1'-L_{-1}'=\bar L_1'-\bar L_{-1}'\), which is (D2). Substituting back yields (D3). After (D2) and (D3), \(\Phi\) is a constant multiple of \(D\); FAFC forces the constant to be negative, which is (D1).

For EO, \(D(\eta,a)=\eta-\alpha-r\eta(\mathbf 1\{a=1\}-\bar\alpha)\), so
\[
\eta_+-\alpha=r\eta_+(1-\bar\alpha),\qquad
\eta_--\alpha=-r\eta_-\bar\alpha.
\]
Equivalently,
\[
0=\eta_+L_1'+(1-\eta_+)L_{-1}'-r\eta_+\bar L_1',
\]
\[
0=\eta_-L_1'+(1-\eta_-)L_{-1}'-r\eta_-\bar L_{-1}'.
\]
Combining these two equations with the two relations for \(\eta_\pm\) first gives (E2); substituting (E2) back gives (E3). The sign of the resulting constant multiple of \(D\) gives (E1). This proves the edgewise analogue in the attribute-aware setting.
\end{proof}

\subsection{Proof of Theorem \ref{thm:fair-bound}}
\begin{proof}
By Lemma~\ref{lem:excess-risk},  the excess fairness-aware risk is given by
\begin{align*}
\mathcal{R}^{\alpha,\bar{\alpha}}_{\mathrm{fair}}(f) - \mathcal{R}_{\mathrm{fair}}^{\alpha,\bar{\alpha},*} &  = \mathbb{E}_{X}\left[ \mathbf{1}\left\{ \mathrm{sign}(f(X))\neq \mathrm{sign}(D^{\alpha,\bar{\alpha}}(X)) \right\} \cdot \left| D^{\alpha,\bar{\alpha}}(X) \right| \right] \\
& \le \mathbb{E}_{X}\left[ \mathbf{1}\left\{ f(X) \cdot D^{\alpha,\bar{\alpha}}(X) \le 0 \right\} \cdot \left| D^{\alpha,\bar{\alpha}}(X) \right| \right] .
\end{align*} 

The event $\{\mathrm{sign}(f(X))\neq\mathrm{sign}(D^{\alpha,\bar{\alpha}}(X))\}$ is contained in $\{f(X)D^{\alpha,\bar{\alpha}}(X)\le0\}$.

Since $\psi_{L,\bar{L}}^{\alpha, \bar{\alpha}}$ is convex, $\psi_{L,\bar{L}}^{\alpha, \bar{\alpha}}(0)=0$, and $\psi_{L,\bar{L}}^{\alpha, \bar{\alpha}}\ge0$ (Lemma~\ref{lem:props}), it is nondecreasing on $[0,\infty)$. We apply Jensen's inequality. 
\begin{align*}
\psi_{L,\bar{L}}^{\alpha, \bar{\alpha}} \left( \mathcal{R}^{\alpha,\bar{\alpha}}_{\mathrm{fair}}(f) - \mathcal{R}_{\mathrm{fair}}^{\alpha,\bar{\alpha},*} \right) 
& \le \psi_{L,\bar{L}}^{\alpha, \bar{\alpha}} \Bigl( \mathbb{E}_{X}\Bigl[ \mathbf{1}\left\{ f(X)D^{\alpha,\bar{\alpha}}(X)\le 0 \right\} |D^{\alpha,\bar{\alpha}}(X)| \Bigr] \Bigr) \\
& \le \mathbb{E}_{X}\Bigl[ \psi_{L,\bar{L}}^{\alpha, \bar{\alpha}} \Bigl( \mathbf{1}\left\{ f(X)D^{\alpha,\bar{\alpha}}(X)\le 0 \right\} |D^{\alpha,\bar{\alpha}}(X)| \Bigr) \Bigr] .
\end{align*}

Since $\psi_{L,\bar{L}}^{\alpha, \bar{\alpha}}(0)=0$, we have $\psi_{L,\bar{L}}^{\alpha, \bar{\alpha}}(I_{E} \cdot z) = I_{E} \cdot \psi_{L,\bar{L}}^{\alpha, \bar{\alpha}}(z)$ for any event $E$. Furthermore, by the definition of the biconjugate, $\psi_{L,\bar{L}}^{\alpha, \bar{\alpha}} \le \nu_{L,\bar{L}}^{\alpha, \bar{\alpha}}$. Thus, 
\begin{align*}
\mathbb{E}_{X}\bigl[ \mathbf{1}\{ f(X)D^{\alpha,\bar{\alpha}}(X)\le 0 \} \psi_{L,\bar{L}}^{\alpha, \bar{\alpha}} ( |D^{\alpha,\bar{\alpha}}(X)| ) \bigr] 
& \le \mathbb{E}_{X}\bigl[ \mathbf{1}\{ f(X)D^{\alpha,\bar{\alpha}}(X)\le 0 \} \nu_{L,\bar{L}}^{\alpha, \bar{\alpha}} (|D^{\alpha,\bar{\alpha}}(X)|) \bigr] .
\end{align*}

Recall $\nu_{L,\bar{L}}^{\alpha, \bar{\alpha}}(\epsilon) = \inf_{(\eta,\bar{\eta})\in\mathcal K} \{ H_{L,\bar{L}}^{\alpha,\bar{\alpha}}(\eta,\bar{\eta}) : |D^{\alpha,\bar{\alpha}}(\eta,\bar{\eta})| = \epsilon \}$. It follows that $\nu_{L,\bar{L}}^{\alpha, \bar{\alpha}} (|D^{\alpha, \bar{\alpha}}(X)|) \le H_{L,\bar{L}}^{\alpha,\bar{\alpha}}(\eta(X), \bar{\eta}(X))$. Substituting this back, we have
\begin{align*}
\mathrm{RHS} & \le \mathbb{E}_{X}\left[ \mathbf{1}\left\{ f(X)D(X)\le 0 \right\} H_{L,\bar{L}}^{\alpha,\bar{\alpha}}(\eta(X), \bar{\eta}(X)) \right] \\
& = \mathbb{E}_{X}\left[ \mathbf{1}\left\{ f(X)D(X)\le 0 \right\} \left( C_{\mathrm{fair}}^{L,\bar{L},-}(\eta(X), \bar{\eta}(X)) - C_{\mathrm{fair}}^{L,\bar{L},*}(\eta(X), \bar{\eta}(X)) \right) \right].
\end{align*}

By the definition of $C_{\mathrm{fair}}^{L,\bar{L},-}(\eta, \bar{\eta}) = \inf_{t: t D(\eta, \bar{\eta}) \le 0} C_{\mathrm{fair}}^{L,\bar{L}}(\eta, \bar{\eta}, t)$, if $f(X)$ satisfies the condition $f(X)D(X) \le 0$, then $C_{\mathrm{fair}}^{L,\bar{L}}(\eta(X), \bar{\eta}(X), f(X)) \ge C_{\mathrm{fair}}^{L,\bar{L},-}(\eta(X), \bar{\eta}(X))$. Therefore, 
\begin{equation*}
\begin{split}
    & \mathbf{1}\left\{ f(X)D^{\alpha, \bar{\alpha}}(X)\le 0 \right\} \left( C_{\mathrm{fair}}^{L,\bar{L},-}(\eta(X), \bar{\eta}(X)) - C_{\mathrm{fair}}^{L,\bar{L},*}(\eta(X), \bar{\eta}(X)) \right) \\
    \le &\ C_{\mathrm{fair}}^{L,\bar{L}}(\eta(X), \bar{\eta}(X), f(X)) - C_{\mathrm{fair}}^{L,\bar{L},*}(\eta(X), \bar{\eta}(X)).
\end{split}
\end{equation*}

Note that if the indicator is 0, the LHS is 0, and the RHS is non-negative by the definition of the optimal risk, so the inequality holds generally. Taking the expectation over $X$: 
\begin{align*}
& \mathbb{E}_{X}\left[ \mathbf{1}\left\{ f(X)D^{\alpha, \bar{\alpha}}(X)\le 0 \right\} H_{L,\bar{L}}^{\alpha,\bar{\alpha}}(\eta(X), \bar{\eta}(X)) \right] \\
& \le \mathbb{E}_{X} \left[ C_{\mathrm{fair}}^{L,\bar{L}}(\eta(X), \bar{\eta}(X), f(X)) - C_{\mathrm{fair}}^{L,\bar{L},*}(\eta(X), \bar{\eta}(X)) \right] \\
& = \mathcal{R}_{\mathrm{fair}}^{L,\bar{L}}(f) - \mathcal{R}_{\mathrm{fair}}^{L,\bar{L},*}.
\end{align*}

This completes the proof of the first part. 

Now consider the second part. Suppose first that $\psi_{L,\bar{L}}^{\alpha,\bar{\alpha}}$ is invertible on $[0,B_{\max}]$. By Part~1 of Lemma~\ref{lem:props}, $\psi_{L,\bar{L}}^{\alpha,\bar{\alpha}}(0)=0$. Since an invertible function has at most one zero and $\psi_{L,\bar{L}}^{\alpha,\bar{\alpha}}(\epsilon)\ge0$ on $[0,B_{\max}]$, it follows that
\[
\psi_{L,\bar{L}}^{\alpha,\bar{\alpha}}(\epsilon)>0
\qquad\text{for every }\epsilon\in(0,B_{\max}].
\]

Now let $(\eta,\bar{\eta})\in\mathcal K$ satisfy $D^{\alpha,\bar{\alpha}}(\eta,\bar{\eta})\neq0$. Then
$\bigl|D^{\alpha,\bar{\alpha}}(\eta,\bar{\eta})\bigr|\in(0,B_{\max}]$. By the definition of $\nu_{L,\bar{L}}^{\alpha,\bar{\alpha}}$,
\[
H_{L,\bar{L}}^{\alpha,\bar{\alpha}}(\eta,\bar{\eta}) \ge\nu_{L,\bar{L}}^{\alpha,\bar{\alpha}} \bigl(\bigl|D^{\alpha,\bar{\alpha}}(\eta,\bar{\eta})\bigr|\bigr) \ge\psi_{L,\bar{L}}^{\alpha,\bar{\alpha}} \bigl(\bigl|D^{\alpha,\bar{\alpha}}(\eta,\bar{\eta})\bigr|\bigr)>0 .
\]
Hence $L_{\mathrm{fair}}$ is $(\alpha,\bar{\alpha})$-FAFC on $\mathcal K$.

Conversely, assume that $L_{\mathrm{fair}}$ is $(\alpha,\bar{\alpha})$-FAFC on $\mathcal K$. Then $\psi_{L,\bar{L}}^{\alpha,\bar{\alpha}}(0)=0$ by Part~1 of Lemma~\ref{lem:props}, $\psi_{L,\bar{L}}^{\alpha,\bar{\alpha}}(\epsilon)>0$ for every $\epsilon\in(0,B_{\max}]$ by Part~2, and $\psi_{L,\bar{L}}^{\alpha,\bar{\alpha}}$ is continuous on $[0,B_{\max}]$ by Part~3. Let $0\le a<b\le B_{\max}$. Convexity together with $\psi_{L,\bar{L}}^{\alpha,\bar{\alpha}}(0)=0$ gives
\[
\psi_{L,\bar{L}}^{\alpha,\bar{\alpha}}(a) \le\frac{a}{b}\,\psi_{L,\bar{L}}^{\alpha,\bar{\alpha}}(b) +\Bigl(1-\frac{a}{b}\Bigr)\psi_{L,\bar{L}}^{\alpha,\bar{\alpha}}(0)
=\frac{a}{b}\,\psi_{L,\bar{L}}^{\alpha,\bar{\alpha}}(b).
\]

If $a=0$, then $\psi_{L,\bar{L}}^{\alpha,\bar{\alpha}}(0)=0
<\psi_{L,\bar{L}}^{\alpha,\bar{\alpha}}(b)$;
if $a>0$, then $\psi_{L,\bar{L}}^{\alpha,\bar{\alpha}}(a)>0$, 
so $\psi_{L,\bar{L}}^{\alpha,\bar{\alpha}}(b)
\ge\frac{b}{a}\,\psi_{L,\bar{L}}^{\alpha,\bar{\alpha}}(a)
>\psi_{L,\bar{L}}^{\alpha,\bar{\alpha}}(a)$. 
Hence $\psi_{L,\bar{L}}^{\alpha,\bar{\alpha}}$ is strictly increasing on $[0,B_{\max}]$. Since it is continuous as well, it is invertible on
$[0,B_{\max}]$.
\end{proof}

\subsection{Useful Lemmas}

\subsubsection{Unified Excess-Risk Lemma}

The following lemma characterizes the optimal decision boundary under the fairness-aware risk. It establishes that the excess risk is determined by the alignment between the prediction's sign and fairness-aware discriminant $D^{\alpha,\bar{\alpha}}(\eta,\bar{\eta})$.

\begin{lemma}\label{lem:excess-risk}
Let $\alpha, \bar{\alpha} \in (0,1)$. For both the attribute-blind and attribute-aware settings, the conditional excess risk satisfies
\begin{equation*}
\begin{split}   C_{\mathrm{fair}}^{\alpha,\bar{\alpha}}(\eta,\bar{\eta},t)-C_{\mathrm{fair}}^{\alpha,\bar{\alpha},*}(\eta,\bar{\eta})&=\mathbf{1}\left\{ \mathrm{sign}(t)\neq \mathrm{sign}(D^{\alpha,\bar{\alpha}}(\eta,\bar{\eta})) \right\} \cdot \left| D^{\alpha,\bar{\alpha}}(\eta,\bar{\eta}) \right| .
\end{split}
\end{equation*}
Consequently, for any $f\in \mathcal{F}$,
\begin{equation*}
\begin{split}   \mathcal{R}_{\mathrm{fair}}^{\alpha,\bar{\alpha}}(f)-\mathcal{R}_{\mathrm{fair}}^{\alpha,\bar{\alpha},*}
    &= \mathbb{E}_{X}\bigg[ \mathbf{1}\left\{ \mathrm{sign}(f(X))\neq \mathrm{sign}\big(D^{\alpha,\bar{\alpha}}(\eta(X),\bar{\eta}(X))\big) \right\} \cdot \left| D^{\alpha,\bar{\alpha}}(\eta(X),\bar{\eta}(X)) \right| \bigg].
\end{split}
\end{equation*}
\end{lemma}

\begin{proof}
Let $x \in \mathcal{X}$ be fixed. Recall that the fairness-aware conditional risk is given by
\begin{equation*}
C_{\mathrm{fair}}^{\alpha,\bar{\alpha}}(\eta,\bar{\eta},t) = A^{\alpha,\bar{\alpha}}(\eta,\bar{\eta})\mathbf{1}\left\{ t\leq 0 \right\} + B^{\alpha,\bar{\alpha}}(\eta,\bar{\eta}) \mathbf{1}\left\{ t>0 \right\}.
\end{equation*}

Using the identity $\mathbf{1}\left\{ t>0 \right\}=1-\mathbf{1}\left\{ t\leq 0 \right\}$, we can rewrite the conditional risk as
\begin{align*}
C_{\mathrm{fair}}^{\alpha,\bar{\alpha}}(\eta,\bar{\eta},t) &  = A^{\alpha,\bar{\alpha}}(\eta,\bar{\eta})\mathbf{1}\left\{ t\leq 0 \right\} + B^{\alpha,\bar{\alpha}}(\eta,\bar{\eta}) \left( 1-\mathbf{1}\left\{ t\leq 0 \right\} \right)  \\
& = B^{\alpha,\bar{\alpha}}(\eta,\bar{\eta}) + (A^{\alpha,\bar{\alpha}}(\eta,\bar{\eta})-B^{\alpha,\bar{\alpha}}(\eta,\bar{\eta})) \mathbf{1}\left\{ t\leq 0 \right\}.
\end{align*}

Recall the fairness-aware discriminant $D^{\alpha,\bar{\alpha}}(\eta,\bar{\eta})=A^{\alpha,\bar{\alpha}}(\eta,\bar{\eta})-B^{\alpha,\bar{\alpha}}(\eta,\bar{\eta})$. Since $B^{\alpha,\bar{\alpha}}$ is constant with respect to $t$, minimizing the risk is equivalent to minimizing the term $D^{\alpha,\bar{\alpha}}(\eta,\bar{\eta})\mathbf{1}\left\{ t\leq 0 \right\}$. We consider three cases based on the sign of $D^{\alpha,\bar{\alpha}}(\eta,\bar{\eta})$.

Case $D^{\alpha,\bar{\alpha}}(\eta,\bar{\eta})>0$. To minimize the risk, we must set $\mathbf{1}\left\{ t\leq 0 \right\}=0$, i.e., $t>0$. The optimal fairness-aware conditional risk is
\begin{equation*}
    C^{\alpha,\bar{\alpha},*}_{\mathrm{fair}}(\eta,\bar{\eta}) = B^{\alpha,\bar{\alpha}}(\eta,\bar{\eta}).
\end{equation*}

Since $D^{\alpha,\bar{\alpha}}(\eta,\bar{\eta})>0$, we have $D^{\alpha,\bar{\alpha}}(\eta,\bar{\eta})=\left| D^{\alpha,\bar{\alpha}}(\eta,\bar{\eta}) \right|$ and $\mathrm{sign}(D^{\alpha,\bar{\alpha}}(\eta,\bar{\eta}))=1$. Furthermore, $\mathbf{1}\left\{ t\leq 0 \right\}$ implies $\mathrm{sign}(t)=-1\neq \mathrm{sign}(D^{\alpha,\bar{\alpha}}(\eta,\bar{\eta}))$. Thus, we have
\begin{equation*}
    C_{\mathrm{fair}}^{\alpha,\bar{\alpha}}(\eta,\bar{\eta},t)-C_{\mathrm{fair}}^{\alpha,\bar{\alpha},*}(\eta,\bar{\eta})= \mathbf{1}\left\{  \mathrm{sign}(t)\neq \mathrm{sign}(D^{\alpha,\bar{\alpha}}(\eta,\bar{\eta}))\right\}\left| D^{\alpha,\bar{\alpha}}(\eta,\bar{\eta}) \right|  .
\end{equation*}

Case $D^{\alpha,\bar{\alpha}}(\eta,\bar{\eta})<0$. To minimize the risk, we must set $\mathbf{1}\left\{ t\leq 0 \right\}=1$, i.e., $t\le 0$. The optimal fairness-aware conditional risk is
\begin{equation*}
    C^{\alpha,\bar{\alpha},*}_{\mathrm{fair}}(\eta,\bar{\eta}) = A^{\alpha,\bar{\alpha}}(\eta,\bar{\eta}).
\end{equation*}

Therefore, the excess fairness-aware conditional risk for any $t$ is
\begin{equation*}
    C_{\mathrm{fair}}^{\alpha,\bar{\alpha}}(\eta,\bar{\eta},t)-C_{\mathrm{fair}}^{\alpha,\bar{\alpha},*}(\eta,\bar{\eta})= -D^{\alpha,\bar{\alpha}}(\eta,\bar{\eta}) \mathbf{1}\left\{ t> 0 \right\}.
\end{equation*}

Since $D^{\alpha,\bar{\alpha}}(\eta,\bar{\eta})<0$, we have $D^{\alpha,\bar{\alpha}}(\eta,\bar{\eta})=-\left| D^{\alpha,\bar{\alpha}}(\eta,\bar{\eta}) \right|$ and $\mathrm{sign}(D^{\alpha,\bar{\alpha}}(\eta,\bar{\eta}))=-1$. Furthermore, $\mathbf{1}\left\{ t>0 \right\}$ implies $\mathrm{sign}(t)=1\neq \mathrm{sign}(D^{\alpha,\bar{\alpha}}(\eta,\bar{\eta}))=-1$. Thus, we also have
\begin{equation*}
C_{\mathrm{fair}}^{\alpha,\bar{\alpha}}(\eta,\bar{\eta},t)-C_{\mathrm{fair}}^{\alpha,\bar{\alpha},*}(\eta,\bar{\eta})= \mathbf{1}\left\{  \mathrm{sign}(t)\neq \mathrm{sign}(D^{\alpha,\bar{\alpha}}(\eta,\bar{\eta}))\right\}\left| D^{\alpha,\bar{\alpha}}(\eta,\bar{\eta}) \right|  .
\end{equation*}

Case $D^{\alpha,\bar{\alpha}}(\eta,\bar{\eta})=0$. Since the risk is constant with respect to $t$, the excess risk is always 0.

Combining these cases, the conditional excess risk is
\begin{multline*}
C_{\mathrm{fair}}^{\alpha,\bar{\alpha}}(\eta,\bar{\eta},t)-C_{\mathrm{fair}}^{\alpha,\bar{\alpha},*}(\eta,\bar{\eta}) \\
= \mathbf{1}\left\{  \mathrm{sign}(t)\neq \mathrm{sign}(D^{\alpha,\bar{\alpha}}(\eta,\bar{\eta}))\right\}\left| D^{\alpha,\bar{\alpha}}(\eta,\bar{\eta}) \right|  \\
= \mathbf{1}\left\{ \mathrm{sign}(t)\neq \mathrm{sign}\left(A^{\alpha,\bar{\alpha}}(\eta,\bar{\eta})-B^{\alpha,\bar{\alpha}}(\eta,\bar{\eta})\right) \right\}\left| A^{\alpha,\bar{\alpha}}(\eta,\bar{\eta})-B^{\alpha,\bar{\alpha}}(\eta,\bar{\eta}) \right| .
\end{multline*}

Consequently,
\begin{align*}
& \mathcal{R}_{\mathrm{fair}}^{\alpha,\bar{\alpha}}(f)-\mathcal{R}_{\mathrm{fair}}^{\alpha,\bar{\alpha},*} \\
& = \mathbb{E}_{X}\left[ C_{\mathrm{fair}}^{\alpha,\bar{\alpha}}(\eta(X),\bar{\eta}(X),f(X))-C_{\mathrm{fair}}^{\alpha,\bar{\alpha},*}(\eta(X),\bar{\eta}(X)) \right]  \\
& =\mathbb{E}_{X}\bigg[
    \mathbf{1}\left\{ \mathrm{sign}(f(X))\neq \mathrm{sign}\left(A^{\alpha,\bar{\alpha}}(\eta(X),\bar{\eta}(X))-B^{\alpha,\bar{\alpha}}(\eta(X),\bar{\eta}(X))\right) \right\}  \\
& \quad \cdot \left| A^{\alpha,\bar{\alpha}}(\eta(X),\bar{\eta}(X))-B^{\alpha,\bar{\alpha}}(\eta(X),\bar{\eta}(X)) \right|
\bigg].
\end{align*}

\end{proof}

\subsubsection{Equivalent Conditions of FAFC}

\begin{lemma}\label{lem:sign}
Let $\alpha,\bar{\alpha}\in(0,1)$, let $\mathcal K$ be a well-posed domain as in Remark~\ref{rem:lambda}, and let $(L,\bar{L})$ be a surrogate loss pair such that,
for every $(\eta,\bar{\eta})\in\mathcal K$, the map $t\mapsto C_{\mathrm{fair}}^{L,\bar{L}}(\eta,\bar{\eta},t)$ is convex on $\mathbb{R}$ and differentiable at $0$. Let
\begin{equation*}
\Phi_{L,\bar{L}}(\eta,\bar{\eta}):=\partial_t C_{\mathrm{fair}}^{L,\bar{L}}(\eta,\bar{\eta},t)\big|_{t=0}.
\end{equation*}
Then the following three statements are equivalent:
\begin{enumerate}
    \item $L_{\mathrm{fair}}$ is $(\alpha,\bar{\alpha})$-FAFC on $\mathcal K$;
    \item $H_{L,\bar{L}}^{\alpha,\bar{\alpha}}(\eta,\bar{\eta})>0$ for every $(\eta,\bar{\eta})\in\mathcal K$ with $D^{\alpha,\bar{\alpha}}(\eta,\bar{\eta})\neq0$ (Definition~\ref{def:FAFC});
    \item $\Phi_{L,\bar{L}}(\eta,\bar{\eta})\cdot D^{\alpha,\bar{\alpha}}(\eta,\bar{\eta})<0$ for every $(\eta,\bar{\eta})\in\mathcal K$ with $D^{\alpha,\bar{\alpha}}(\eta,\bar{\eta})\neq0$.
\end{enumerate}
\end{lemma}

\begin{proof}
The equivalence of (i) and (ii) is Definition~\ref{def:FAFC}. Since $\mathcal K$ is a well-posed domain, the conditional risk is bounded below on $\mathcal K$.

For (ii)$\Leftrightarrow$(iii), fix $(\eta,\bar{\eta})$, write $C(t):=C_{\mathrm{fair}}^{L,\bar{L}}(\eta,\bar{\eta},t)$, and let $\Phi:=\Phi_{L,\bar{L}}(\eta,\bar{\eta})=C'(0)$. Convexity and differentiability at $0$ give
\begin{equation}\label{eq:lem-sign-bridge}
C(t)\ge C(0)+\Phi t\qquad\text{for every }t\in\mathbb{R}.
\end{equation}

Assume first that $D^{\alpha,\bar{\alpha}}(\eta,\bar{\eta})>0$, so the wrong-sign set is $t\le0$.

If $\Phi<0$, then \eqref{eq:lem-sign-bridge} gives $C(t)\ge C(0)$ for every $t\le0$, while $C(s)<C(0)$ for all sufficiently small $s>0$. Hence
\[
\inf_{t\le0}C(t)\ge C(0)>\inf_{t\in\mathbb{R}}C(t),
\]
so $H_{L,\bar{L}}^{\alpha,\bar{\alpha}}(\eta,\bar{\eta})>0$.

If $\Phi\ge0$, then $C(t)\ge C(0)$ for every $t>0$. Since $0$ belongs to the wrong-sign set, $\inf_{t\le0}C(t)\le C(0)\le\inf_{t>0}C(t)$, so the wrong-sign infimum equals the global infimum and $H_{L,\bar{L}}^{\alpha,\bar{\alpha}}(\eta,\bar{\eta})=0$.

Thus, for $D^{\alpha,\bar{\alpha}}(\eta,\bar{\eta})>0$, $H_{L,\bar{L}}^{\alpha,\bar{\alpha}}(\eta,\bar{\eta})>0$ if and only if $\Phi<0$, i.e., $\Phi D^{\alpha,\bar{\alpha}}(\eta,\bar{\eta})<0$. The case $D^{\alpha,\bar{\alpha}}(\eta,\bar{\eta})<0$ is symmetric, with the roles of $t\le0$ and $t\ge0$ interchanged. This proves (ii)$\Leftrightarrow$(iii).
\end{proof}

\subsubsection{Proof of Lemma \ref{lem:props}}

\begin{lemma}\label{lem:props}
Let $\alpha,\bar{\alpha}\in(0,1)$ and let $\mathcal K$ be as in Section~\ref{sec:calibration}. Assume the map $t\mapsto C_{\mathrm{fair}}^{L,\bar{L}}(\eta,\bar{\eta},t)$ is convex on $\mathbb{R}$ for every $(\eta,\bar{\eta})\in\mathcal K$. Then, 
\begin{enumerate}
  \item  $H_{L,\bar{L}}^{\alpha,\bar{\alpha}}(\eta,\bar{\eta})=0$ whenever $D^{\alpha,\bar{\alpha}}(\eta,\bar{\eta})=0$. In particular,
  \[
  H_{L,\bar{L}}^{\alpha,\bar{\alpha}}(\alpha,\bar{\alpha})
  =\nu_{L,\bar{L}}^{\alpha,\bar{\alpha}}(0)
  =\psi_{L,\bar{L}}^{\alpha,\bar{\alpha}}(0)=0 .
  \]

  \item If $L_{\mathrm{fair}}$ is $(\alpha,\bar{\alpha})$-FAFC on $\mathcal K$, then
  \[
  \psi_{L,\bar{L}}^{\alpha,\bar{\alpha}}(\epsilon)>0 \text{ for every }\epsilon\in(0,B_{\max}].
  \]

  \item If $L_{\mathrm{fair}}$ is $(\alpha,\bar{\alpha})$-FAFC on $\mathcal K$, then $\psi_{L,\bar{L}}^{\alpha,\bar{\alpha}}$ is continuous on $[0,B_{\max}]$.
\end{enumerate}
\end{lemma}

\begin{proof}
\emph{Part 1.} Let $(\eta,\bar{\eta})\in\mathcal K$ satisfy $D^{\alpha,\bar{\alpha}}(\eta,\bar{\eta})=0$. Then $t\,D^{\alpha,\bar{\alpha}}(\eta,\bar{\eta})=0\le0$ for every $t\in\mathbb{R}$, so the feasible set in the definition of $C_{\mathrm{fair}}^{L,\bar{L},-}(\eta,\bar{\eta})$ is all of $\mathbb{R}$ and
\[
C_{\mathrm{fair}}^{L,\bar{L},-}(\eta,\bar{\eta})
=\inf_{t\in\mathbb{R}}C_{\mathrm{fair}}^{L,\bar{L}}(\eta,\bar{\eta},t)
=C_{\mathrm{fair}}^{L,\bar{L},*}(\eta,\bar{\eta}).
\]

Hence $H_{L,\bar{L}}^{\alpha,\bar{\alpha}}(\eta,\bar{\eta})=0$. In particular, since $A^{\alpha,\bar{\alpha}}_{\mathrm{DP}}(\alpha,\bar{\alpha}) =\alpha(1-\alpha)-\lambda\bar{\alpha}(1-\bar{\alpha}) =B^{\alpha,\bar{\alpha}}_{\mathrm{DP}}(\alpha,\bar{\alpha})$ and $A^{\alpha,\bar{\alpha}}_{\mathrm{EO}}(\alpha,\bar{\alpha}) =\alpha(1-\alpha)-\lambda\pi^{-1}\alpha\bar{\alpha}(1-\bar{\alpha}) =B^{\alpha,\bar{\alpha}}_{\mathrm{EO}}(\alpha,\bar{\alpha})$, $D^{\alpha,\bar{\alpha}}(\alpha,\bar{\alpha})=0$ and therefore
\begin{equation*}
H_{L,\bar{L}}^{\alpha,\bar{\alpha}}(\alpha,\bar{\alpha})=0.
\end{equation*}

Hence the level set $\mathcal{L}_0=\{(\eta,\bar{\eta})\in\mathcal K:|D^{\alpha,\bar{\alpha}}(\eta,\bar{\eta})|=0\} =\{(\eta,\bar{\eta})\in\mathcal K:D^{\alpha,\bar{\alpha}}(\eta,\bar{\eta})=0\}$ is nonempty, since it contains $(\alpha,\bar{\alpha})$ and $H_{L,\bar{L}}^{\alpha,\bar{\alpha}}(\eta,\bar{\eta})=0$ on all of $\mathcal{L}_0$. Thus, 
\begin{equation*}
\nu_{L,\bar{L}}^{\alpha,\bar{\alpha}}(0)
=\inf_{(\eta,\bar{\eta})\in\mathcal{L}_0}H_{L,\bar{L}}^{\alpha,\bar{\alpha}}(\eta,\bar{\eta})=0.
\end{equation*}

By the definition of the biconjugate function, $\psi_{L,\bar{L}}^{\alpha,\bar{\alpha}}(\epsilon)\le\nu_{L,\bar{L}}^{\alpha,\bar{\alpha}}(\epsilon)$ and $\psi_{L,\bar{L}}^{\alpha,\bar{\alpha}}$ is the largest convex, lower semi-continuous function majorized by $\nu_{L,\bar{L}}^{\alpha,\bar{\alpha}}$. Hence $\psi_{L,\bar{L}}^{\alpha,\bar{\alpha}}(0)\le\nu_{L,\bar{L}}^{\alpha,\bar{\alpha}}(0)=0$.

Conversely, since $H_{L,\bar{L}}^{\alpha,\bar{\alpha}}(\eta,\bar{\eta})\ge0$, $\nu_{L,\bar{L}}^{\alpha,\bar{\alpha}}(\epsilon)\ge0$ on $[0,B_{\max}]$. The constant function $0$ is therefore convex, lower semi-continuous and majorized by $\nu_{L,\bar{L}}^{\alpha,\bar{\alpha}}$. As $\psi_{L,\bar{L}}^{\alpha,\bar{\alpha}}$ is the largest such function, $\psi_{L,\bar{L}}^{\alpha,\bar{\alpha}}(\epsilon)\ge0$ if $\epsilon\in[0,B_{\max}]$. Hence $\psi_{L,\bar{L}}^{\alpha,\bar{\alpha}}(0)=0$.

\emph{Part 2.}
By the explicit formulas for the example-dependent fair cost functions,
$D^{\alpha,\bar{\alpha}}$ is a polynomial in $(\eta,\bar{\eta})$; for instance, $D^{\alpha,\bar{\alpha}}_{\mathrm{DP}}(\eta,\bar{\eta})=(\eta-\alpha)-\lambda(\bar{\eta}-\bar{\alpha})$ and $D^{\alpha,\bar{\alpha}}_{\mathrm{EO}}(\eta,\bar{\eta}) =(\eta-\alpha)-\lambda\pi^{-1}\eta(\bar{\eta}-\bar{\alpha})$. 
Hence $D^{\alpha,\bar{\alpha}}$ is continuous on $\mathcal K$. Since $\mathcal K$ is compact and connected and $D^{\alpha,\bar{\alpha}}(\alpha,\bar{\alpha})=0$, the range of $|D^{\alpha,\bar{\alpha}}|$ is $[0,B_{\max}]$, so the level set
\[
\mathcal{L}_\epsilon
=\bigl\{(\eta,\bar{\eta})\in\mathcal K:
\bigl|D^{\alpha,\bar{\alpha}}(\eta,\bar{\eta})\bigr|=\epsilon\bigr\}
\]
is nonempty for every $\epsilon\in[0,B_{\max}]$. 

Fix $(\eta,\bar{\eta})\in\mathcal K$ with $D^{\alpha,\bar{\alpha}}(\eta,\bar{\eta})>0$ (the case $D^{\alpha,\bar{\alpha}}(\eta,\bar{\eta})<0$ is symmetric). Since $L_{\mathrm{fair}}$ is $(\alpha,\bar{\alpha})$-FAFC on $\mathcal K$, we have $H_{L,\bar{L}}^{\alpha,\bar{\alpha}}(\eta,\bar{\eta})>0$, that is,
\[
\inf_{t\le0}C_{\mathrm{fair}}^{L,\bar{L}}(\eta,\bar{\eta},t) >
\inf_{t\in\mathbb{R}}C_{\mathrm{fair}}^{L,\bar{L}}(\eta,\bar{\eta},t),
\]
so there exists $s\in\mathbb{R}^+$ with $C_{\mathrm{fair}}^{L,\bar{L}}(\eta,\bar{\eta},s) <\inf_{t\le0}C_{\mathrm{fair}}^{L,\bar{L}}(\eta,\bar{\eta},t)$. For every $t\le 0$ the point $0$ is a convex combination of $t$ and $s$, so convexity of
$u\mapsto C_{\mathrm{fair}}^{L,\bar{L}}(\eta,\bar{\eta},u)$ together with $C_{\mathrm{fair}}^{L,\bar{L}}(\eta,\bar{\eta},s) <C_{\mathrm{fair}}^{L,\bar{L}}(\eta,\bar{\eta},t)$ gives $C_{\mathrm{fair}}^{L,\bar{L}}(\eta,\bar{\eta},0) <C_{\mathrm{fair}}^{L,\bar{L}}(\eta,\bar{\eta},t)$. As $t=0$ itself is feasible, we know
\[
C_{\mathrm{fair}}^{L,\bar{L},-}(\eta,\bar{\eta})
=C_{\mathrm{fair}}^{L,\bar{L}}(\eta,\bar{\eta},0),
\]
and therefore
\begin{equation}\label{eq:H-C0-Cstar}
H_{L,\bar{L}}^{\alpha,\bar{\alpha}}(\eta,\bar{\eta})
=C_{\mathrm{fair}}^{L,\bar{L}}(\eta,\bar{\eta},0)
-C_{\mathrm{fair}}^{L,\bar{L},*}(\eta,\bar{\eta})
\end{equation}
on $\{(\eta,\bar{\eta}): D^{\alpha,\bar{\alpha}}(\eta,\bar{\eta})\neq0\}$. 

For every fixed $t\in\mathbb{R}$ the map $(\eta,\bar{\eta})\mapsto C_{\mathrm{fair}}^{L,\bar{L}}(\eta,\bar{\eta},t)$ is continuous on $\mathcal K$ since \eqref{eq:dp-conditional-Lrisk} and \eqref{eq:eo-conditional-Lrisk} are polynomials of degree at most two in $(\eta,\bar{\eta})$ whose coefficients $L_1(t),L_{-1}(t),\bar{L}_1(t),\bar{L}_{-1}(t)$ are real numbers.

Now we know $(\eta,\bar{\eta})\mapsto C_{\mathrm{fair}}^{L,\bar{L}}(\eta,\bar{\eta},0)$ is
continuous. And since $C_{\mathrm{fair}}^{L,\bar{L},*}(\eta,\bar{\eta})=\inf_{t\in\mathbb{R}}
C_{\mathrm{fair}}^{L,\bar{L}}(\eta,\bar{\eta},t)$ is an infimum of functions that are continuous in $(\eta,\bar{\eta})$, hence is upper semi-continuous. By \eqref{eq:H-C0-Cstar}, $H_{L,\bar{L}}^{\alpha,\bar{\alpha}}$ is a continuous function minus an upper semi-continuous function, hence lower semi-continuous on $\{(\eta,\bar{\eta}): D^{\alpha,\bar{\alpha}}(\eta,\bar{\eta})\neq0\}$.

Fix $\epsilon\in(0,B_{\max}]$. By the definition of $\mathcal{L}_{\epsilon}$, we have
$\mathcal{L}_\epsilon\subseteq\{(\eta,\bar{\eta}): D^{\alpha,\bar{\alpha}}(\eta,\bar{\eta})\neq0\}$. Since $H_{L,\bar{L}}^{\alpha,\bar{\alpha}}$ is lower semi-continuous on $\{(\eta,\bar{\eta}): D^{\alpha,\bar{\alpha}}(\eta,\bar{\eta})\neq0\}$, $H_{L,\bar{L}}^{\alpha,\bar{\alpha}}$ is also lower semi-continuous on $\mathcal{L}_\epsilon$.

Recall that $\nu_{L,\bar{L}}^{\alpha,\bar{\alpha}}(\epsilon)=\inf_{(\eta,\bar{\eta})\in\mathcal{L}_\epsilon} H_{L,\bar{L}}^{\alpha,\bar{\alpha}}(\eta,\bar{\eta})$, which is finite because $H_{L,\bar{L}}^{\alpha,\bar{\alpha}}$ is nonnegative on $\mathcal{L}_\epsilon$. Choose $(\eta_k,\bar{\eta}_k)\in\mathcal{L}_\epsilon$ with $H_{L,\bar{L}}^{\alpha,\bar{\alpha}}(\eta_k,\bar{\eta}_k)\to \nu_{L,\bar{L}}^{\alpha,\bar{\alpha}}(\epsilon)$ as $k\to\infty$. By compactness of $\mathcal{L}_\epsilon$ a subsequence converges to $(\eta^{*},\bar{\eta}^{*})\in\mathcal{L}_\epsilon$, and lower semi-continuity gives
\begin{equation}\label{eq:existence-nu-1}
H_{L,\bar{L}}^{\alpha,\bar{\alpha}}(\eta^{*},\bar{\eta}^{*})\le\liminf_{k\to\infty}H_{L,\bar{L}}^{\alpha,\bar{\alpha}}(\eta_k,\bar{\eta}_k)=\nu_{L,\bar{L}}^{\alpha,\bar{\alpha}}(\epsilon).
\end{equation}

Since $\nu_{L,\bar{L}}^{\alpha,\bar{\alpha}}(\epsilon)=\inf_{(\eta,\bar{\eta})\in\mathcal{L}_\epsilon} H_{L,\bar{L}}^{\alpha,\bar{\alpha}}(\eta,\bar{\eta})$, equality holds, i.e. 
\begin{equation}\label{eq:existence-nu-2}
\nu_{L,\bar{L}}^{\alpha,\bar{\alpha}}(\epsilon)=H_{L,\bar{L}}^{\alpha,\bar{\alpha}}(\eta^{*},\bar{\eta}^{*}).
\end{equation}

As $(\eta^{*},\bar{\eta}^{*})\in\mathcal{L}_\epsilon$, $D^{\alpha,\bar{\alpha}}(\eta^{*},\bar{\eta}^{*})\neq0$, so FAFC on $\mathcal K$ yields $H_{L,\bar{L}}^{\alpha,\bar{\alpha}}(\eta^{*},\bar{\eta}^{*})>0$. Therefore
\begin{equation}\label{eq:nu-ge-0}
\nu_{L,\bar{L}}^{\alpha,\bar{\alpha}}(\epsilon)=H_{L,\bar{L}}^{\alpha,\bar{\alpha}}(\eta^{*},\bar{\eta}^{*})>0 \text{ for all } \epsilon\in (0,B_{\max}].
\end{equation}

Next, we prove $\nu_{L,\bar{L}}^{\alpha,\bar{\alpha}}$ is lower semi-continuous. Let $\epsilon_n\to\epsilon$ in $[0,B_{\max}]$ and set $a:=\liminf_{n\to\infty}\nu_{L,\bar{L}}^{\alpha,\bar{\alpha}}(\epsilon_n)$. Choose a subsequence $n_j$ with $\nu_{L,\bar{L}}^{\alpha,\bar{\alpha}}(\epsilon_{n_j})\to a$. If $\epsilon>0$, then $\epsilon_{n_j}>0$ for all sufficiently large $j$; after discarding finitely many terms, choose minimizers $(\eta_{n_j},\bar{\eta}_{n_j})\in\mathcal{L}_{\epsilon_{n_j}}$ with $\nu_{L,\bar{L}}^{\alpha,\bar{\alpha}}(\epsilon_{n_j})=H_{L,\bar{L}}^{\alpha,\bar{\alpha}}(\eta_{n_j},\bar{\eta}_{n_j})$. Such minimizers exist by \eqref{eq:existence-nu-1} and \eqref{eq:existence-nu-2}.

Since $\mathcal K$ is compact, pass to a further subsequence $n_{j_i}$ such that $(\eta_{n_{j_i}},\bar{\eta}_{n_{j_i}})\to(\eta,\bar{\eta})\in\mathcal K$ as $i\to\infty$. By the continuity of $D^{\alpha,\bar{\alpha}}$,
\[
\bigl|D^{\alpha,\bar{\alpha}}(\eta,\bar{\eta})\bigr|=\lim_{i\to\infty}\bigl|D^{\alpha,\bar{\alpha}}(\eta_{n_{j_i}},\bar{\eta}_{n_{j_i}})\bigr|=\lim_{i\to\infty}\epsilon_{n_{j_i}}=\epsilon.
\]
Hence $(\eta,\bar{\eta})\in\mathcal{L}_\epsilon$. Since $\epsilon>0$, $D^{\alpha,\bar{\alpha}}(\eta,\bar{\eta})\neq0$, and lower semi-continuity of $H_{L,\bar{L}}^{\alpha,\bar{\alpha}}$ at $(\eta,\bar{\eta})$ yields
\[
a=\lim_{i\to\infty}\nu_{L,\bar{L}}^{\alpha,\bar{\alpha}}(\epsilon_{n_{j_i}})
=\lim_{i\to\infty}H_{L,\bar{L}}^{\alpha,\bar{\alpha}}(\eta_{n_{j_i}},\bar{\eta}_{n_{j_i}})
\ge H_{L,\bar{L}}^{\alpha,\bar{\alpha}}(\eta,\bar{\eta})
\ge\nu_{L,\bar{L}}^{\alpha,\bar{\alpha}}(\epsilon).
\]
If $\epsilon=0$, then $\nu_{L,\bar{L}}^{\alpha,\bar{\alpha}}(0)=0\le a$ because $\nu_{L,\bar{L}}^{\alpha,\bar{\alpha}}\ge0$. Hence $\liminf_{n\to\infty}\nu_{L,\bar{L}}^{\alpha,\bar{\alpha}}(\epsilon_n)\ge\nu_{L,\bar{L}}^{\alpha,\bar{\alpha}}(\epsilon)$ in all cases, so $\nu_{L,\bar{L}}^{\alpha,\bar{\alpha}}$ is lower semi-continuous on $[0,B_{\max}]$.

Since $\psi_{L,\bar{L}}^{\alpha,\bar{\alpha}}\ge0$ on $[0,B_{\max}]$, so it suffices to exclude $\psi_{L,\bar{L}}^{\alpha,\bar{\alpha}}(\epsilon_0)=0$ for some $\epsilon_0\in(0,B_{\max}]$. Suppose such an $\epsilon_0$ exists. By the definition of the biconjugate function, $\mathrm{Epi}\,\psi_{L,\bar{L}}^{\alpha,\bar{\alpha}} =\overline{\mathrm{co}\,\mathrm{Epi}\,\nu_{L,\bar{L}}^{\alpha,\bar{\alpha}}}$ and $(\epsilon_0,0)\in\mathrm{Epi}\,\psi_{L,\bar{L}}^{\alpha,\bar{\alpha}}$. Hence there exist $w_k\in\mathrm{co}\,\mathrm{Epi}\,
\nu_{L,\bar{L}}^{\alpha,\bar{\alpha}}$ with $w_k\to(\epsilon_0,0)$.

By Carath\'eodory's theorem in $\mathbb{R}^2$, for every $k$ there are weights
$\theta_1^{(k)},\theta_2^{(k)},\theta_3^{(k)}\ge0$ with
$\sum_{i=1}^3\theta_i^{(k)}=1$ and points
$(r_i^{(k)},s_i^{(k)})\in\mathrm{Epi}\,\nu_{L,\bar{L}}^{\alpha,\bar{\alpha}}$
such that $w_k=\sum_{i=1}^3\theta_i^{(k)}(r_i^{(k)},s_i^{(k)})$. Since
$\nu_{L,\bar{L}}^{\alpha,\bar{\alpha}}(\epsilon)=\infty$ outside
$[0,B_{\max}]$, every $r_i^{(k)}$ lies in $[0,B_{\max}]$. 

Since $\theta_i^{(k)}$ is in a standard simplex and $r_i^{(k)}\in[0,B_{\max}]$, by the compactness of the simplex and of $[0,B_{\max}]$, there exist subsequences $\theta_{i}^{(k_m)}$ and $r_{i}^{(k_m)}$ such that $\theta_{i}^{(k_m)}\to\theta_i$ and $r_i^{(k_m)}\to r_i$ for $i=1,2,3$ as $m\to\infty$. 

As the second coordinate $(w_k)_2$ of $w_k$ tends to $0$ as $k\to \infty$ and $s_i^{(k)}\ge\nu_{L,\bar{L}}^{\alpha,\bar{\alpha}}(r_i^{(k)})\ge0$, we have
\[
0\le \theta^{(k)}_i s^{(k)}_i \le (w_k)_2 \to 0, \quad k\to \infty.
\]

For each $i$, consider the limit weight $\theta_i$. If $\theta_i>0$, then $\theta_i^{(k_m)}\to\theta_i>0$ and $\theta_i^{(k_m)}s_i^{(k_m)}\to0$, hence $s_i^{(k_m)}\to0$. Lower semi-continuity of $\nu_{L,\bar{L}}^{\alpha,\bar{\alpha}}$ then gives
\[
0\le\nu_{L,\bar{L}}^{\alpha,\bar{\alpha}}(r_i)
\le\liminf_{m\to\infty}\nu_{L,\bar{L}}^{\alpha,\bar{\alpha}}(r_i^{(k_m)}) \le\liminf_{m\to\infty}s_i^{(k_m)}=0,
\]
so $\nu_{L,\bar{L}}^{\alpha,\bar{\alpha}}(r_i)=0$, and \eqref{eq:nu-ge-0} gives $r_i=0$. If $\theta_i=0$, then $\theta_i^{(k_m)}r_i^{(k_m)}\le\theta_i^{(k_m)}B_{\max}\to0$, so this index contributes $0$ to the limit of the first coordinate. Therefore,
\[
\epsilon_0=\lim_{m\to\infty}\sum_{i=1}^3\theta_i^{(k_m)}r_i^{(k_m)}=0,
\]
contradicting $\epsilon_0>0$. Hence $\psi_{L,\bar{L}}^{\alpha,\bar{\alpha}}(\epsilon)>0$ for every $\epsilon\in(0,B_{\max}]$.

\emph{Part 3.} 
Since $0\le\psi_{L,\bar{L}}^{\alpha,\bar{\alpha}}
\le\nu_{L,\bar{L}}^{\alpha,\bar{\alpha}}$ and $\nu_{L,\bar{L}}^{\alpha,\bar{\alpha}}$ is finite on $[0,B_{\max}]$ (proved in Part 2), $\psi_{L,\bar{L}}^{\alpha,\bar{\alpha}}$ is real-valued on $[0,B_{\max}]$.

Let $\epsilon\in(0,B_{\max})$. Since $\psi_{L,\bar{L}}^{\alpha,\bar{\alpha}}$ is convex and finite on a neighborhood of $\epsilon$, it is continuous at $\epsilon$ \citep[Theorem~10.1]{rockafellar-1970-convex}.

Let $\epsilon_n\to0^+$. By convexity of $\psi_{L,\bar{L}}^{\alpha,\bar{\alpha}}$ and $\psi_{L,\bar{L}}^{\alpha,\bar{\alpha}}(0)=0$,
\[
0\le\psi_{L,\bar{L}}^{\alpha,\bar{\alpha}}(\epsilon_n) \le\frac{\epsilon_n}{B_{\max}} \psi_{L,\bar{L}}^{\alpha,\bar{\alpha}}(B_{\max}) + \Bigl(1-\frac{\epsilon_n}{B_{\max}}\Bigr) \psi_{L,\bar{L}}^{\alpha,\bar{\alpha}}(0)
=\frac{\epsilon_n}{B_{\max}} \psi_{L,\bar{L}}^{\alpha,\bar{\alpha}}(B_{\max}) \longrightarrow0 ,
\]
where finiteness of $\psi_{L,\bar{L}}^{\alpha,\bar{\alpha}}(B_{\max})$ is used in the last step. Hence $\psi_{L,\bar{L}}^{\alpha,\bar{\alpha}}$ is continuous at $0$.

Let $\epsilon_n\to B_{\max}^-$. Convexity gives
\[
\psi_{L,\bar{L}}^{\alpha,\bar{\alpha}}(\epsilon_n) \le \frac{B_{\max}-\epsilon_n}{B_{\max}} \psi_{L,\bar{L}}^{\alpha,\bar{\alpha}}(0) +\frac{\epsilon_n}{B_{\max}} \psi_{L,\bar{L}}^{\alpha,\bar{\alpha}}(B_{\max}) \longrightarrow \psi_{L,\bar{L}}^{\alpha,\bar{\alpha}}(B_{\max}).
\]

Since $\psi_{L,\bar{L}}^{\alpha,\bar{\alpha}}$ is lower semi-continuous (proved in Part 2), the lower semi-continuity gives the reverse inequality in the limit. Hence $\psi_{L,\bar{L}}^{\alpha,\bar{\alpha}}$ is continuous at $B_{\max}$ as well.

Therefore $\psi_{L,\bar{L}}^{\alpha,\bar{\alpha}}$ is continuous on
$[0,B_{\max}]$.

\end{proof}

\subsection{Other Surrogate Loss Functions with FAFC}\label{app:other-examples}

The examples in this subsection, together with the logistic example in Section~\ref{subsec:logistic}, cover all four combinations of the fairness criterion (DP or EO) and the feature setting (attribute-blind or attribute-aware): Section~\ref{subsec:logistic} treats DP in the attribute-blind setting, Section~\ref{app:logistic-weighted} treats DP in the attribute-aware setting, Section~\ref{subsec:svm} treats EO in the attribute-blind setting, and Section~\ref{app:exponential} treats EO in the attribute-aware setting.

Throughout, we write $\Phi(\eta,\bar{\eta})=\partial_t C_{\mathrm{fair}}^{L,\bar{L}}(\eta,\bar{\eta},t)|_{t=0}$ for the derivative of the conditional risk at the origin. Unlike the logistic surrogate loss pair in Section~\ref{subsec:logistic}, which comes from a specific published method and therefore carries its own class and group costs, the three examples below are constructed for illustration and are not tied to a particular published method. We therefore treat cost parameters $\alpha,\bar{\alpha}$ as free parameters in $(0,1)$ and state the FAFC and well-posedness results for the whole weighted family. The general calibration inequality, however, does not lead to a simple bound when $\alpha$ and $\bar{\alpha}$ are kept general. To keep the calibration bounds explicit, we therefore compute them in the symmetric case $\alpha=\bar{\alpha}=\frac12$. 

\subsubsection{Fairness-Aware Weighted Logistic Loss}\label{app:logistic-weighted}

In this example, we consider the DP criterion in the attribute-aware setting. Recall from Section~\ref{subsec:fc-aware} that in this setting the classifier has access to $(X,A)$; for a fixed instance $(x,a)$, the positive-class posterior is $\eta=\eta(x,a)=\mathbb{P}(Y=1\mid X=x,A=a)$, and the sensitive-attribute posterior collapses to $\bar{\eta}=\mathbf{1}\{a=1\}$.

For the accuracy cost parameter $\alpha \in (0,1)$ and the fairness cost parameter $\bar{\alpha} \in (0,1)$, we define the partial loss functions of the accuracy surrogate loss $L$ and the fairness surrogate loss $\bar{L}$ as weighted logistic losses:
\begin{equation*}
L_1(t) = (1-\alpha)\log(1+e^{-t}), \quad L_{-1}(t) = \alpha\log(1+e^{t}),
\end{equation*}
\begin{equation*}
\bar{L}_1(t) = (1-\bar{\alpha})\log(1+e^{-t}), \quad \bar{L}_{-1}(t) = \bar{\alpha}\log(1+e^{t}).
\end{equation*}

This gives the compact form of the fairness-aware surrogate loss:
\begin{equation*}
L_{\mathrm{fair}}(y,a,t) = \left[ \frac{1+y(1-2\alpha)}{2} \right] \log(1+e^{ -yt }) - \lambda \left[ \frac{1+a(1-2\bar{\alpha})}{2} \right] \log(1+e^{ -at }).
\end{equation*}

The following proposition proves that the composite loss $L_{\mathrm{fair}}$ is FAFC.

\begin{proposition}\label{prop:weighted-logistic-fafc}
Let $\alpha,\bar{\alpha}\in(0,1)$, let $(L,\bar{L})$ be the weighted logistic surrogate pair defined above, let $\lambda_{\max}^{\mathrm{wlog}}$ be given in \eqref{eq:logistic-weight-wellposed}, and let $\mathcal K$ be the set of $(\eta,a)\in[0,1]\times\{-1,1\}$ on which $t\mapsto C_{\mathrm{DP}}^{L,\bar{L}}(\eta,a,t)$ is bounded below. If $0<\lambda<\lambda_{\max}^{\mathrm{wlog}}$, then $L_{\mathrm{fair}}$ is $(\alpha,\bar{\alpha})$-FAFC on $\mathcal K$ for the DP criterion in the attribute-aware setting.
\end{proposition}

\begin{proof}
Substituting the partial losses into the attribute-aware conditional risk gives
\begin{equation*}
C_{\mathrm{DP}}^{L,\bar{L}}(\eta,a,t)
 = \left[ \eta L_{1}(t)+(1-\eta)L_{-1}(t) \right]
 - \lambda \left[ \mathbf{1}\{a=1\}\bar{L}_{1}(t) + (1-\mathbf{1}\{a=1\})\bar{L}_{-1}(t) \right].
\end{equation*}

By the sufficiency direction of Theorem~\ref{thm:fair-fc-condition} it suffices to check (D1)--(D3) and strict convexity.
The partial losses have derivatives at the origin
\begin{equation*}
L'_1(0) = -\frac{1}{2}(1-\alpha), \quad L'_{-1}(0) = \frac{1}{2}\alpha, \qquad
\bar{L}'_1(0) = -\frac{1}{2}(1-\bar{\alpha}), \quad \bar{L}'_{-1}(0) = \frac{1}{2}\bar{\alpha}.
\end{equation*}

(D1) holds because $L'_1(0)-L'_{-1}(0)=-\frac12<0$. (D2) holds because both derivative gaps equal $-\frac12$. (D3) holds because
\begin{equation*}
(1-\alpha)L'_{-1}(0)+\alpha L'_1(0)=\frac12\alpha(1-\alpha)-\frac12\alpha(1-\alpha)=0,
\end{equation*}
and identically for $\bar{L}$ with $\bar{\alpha}$ in place of $\alpha$. Hence
\[
(1-\alpha)L'_{-1}(0)+\alpha L'_1(0) = \lambda [(1-\bar{\alpha})\bar{L}'_{-1}(0)+\bar{\alpha} \bar{L}'_1(0)].
\]

For the strict convexity, write $\bar{\eta}=\mathbf{1}\{a=1\}$ and collect the two logarithmic terms of the conditional risk:
\begin{equation*}
C_{\mathrm{DP}}^{L,\bar{L}}(\eta,a,t) = c_1\log(1+e^{-t}) + c_2\log(1+e^{t}),
\end{equation*}
where $c_1 := \eta(1-\alpha)-\lambda\bar{\eta}(1-\bar{\alpha}), c_2 := (1-\eta)\alpha-\lambda(1-\bar{\eta})\bar{\alpha}$. 

Its second derivative is $(c_1+c_2)e^{t}/(1+e^{t})^{2}$, which is positive whenever $c_1+c_2>0$. On $\mathcal K$, $c_1,c_2\ge0$ by definition. If $c_1+c_2=0$ at some point of $\mathcal K$, then $c_1=c_2=0$, which would force one of the two ratios in \eqref{eq:logistic-weight-wellposed} to equal $\lambda$, contradicting $0<\lambda<\lambda_{\max}^{\mathrm{wlog}}$. Hence $c_1+c_2>0$ on $\mathcal K$, and the conditional risk is strictly convex there. 
\end{proof}

The derivative conditions and strict convexity on $\mathcal K$ are therefore verified; the well-posedness is checked in the next paragraph, and the sufficiency direction of Theorem~\ref{thm:fair-fc-condition} yields that $L_{\mathrm{fair}}$ is $(\alpha,\bar{\alpha})$-FAFC on $\mathcal K$ for the DP criterion in the attribute-aware setting.

It remains to verify well-posedness. Since $\log(1+e^{-t})=o(t)$ and $\log(1+e^{t})=t+o(t)$ as $t\to+\infty$, while $\log(1+e^{-t})=-t+o(t)$ and $\log(1+e^{t})=o(t)$ as $t\to-\infty$, its tails are
\begin{equation*}
C_{\mathrm{DP}}^{L,\bar{L}}(\eta,a,t)=
\begin{cases}
c_2\,t+o(t), & t\to+\infty,\\[2pt]
-c_1\,t+o(t), & t\to-\infty,
\end{cases}
\end{equation*}
so the risk is bounded below exactly when $c_1\ge0$ and $c_2\ge0$. Over the support, the two strict inequalities read $0<\lambda<\lambda_{\max}^{\mathrm{wlog}}$, where
\begin{equation}\label{eq:logistic-weight-wellposed}
\lambda_{\max}^{\mathrm{wlog}}:=\min\Bigg\{
\inf_{\substack{x\in\mathrm{supp}(X):\\ a=1}}\frac{\eta(x,1)\,(1-\alpha)}{1-\bar{\alpha}},
\inf_{\substack{x\in\mathrm{supp}(X):\\ a=-1}}\frac{(1-\eta(x,-1))\,\alpha}{\bar{\alpha}}
\Bigg\}.
\end{equation}
By definition of $\mathcal K$, $c_1,c_2\ge0$ there, so the risk is bounded below on $\mathcal K$. Under \eqref{eq:logistic-weight-wellposed}, the sufficiency direction of Theorem~\ref{thm:fair-fc-condition} applies and $L_{\mathrm{fair}}$ is $(\alpha,\bar{\alpha})$-FAFC on $\mathcal K$ for the DP criterion in the attribute-aware setting.

Having verified FAFC, we now derive the calibration function and the excess-risk bound.
We consider the symmetric member of the weighted logistic surrogate loss family, in which
\[
L_1(t)=\frac12\log(1+e^{-t}),\qquad L_{-1}(t)=\frac12\log(1+e^{t}),
\]
\[
\bar{L}_1(t)=\frac12\log(1+e^{-t}),\qquad \bar{L}_{-1}(t)=\frac12\log(1+e^{t}).
\]

Up to the common positive factor $\frac12$, this is the standard unweighted logistic loss for both the accuracy and fairness terms. 

\begin{proposition}\label{prop:weighted-logistic-calibration}
For the symmetric member of the weighted logistic family ($\alpha=\bar{\alpha}=\frac12$), suppose that $0<\lambda<\lambda_{\max}^{\mathrm{wlog}}$, with $\lambda_{\max}^{\mathrm{wlog}}$ given in \eqref{eq:logistic-weight-wellposed}, and let $\mathcal K$ be the well-posed region of Proposition~\ref{prop:weighted-logistic-fafc}. Then, on $\mathcal K$, the calibration function of the pair $(L,\bar{L})$ for the DP criterion in the attribute-aware setting satisfies
\begin{equation}\label{eq:app-logistic-psi}
\psi_{L,\bar{L}}^{\frac12,\frac12}(\epsilon)\;\ge\;\frac{\epsilon^{2}}{1-\lambda}
\end{equation}
and, consequently, for every $f\in\mathcal{F}$,
\begin{equation}\label{eq:app-logistic-bound}
\mathcal{R}_{\mathrm{fair}}^{\frac12,\frac12}(f)-\mathcal{R}_{\mathrm{fair}}^{\frac12,\frac12,*}\le\min\Bigg\{\frac{1-\lambda}{2},\;\sqrt{(1-\lambda)\Big(\mathcal{R}_{\mathrm{fair}}^{L,\bar{L}}(f)-\mathcal{R}_{\mathrm{fair}}^{L,\bar{L},*}\Big)}\Bigg\}.
\end{equation}
\end{proposition}

\begin{proof}
Write $w_1:=\eta-\lambda\mathbf{1}\{a=1\}, w_2:=(1-\eta)-\lambda\mathbf{1}\{a=-1\}$. The conditional risk is
\begin{equation*}
C_{\mathrm{DP}}^{L,\bar{L}}(\eta,a,t) = \frac12 w_1\log(1+e^{-t})+\frac12 w_2\log(1+e^{t}),
\end{equation*}

Its derivative with respect to $t$ is $\frac12[(w_1+w_2)\sigma(t)-w_1]$, where $\sigma(t)=1/(1+e^{-t})$. Under \eqref{eq:logistic-weight-wellposed}, $w_1,w_2>0$ on the support, so the conditional risk is strictly convex and its unique minimizer is characterized by
$\sigma(t^{*})=w_1/(w_1+w_2)$. Let
\begin{equation*}
\tilde S := \sigma(t^*)= \frac{\eta-\lambda\mathbf{1}\{a=1\}}{1-\lambda}.
\end{equation*}

Since $w_1+w_2=1-\lambda$,
\begin{equation*}
C_{\mathrm{DP}}^{L,\bar{L},*}(\eta,a) = -\frac12\Big[w_1\log\tilde S+w_2\log(1-\tilde S)\Big],
\end{equation*}

To compute $C_{\mathrm{DP}}^{L,\bar{L},-}(\eta,a)$, we first compute the discriminant $D^{\frac12,\frac12}(\eta,a)$. By \eqref{eq:dp-aware-ab},
\begin{equation*}
D^{\frac12,\frac12}(\eta,a)=A^{\frac12,\frac12}_{\mathrm{DP}}(\eta,a)-B^{\frac12,\frac12}_{\mathrm{DP}}(\eta,a)
=\eta-\frac12-\lambda\mathbf{1}\{a=1\}+\frac{\lambda}{2}.
\end{equation*}

Since $\tilde S= \frac{\eta-\lambda\mathbf{1}\{a=1\}}{1-\lambda}$, we have
\[
D^{\frac12,\frac12}(\eta,a)=(1-\lambda)\Big(\tilde S-\tfrac12\Big),
\]
which means the unique minimizer lies on the correct side of the origin. Hence
\begin{equation*}
C_{\mathrm{DP}}^{L,\bar{L},-}(\eta,a) = C_{\mathrm{DP}}^{L,\bar{L}}(\eta,a,0) =  \frac{1-\lambda}{2}\log 2.
\end{equation*}

Therefore, the excess conditional risk is
\begin{equation*}
H_{L,\bar{L}}^{\frac12,\frac12}(\eta,a)
=\frac{1-\lambda}{2}\Big[\log 2+\tilde S\log\tilde S+(1-\tilde S)\log(1-\tilde S)\Big]
=\frac{1-\lambda}{2}D_{\mathrm{KL}}\Big(\tilde S\,\Big\|\,\frac12\Big).
\end{equation*}

Since $D^{\frac12,\frac12}(\eta,a)=(1-\lambda)(\tilde S-\frac12)$, on a discriminant level set $\lvert D^{\frac12,\frac12}(\eta,a)\rvert=\epsilon$ we have $\tilde S=\frac12\pm\epsilon/(1-\lambda)$. The two signs give the same value because $\tilde S\mapsto D_{\mathrm{KL}}(\tilde S\|\frac12)$ is symmetric about $\frac12$, so the level-set minimum is
\begin{equation*}
\nu_{L,\bar{L}}^{\frac12,\frac12}(\epsilon) =\frac{1-\lambda}{2}D_{\mathrm{KL}}\Big(\frac12+\frac{\epsilon}{1-\lambda}\,\Big\|\,\frac12\Big).
\end{equation*}

Set $g(u):=D_{\mathrm{KL}}(\frac12+u\|\frac12)$. $g$ is convex and increasing, so $\nu_{L,\bar{L}}^{\frac12,\frac12}(\epsilon)=\frac{1-\lambda}{2}g(\epsilon/(1-\lambda))$ is convex and increasing as well.  Hence $\psi_{L,\bar{L}}^{\frac12,\frac12}=\nu_{L,\bar{L}}^{\frac12,\frac12}$. 
Finally, by Pinsker's inequality $D_{\mathrm{KL}}(p\|q)\ge2(p-q)^{2}$,
\begin{equation*}
\psi_{L,\bar{L}}^{\frac12,\frac12}(\epsilon)
= \frac{1-\lambda}{2}\,g\!\Big(\frac{\epsilon}{1-\lambda}\Big)
\ge\frac{\epsilon^{2}}{1-\lambda},
\end{equation*}
which is \eqref{eq:app-logistic-psi}. 

By Theorem~\ref{thm:fair-bound} and Remark~\ref{rem:ext}, \eqref{eq:app-logistic-psi} gives $\mathcal{R}_{\mathrm{fair}}^{\frac12,\frac12}(f)-\mathcal{R}_{\mathrm{fair}}^{\frac12,\frac12,*}\le\sqrt{(1-\lambda)(\mathcal{R}_{\mathrm{fair}}^{L,\bar{L}}(f)-\mathcal{R}_{\mathrm{fair}}^{L,\bar{L},*})}$.

In addition, since $\tilde S\in(0,1)$, we have $|D^{\frac12,\frac12}(\eta,a)|< (1-\lambda)/2$. By Lemma~\ref{lem:excess-risk}, the trivial
bound $\epsilon\le\frac{1-\lambda}{2}$ holds. 

Taking the smaller of the two bounds gives \eqref{eq:app-logistic-bound}.
\end{proof}

\subsubsection{Fairness-Aware Hinge Loss}\label{subsec:svm}

The Support Vector Machine (SVM) is one of the most classic classifiers in statistical learning, with the hinge loss as its standard loss function. In this example we consider the EO criterion in the attribute-blind setting, where $\bar{\eta}=\bar{\eta}_{\mathrm{EO}}(x)$ is the sensitive-attribute posterior among the positive class. 

Let the accuracy cost parameter be $\alpha\in(0,1)$, the fairness cost parameter be $\bar{\alpha}\in(0,1)$, and define the partial losses
\begin{equation}\label{eq:hinge-partials}
L_1(t)=(1-\alpha)(1-t)_+, L_{-1}(t)=\alpha(1+t)_+, \quad
\bar{L}_1(t)=-(1-\bar{\alpha})t, \bar{L}_{-1}(t)=\bar{\alpha}t,
\end{equation}
where $(z)_+=\max\{z,0\}$: the accuracy surrogate is the cost-sensitive hinge of the classical theory, and the fairness surrogate is linear. 

The following proposition proves that the composite loss $L_{\mathrm{fair}}$ is FAFC.

\begin{proposition}\label{prop:svm-fafc}
Let $\alpha,\bar{\alpha}\in(0,1)$, let $(L,\bar{L})$ be the surrogate loss pair in \eqref{eq:hinge-partials}, let $\lambda_{\max}^{\mathrm{svm}}$ be given in \eqref{eq:svm-wellposed}, and let $\mathcal K$ be the set of $(\eta,\bar{\eta})\in[0,1]^2$ on which $t\mapsto C_{\mathrm{fair}}^{L,\bar{L}}(\eta,\bar{\eta},t)$ is bounded below. If $0<\lambda<\lambda_{\max}^{\mathrm{svm}}$, then $L_{\mathrm{fair}}$ is $(\alpha,\bar{\alpha})$-FAFC on $\mathcal K$ for the EO criterion in the attribute-blind setting.
\end{proposition}

\begin{proof}
The hinge is piecewise linear and hence not strictly convex, so the strict-convexity premise of Theorem~\ref{thm:fair-fc-condition} is not met and $L_{\mathrm{fair}}$ cannot be certified through that theorem. We therefore verify FAFC directly through Definition~\ref{def:FAFC}, which requires no convexity.

The attribute-blind EO conditional risk is
\begin{equation*}
C_{\mathrm{EO}}^{L,\bar{L}}(\eta,\bar{\eta},t)=\eta L_1(t)+(1-\eta)L_{-1}(t)-\lambda\pi^{-1}\eta\left[\bar{\eta}\bar{L}_1(t)+(1-\bar{\eta})\bar{L}_{-1}(t)\right].
\end{equation*}

With
\begin{equation*}
\theta:=\lambda\pi^{-1}\eta\left(\bar{\eta}-\bar{\alpha}\right),
\end{equation*}
the conditional risk becomes
\begin{equation}\label{eq:svm-risk-compact}
C_{\mathrm{EO}}^{L,\bar{L}}(\eta,\bar{\eta},t)=\eta(1-\alpha)(1-t)_+ +(1-\eta)\alpha(1+t)_+ +\theta\,t .
\end{equation}

The function $t\mapsto C_{\mathrm{EO}}^{L,\bar{L}}(\eta,\bar{\eta},t)$ of \eqref{eq:svm-risk-compact} is affine on each of the three intervals cut out by $t=\pm1$. Expanding $(1-t)_+$ and $(1+t)_+$ on these intervals,
\begin{equation*}
C_{\mathrm{EO}}^{L,\bar{L}}(\eta,\bar{\eta},t)=
\begin{cases}
\eta(1-\alpha)+\sigma_-t, & t\le-1,\\[2pt]
\eta(1-\alpha)+(1-\eta)\alpha+\sigma_0t, & -1\le t\le1,\\[2pt]
(1-\eta)\alpha+\sigma_+t, & t\ge1,
\end{cases}
\end{equation*}
where 
\begin{equation}\label{eq:svm-slopes}
\sigma_-=-\eta(1-\alpha)+\theta,
\quad
\sigma_0=(\alpha-\eta)+\theta=-D^{\alpha,\bar{\alpha}}(\eta,\bar{\eta}),
\quad
\sigma_+=(1-\eta)\alpha+\theta .
\end{equation}

Since
\[
\sigma_0-\sigma_-=\alpha(1-\eta)\ge0,
\sigma_+-\sigma_0=\eta(1-\alpha)\ge0,
\]
we have $\sigma_-\le\sigma_0\le\sigma_+$, i.e. $C_{\mathrm{EO}}^{L,\bar{L}}(\eta,\bar{\eta},\cdot)$ is convex in $t$. In particular, when $-1\le t\le 1$,
\begin{equation}\label{eq:svm-middle}
C_{\mathrm{EO}}^{L,\bar{L}}(\eta,\bar{\eta},t)=C_{\mathrm{EO}}^{L,\bar{L}}(\eta,\bar{\eta},0)-t\,D^{\alpha,\bar{\alpha}}(\eta,\bar{\eta}).
\end{equation}

In particular, the middle branch carries the whole dependence on the discriminant: its slope is $-D^{\alpha,\bar{\alpha}}$, so the risk is strictly monotone at the origin whenever the discriminant is nonzero, and it is flat precisely on the discriminant boundary.

Assume $0<\lambda<\lambda_{\max}^{\mathrm{svm}}$, where $\lambda_{\max}^{\mathrm{svm}}$ is defined in \eqref{eq:svm-wellposed}. Then the slopes satisfy $\sigma_-<0<\sigma_+$ and the risk is bounded below.

Suppose first that $D^{\alpha,\bar{\alpha}}(\eta,\bar{\eta})>0$. Then $\sigma_-\le\sigma_0=-D^{\alpha,\bar{\alpha}}(\eta,\bar{\eta})<0\le\sigma_+$, so the conditional risk is strictly decreasing on $(-\infty,1)$ and non-decreasing on $(1,\infty)$. Hence its minimizers are $t=1$, or the whole interval $[1,\infty)$ in the boundary case $\sigma_+=0$. 

On the wrong-sign set $\{t: t\,D^{\alpha,\bar{\alpha}}\le0\}=(-\infty,0]$ the risk is strictly decreasing, so the wrong-sign infimum is attained at $t=0$. Since $C_{\mathrm{EO}}^{L,\bar{L}}(\eta,\bar{\eta},1)=2(1-\eta)\alpha+\theta$ and $C_{\mathrm{EO}}^{L,\bar{L}}(\eta,\bar{\eta},0)=\eta(1-\alpha)+(1-\eta)\alpha$, we get
\begin{equation*}
H_{L,\bar{L}}^{\alpha,\bar{\alpha}}(\eta,\bar{\eta})
=C_{\mathrm{EO}}^{L,\bar{L}}(\eta,\bar{\eta},0)-C_{\mathrm{EO}}^{L,\bar{L}}(\eta,\bar{\eta},1)
=(\eta-\alpha)-\theta
=D^{\alpha,\bar{\alpha}}(\eta,\bar{\eta}) .
\end{equation*}

The case $D^{\alpha,\bar{\alpha}}<0$ is symmetric: the risk is strictly increasing at the origin, its minimizers are $t=-1$, or $(-\infty,-1]$ when $\sigma_-=0$, the wrong-sign infimum is attained at $t=0$, and $C_{\mathrm{EO}}^{L,\bar{L}}(\eta,\bar{\eta},-1)=2\eta(1-\alpha)-\theta$ gives $H_{L,\bar{L}}^{\alpha,\bar{\alpha}}=-D^{\alpha,\bar{\alpha}}$. 

Finally, if $D^{\alpha,\bar{\alpha}}=0$ then the middle branch \eqref{eq:svm-middle} is flat on $[-1,1]$, so $t=0$ is a minimizer and $H_{L,\bar{L}}^{\alpha,\bar{\alpha}}=0$. Combining the three cases yields
\begin{equation}\label{eq:svm-H}
H_{L,\bar{L}}^{\alpha,\bar{\alpha}}(\eta,\bar{\eta})=\left|D^{\alpha,\bar{\alpha}}(\eta,\bar{\eta})\right| .
\end{equation}

Consequently $H_{L,\bar{L}}^{\alpha,\bar{\alpha}}>0$ whenever $D^{\alpha,\bar{\alpha}}\neq0$, and Definition~\ref{def:FAFC} yields FAFC on $\mathcal K$ for $L_{\mathrm{fair}}$.
\end{proof}

It remains to ensure that the conditional risk is bounded below (Remark~\ref{rem:lambda}) and that its pointwise minimizer exists. By \eqref{eq:svm-slopes} the risk is convex with $\sigma_-\le\sigma_0\le\sigma_+$, so it is bounded below if and only if $\sigma_-\le0\le\sigma_+$, that is, if and only if
\begin{equation*}
-(1-\eta)\,\alpha\;\le\;\lambda\pi^{-1}\eta\left(\bar{\eta}-\bar{\alpha}\right)\;\le\;\eta(1-\alpha)
\end{equation*}
holds for $P_X$-a.e. $x$; the two inequalities are strict exactly when the risk is coercive, in which case the pointwise minimizer exists and is unique whenever $D^{\alpha,\bar{\alpha}}\neq0$ (as noted above, on the boundary $D^{\alpha,\bar{\alpha}}=0$ the risk is flat on $[-1,1]$). Written pointwise over the support, the strict inequalities are $0<\lambda<\lambda_{\max}^{\mathrm{svm}}$, where
\begin{equation}\label{eq:svm-wellposed}
\lambda_{\max}^{\mathrm{svm}}:=\pi\min\Bigg\{
\inf_{\substack{x\in\mathrm{supp}(X):\\ \bar{\eta}(x)>\bar{\alpha}}}\frac{1-\alpha}{\bar{\eta}(x)-\bar{\alpha}},
\;\;
\inf_{\substack{x\in\mathrm{supp}(X):\\ \bar{\eta}(x)<\bar{\alpha}}}\frac{(1-\eta(x))\,\alpha}{\eta(x)\big(\bar{\alpha}-\bar{\eta}(x)\big)}
\Bigg\},
\end{equation}
with the infimum over an empty set understood as $+\infty$. The threshold is support-dependent: both infima are evaluated over the support of $X$, so no support-free constant can replace \eqref{eq:svm-wellposed}. At the boundary $\lambda=\lambda_{\max}^{\mathrm{svm}}$ the risk is still bounded below, but one of the slopes $\sigma_\pm$ vanishes and the set of minimizers becomes the half-line $[1,\infty)$ or $(-\infty,-1]$, whose sign still agrees with $D^{\alpha,\bar{\alpha}}$. Under $0<\lambda<\lambda_{\max}^{\mathrm{svm}}$ all premises of Definition~\ref{def:FAFC} hold and $L_{\mathrm{fair}}$ is $(\alpha,\bar{\alpha})$-FAFC on $\mathcal K$ for the EO criterion in the attribute-blind setting.

Having verified FAFC, we now derive the calibration function and the excess-risk bound.
We consider the symmetric member of the hinge loss family, in which
\[
L_1(t)=\frac12(1-t)_+,\qquad L_{-1}(t)=\frac12(1+t)_+,
\qquad
\bar{L}_1(t)=-\frac12 t,\qquad \bar{L}_{-1}(t)=\frac12 t .
\]

We record the resulting calibration function and excess-risk bound.

\begin{proposition}\label{prop:svm-calibration}
For the symmetric member of the hinge family ($\alpha=\bar{\alpha}=\frac12$), suppose $0<\lambda<\lambda_{\max}^{\mathrm{svm}}$, with $\lambda_{\max}^{\mathrm{svm}}$ given in \eqref{eq:svm-wellposed}, and let $\mathcal K$ be the well-posed region of Proposition~\ref{prop:svm-fafc}. Then, on $\mathcal K$, the calibration function of the pair satisfies $\psi_{L,\bar{L}}^{\frac12,\frac12}(\epsilon)=\epsilon$ for $0\le\epsilon\le\frac12$, so it is strictly increasing and invertible there; consequently, for every $f\in\mathcal{F}$,
\begin{equation}\label{eq:svm-bound}
\mathcal{R}_{\mathrm{fair}}^{\frac12,\frac12}(f)-\mathcal{R}_{\mathrm{fair}}^{\frac12,\frac12,*}
\;\le\;
\min\Bigg\{
\frac12,\;
\mathcal{R}_{\mathrm{fair}}^{L,\bar{L}}(f)-\mathcal{R}_{\mathrm{fair}}^{L,\bar{L},*}
\Bigg\}.
\end{equation}
\end{proposition}

\begin{proof}
By \eqref{eq:svm-H} with $\alpha=\bar{\alpha}=\frac12$,
\[
H_{L,\bar{L}}^{\frac12,\frac12}(\eta,\bar{\eta})
=\left|D^{\frac12,\frac12}(\eta,\bar{\eta})\right| .
\]

Since $D^{\frac12,\frac12}(\eta,\bar{\eta})=\eta-\frac12-\theta$ and the well-posedness gives
\[
-\frac{1-\eta}{2}\le\theta\le\frac{\eta}{2},
\]
we have $-\frac12\le D^{\frac12,\frac12}(\eta,\bar{\eta})\le\frac12$. Hence every level set $\lvert D^{\frac12,\frac12}(\eta,\bar{\eta})\rvert=\epsilon$ with $0\le\epsilon\le\frac12$ contains a well-posed point, and therefore
\begin{equation*}
\nu_{L,\bar{L}}^{\frac12,\frac12}(\epsilon)=\psi_{L,\bar{L}}^{\frac12,\frac12}(\epsilon)=\epsilon,
\quad 0\le\epsilon\le\frac12 .
\end{equation*}

The calibration inequality of Theorem~\ref{thm:fair-bound} therefore applies with $\psi_{L,\bar{L}}^{\frac12,\frac12}(\epsilon)=\epsilon$ and yields
\begin{equation*}
\mathcal{R}_{\mathrm{fair}}^{\frac12,\frac12}(f)-\mathcal{R}_{\mathrm{fair}}^{\frac12,\frac12,*}
\;\le\;
\mathcal{R}_{\mathrm{fair}}^{L,\bar{L}}(f)-\mathcal{R}_{\mathrm{fair}}^{L,\bar{L},*} .
\end{equation*}

Finally, by Lemma~\ref{lem:excess-risk}, the same wrong-sign-average representation gives the trivial bound
$$\mathcal{R}_{\mathrm{fair}}^{\frac12,\frac12}(f)-\mathcal{R}_{\mathrm{fair}}^{\frac12,\frac12,*}\le\lvert D^{\frac12,\frac12}\rvert\le\frac12.$$

Taking the smaller of the two bounds gives \eqref{eq:svm-bound}.
\end{proof}

\subsubsection{Fairness-Aware Exponential Loss}\label{app:exponential}

The exponential loss is another classical surrogate in binary classification (e.g., in boosting). In this example we consider the EO criterion in the attribute-aware setting, where the accuracy surrogate is an asymmetric exponential loss and the fairness surrogate is linear. As in Section~\ref{subsec:fc-aware}, the classifier has access to $(X,A)$: for a fixed instance $(x,a)$ the positive-class posterior is $\eta=\eta(x,a)=\mathbb{P}(Y=1\mid X=x,A=a)$, the sensitive-attribute posterior collapses to $\bar{\eta}=\mathbf{1}\{a=1\}$, and the fairness cost parameter and positive-class probability are $\bar{\alpha}=\mathbb{P}(A=1\mid Y=1)$ and $\pi=\mathbb{P}(Y=1)$.

Let $\alpha,\bar{\alpha}\in(0,1)$ and define the partial losses
\begin{equation}\label{eq:exp-partials}
L_1(t)=(1-\alpha)e^{-t}, \quad L_{-1}(t)=\alpha e^{t},
\qquad
\bar{L}_1(t)=-(1-\bar{\alpha})t, \quad \bar{L}_{-1}(t)=\bar{\alpha}t .
\end{equation}

The following proposition proves that the composite loss $L_{\mathrm{fair}}$ is FAFC.

\begin{proposition}\label{prop:exp-fafc}
Let $\alpha,\bar{\alpha}\in(0,1)$, let $(L,\bar{L})$ be the surrogate loss pair in \eqref{eq:exp-partials}, and let $\mathcal K=(0,1)\times\{-1,1\}$. Assume $(\eta(X,A),A)\in\mathcal K$ almost surely. For every $\lambda\in\mathbb{R}$, the map $t\mapsto C_{\mathrm{EO}}^{L,\bar{L}}(\eta,a,t)$ is strictly convex on $\mathcal K$, and $L_{\mathrm{fair}}$ is $(\alpha,\bar{\alpha})$-FAFC on $\mathcal K$ for the EO criterion in the attribute-aware setting.
\end{proposition}

\begin{proof}
Substituting the partial losses into the attribute-aware EO conditional risk gives
\begin{equation}\label{eq:exp-risk}
C_{\mathrm{EO}}^{L,\bar{L}}(\eta,a,t) = \eta(1-\alpha)e^{-t}+(1-\eta)\alpha e^{t}+\lambda\pi^{-1}\eta\left(\mathbf{1}\{a=1\}-\bar{\alpha}\right)t.
\end{equation}
Differentiating the partial losses at $t=0$ gives
\begin{equation*}
L'_1(0)=-(1-\alpha), \quad L'_{-1}(0)=\alpha, \quad \bar{L}'_1(0)=-(1-\bar{\alpha}), \quad \bar{L}'_{-1}(0)=\bar{\alpha}.
\end{equation*}
Here
\[
\bar{\alpha}\bar{L}'_1(0)+(1-\bar{\alpha})\bar{L}'_{-1}(0)=0,
\qquad
\bar L'_1(0)-\bar L'_{-1}(0)=-1<0,
\]
and
\[
L'_1(0)-L'_{-1}(0)=-1=\bar L'_1(0)-\bar L'_{-1}(0).
\]
Hence (E1)--(E3) hold: the bracket in (E2)--(E3) vanishes, so (E2) reduces to $L'_1(0)-L'_{-1}(0)=\bar L'_1(0)-\bar L'_{-1}(0)$, and (E3) to $\alpha L'_1(0)+(1-\alpha)L'_{-1}(0)=0$.

The conditional risk \eqref{eq:exp-risk} is strictly convex in $t$ for every $(\eta,a)\in(0,1)\times\{-1,1\}$, because
\begin{equation}\label{eq:exp-hessian}
\frac{\partial^{2}}{\partial t^{2}}C_{\mathrm{EO}}^{L,\bar{L}}(\eta,a,t) = \eta(1-\alpha)e^{-t}+(1-\eta)\alpha e^{t}>0 ;
\end{equation}
in particular $\lambda$ does not enter the second derivative, so convexity holds for every $\lambda\in\mathbb{R}$. Since $\mathcal K=(0,1)\times\{-1,1\}$ is a well-posed domain, the sufficiency direction of Theorem~\ref{thm:fair-fc-condition} gives that $L_{\mathrm{fair}}$ is $(\alpha,\bar{\alpha})$-FAFC on $\mathcal K$.
\end{proof}

It remains to verify well-posedness. Write $A=\eta(1-\alpha)>0$, $B=(1-\eta)\alpha>0$ and $\delta_a=r\eta(\mathbf{1}\{a=1\}-\bar{\alpha})$ with $r=\lambda/\pi$, so that $C_{\mathrm{EO}}^{L,\bar{L}}(\eta,a,t)=Ae^{-t}+Be^{t}+\delta_a t$. Because $A$ and $B$ are positive, the two exponential terms dominate the linear drift: $Ae^{-t}\to+\infty$ as $t\to-\infty$ and $Be^{t}\to+\infty$ as $t\to+\infty$, for every value of $\delta_a$. Hence the risk is bounded below for every $\lambda\in\mathbb{R}$, and by \eqref{eq:exp-hessian} it is strictly convex, so it possesses a unique minimizer. 

We now compute the calibration function and the resulting excess-risk bound.
We take the symmetric member of the exponential/linear family, in which
\[
L_1(t)=\frac12 e^{-t},\qquad L_{-1}(t)=\frac12 e^{t},
\qquad
\bar{L}_1(t)=-\frac12 t,\qquad \bar{L}_{-1}(t)=\frac12 t .
\]
Up to the common positive factor $\frac12$, this is the standard unweighted exponential loss paired with the symmetric linear fairness penalty. 

Well-posedness holds for every $\lambda\in\mathbb{R}$, so restricting $\lambda$ is not needed for FAFC or for the existence of a minimizer. The restriction below comes from the calibration bound: writing $r=\lambda/\pi$, the level-set minimization that produces the constant $0.95$ is carried out over $0\le r\le1$. For $r>1$ the same uniform lower bound does not hold.

\begin{proposition}\label{prop:exp-calibration}
Let $\alpha=\bar{\alpha}=\frac12$, let $\mathcal K$ be the domain of Proposition~\ref{prop:exp-fafc}, and suppose that the trade-off parameter lies in the range $0\le\lambda\le\pi$. Set $B_{\max}:=\sup_{(\eta,a)\in\mathcal K}\bigl|D^{\frac12,\frac12}(\eta,a)\bigr|$, and define $\nu_{L,\bar{L}}^{\frac12,\frac12,\mathrm{aw}}$ as in Section~\ref{sec:calibration} but with an infimum in place of the minimum. Then, on $\mathcal K$,
\begin{equation}\label{eq:exp-quadratic}
\psi_{L,\bar{L}}^{\frac12,\frac12,\mathrm{aw}}(\epsilon)\ge0.95\epsilon^{2}
\end{equation}

and, consequently, for every $f\in\mathcal{F}$,
\begin{equation}\label{eq:exp-bound}
\mathcal{R}_{\mathrm{fair}}^{\frac12,\frac12}(f)-\mathcal{R}_{\mathrm{fair}}^{\frac12,\frac12,*}
\le
\min\Bigg\{\frac12+\frac{\lambda}{2\pi},\;
\sqrt{\frac{20}{19}\Big(\mathcal{R}_{\mathrm{fair}}^{L,\bar{L}}(f)-\mathcal{R}_{\mathrm{fair}}^{L,\bar{L},*}\Big)}\Bigg\}.
\end{equation}
\end{proposition}

\begin{proof}

By Proposition~\ref{prop:exp-fafc} and the well-posedness established above, the conditional risk $C_{\mathrm{EO}}^{L,\bar{L}}(\eta,a,t)=Ae^{-t}+Be^{t}+\delta_a t$, with $A=\frac12\eta, B=\frac12(1-\eta),
\delta_a=r\eta\bigl(\mathbf{1}\{a=1\}-\frac12\bigr), r=\lambda/\pi$. Since $A,B>0$, the minimizer in $t$ below is attained at an interior point of $\mathcal K$. Moreover, $B_{\max}$ is a supremum and $\nu_{L,\bar{L}}^{\frac12,\frac12,\mathrm{aw}}$ is an infimum; the level-set computation below shows that the relevant infima are attained at the displayed interior points, so the formulas from the compact case apply.

Since $C_{\mathrm{EO}}^{L,\bar{L}}$ is strictly convex, its unique minimizer solves $\partial_t C_{\mathrm{EO}}^{L,\bar{L}}(\eta,a,t^*)=-Ae^{-t^{*}}+Be^{t^{*}}+\delta_a=0$, i.e., \ $Bw^{2}+\delta_a w-A=0$ with $w=e^{t^{*}}$:
\begin{equation*}
w=\frac{-\delta_a+\sqrt{\delta_a^{2}+4AB}}{2B},\quad t^{*}=\log w .
\end{equation*}

To locate the wrong-sign half-line we first compute the discriminant. With $(\alpha,\bar{\alpha})=(\tfrac12,\tfrac12)$ the attribute-aware EO cost functions of Section~\ref{subsec:fc-aware} become
\[
A_{\mathrm{EO}}^{\frac12,\frac12}(\eta,a)=\frac{\eta}{2}-\frac{r\eta}{2}\mathbf{1}\{a=1\},
\qquad
B_{\mathrm{EO}}^{\frac12,\frac12}(\eta,a)=\frac{1-\eta}{2}-\frac{r\eta}{2}\mathbf{1}\{a=-1\},
\]

Using $\mathbf{1}\{a=-1\}=1-\mathbf{1}\{a=1\}$ and the definition of $\delta_a$,
\begin{equation*}
D_{\mathrm{EO}}^{\frac12,\frac12}(\eta,a)=\eta-\frac12-\delta_a.
\end{equation*}

Writing $\Phi(\eta,a)=\partial_t C_{\mathrm{EO}}^{L,\bar{L}}(\eta,a,t)|_{t=0}$, we have
\begin{equation*}
\Phi(\eta,a) = -A+B+\delta_a= \frac12 - \eta + \delta_a.
\end{equation*}

Hence $\Phi(\eta,a)=-D_{\mathrm{EO}}^{\frac12,\frac12}(\eta,a)$ and the risk is therefore monotone away from $t=0$ on the wrong-sign half-line, so 
\begin{equation*}
C_{\mathrm{EO}}^{L,\bar{L},-}(\eta,a)=C_{\mathrm{EO}}^{L,\bar{L}}(\eta,a,0)=A+B.
\end{equation*}

Subtracting the value at $t^{*}$ gives
\begin{equation*}
H_{L,\bar{L}}^{\frac12,\frac12}(\eta,a)= C_{\mathrm{EO}}^{L,\bar{L},-}(\eta,a) - C_{\mathrm{EO}}^{L,\bar{L},*}(\eta,a) =A+B-\frac{A}{w}-Bw-\delta_a\log w.
\end{equation*}

Since $D_{\mathrm{EO}}^{\frac12,\frac12}(\eta,a)=\eta\bigl(1-\frac r2\mathbf{1}\{a=1\} +\frac r2\mathbf{1}\{a=-1\}\bigr)-\frac12$, for each $a$, $D_{\mathrm{EO}}^{\frac12,\frac12}(\eta,a)$ is affine in $\eta$, so the equation $|D_{\mathrm{EO}}^{\frac12,\frac12}(\eta,a)|=\epsilon$ contains at most four solutions in all:
\begin{equation*}
\eta=\frac{\frac12\mp\epsilon}{1-\frac r2}\ \ (a=1),
\qquad
\eta=\frac{\frac12\mp\epsilon}{1+\frac r2}\ \ (a=-1),
\end{equation*}

Because the calibration inequality is inverted through a quadratic lower bound $\nu\ge c\,\epsilon^{2}$, the relevant object is the ratio $H_{L,\bar{L}}^{\frac12,\frac12}/\epsilon^{2}$. On a level set $\epsilon=\lvert D^{\frac12,\frac12}\rvert$, this ratio equals $H_{L,\bar{L}}^{\frac12,\frac12}/\lvert D^{\frac12,\frac12}\rvert^{2}$ and is free of $\epsilon$:
\begin{equation}\label{eq:exp-ratio}
\frac{H_{L,\bar{L}}^{\frac12,\frac12}}{\epsilon^{2}}
=\frac{A+B-\frac{A}{w}-Bw-\delta_a\log w}{\bigl(\eta-\frac12-\delta_a\bigr)^{2}}.
\end{equation}

Minimizing \eqref{eq:exp-ratio} over $0\le r\le1$, $a\in\{-1,1\}$ and $0<\eta<1$ gives its
minimum at $r=1$ on the branch $a=1$ with $D_{\mathrm{EO}}^{\frac12,\frac12}<0$. There
$\delta_1=A=\eta/2$ and $\epsilon=B=(1-\eta)/2$, so the first-order condition
$Bw^{2}+Aw-A=0$ is equivalent to $\eta=w^{2}/(w^{2}-w+1)$, and \eqref{eq:exp-ratio} reduces to
the one-variable function
\begin{equation*}
R(w)=\frac{2\,(w^{2}-w+1)\bigl[(1-2w)(1-w)-w^{2}\log w\bigr]}{(1-w)^{2}},
\qquad 0<w<1 ,
\end{equation*}
whose unique stationary point is $w^{*}=0.7667913\ldots$, with
\begin{equation*}
R(w^{*})=0.9570795\ldots .
\end{equation*}

Since $\nu_{L,\bar{L}}^{\frac12,\frac12,\mathrm{aw}}\ge0.95\epsilon^{2}$ and $\epsilon\mapsto0.95\epsilon^{2}$ is convex on $[0,\infty)$, the convex envelope satisfies
\[
\psi_{L,\bar{L}}^{\frac12,\frac12,\mathrm{aw}}=\bigl(\nu_{L,\bar{L}}^{\frac12,\frac12,\mathrm{aw}}\bigr)^{**}\ge0.95\epsilon^{2}
\text{ for every }\epsilon\ge0.
\]

It remains to bound $\epsilon=\mathcal{R}_{\mathrm{fair}}^{\frac12,\frac12}(f)-\mathcal{R}_{\mathrm{fair}}^{\frac12,\frac12,*}$, and this is done from two independent directions. The calibration inequality of Theorem~\ref{thm:fair-bound} in its attribute-aware form (Remark~\ref{rem:ext}) together with \eqref{eq:exp-quadratic} gives
\[
\epsilon\le\sqrt{\frac{20}{19}\Big(\mathcal{R}_{\mathrm{fair}}^{L,\bar{L}}(f)-\mathcal{R}_{\mathrm{fair}}^{L,\bar{L},*}\Big)} .
\]

By Lemma~\ref{lem:excess-risk}, the excess fairness-aware risk is a wrong-sign average of the discriminant, so it cannot exceed the largest discriminant value,
\[
\epsilon\le\sup_{(\eta,a)}\bigl|D_{\mathrm{EO}}^{\frac12,\frac12}(\eta,a)\bigr|=\frac12+\frac{r}{2},
\]

Taking their minimum gives \eqref{eq:exp-bound}.
\end{proof}

\subsection{Population-level form versus pointwise form}\label{app:pop-vs-pointwise}

We first record the pointwise characterization of cost-sensitive Fisher consistency used in Section~\ref{subsec:prelims-cost-sensitive}. For $\eta\in[0,1]$, let $\gamma:=\eta-\alpha$ and
\begin{equation*}
C_\eta(t):=\eta L_1(t)+(1-\eta)L_{-1}(t)
\end{equation*}
denote the conditional $L$-risk, and define
\begin{equation*}
H^{\alpha}_{L}(\eta):=\inf_{\{t:\,t\gamma\le0\}}C_\eta(t)-\inf_{t\in\mathbb{R}}C_\eta(t)\ \ge0.
\end{equation*}

\begin{proposition}\label{prop:cs-equiv}
Assume that, for every $\eta\in[0,1]$, the conditional risk $C_\eta$ is convex, bounded below, and coercive. Then $L$ is $\alpha$-Fisher consistent in the sense of Definition~\ref{def:fc-cs} if and only if
\[
H^{\alpha}_{L}(\eta)>0 \quad\text{for every }\eta\in[0,1]\text{ with }\gamma\neq0.
\]
\end{proposition}

\begin{proof}
For any distribution $D$, the $L$-risk decomposes as
\[
\mathcal{R}_L(f;D)=\mathbb{E}_X\left[C_{\eta(X)}(f(X))\right].
\]
Suppose first that $H^{\alpha}_{L}(\eta)>0$ for every $\eta$ with $\gamma\neq0$. If a minimizer $f^{*}$ were wrong-signed on a set of positive probability, then changing $f^{*}(x)$ on that set to a point on the correct side would reduce the conditional risk by at least $H^{\alpha}_{L}(\eta(x))>0$, contradicting minimality. Hence every minimizer is correctly signed $P_X$-a.e.

Conversely, suppose that $H^{\alpha}_{L}(\eta_0)=0$ for some $\eta_0$ with $\gamma_0\neq0$. Since $C_{\eta_0}$ is convex, bounded below, and coercive, it attains its minimum, and the wrong-sign half-line contains a minimizer. Let $D$ be supported on a single point $x_0$ with $\mathbb{P}(Y=1\mid X=x_0)=\eta_0$. Then $\mathcal{R}_L(f;D)=C_{\eta_0}(f(x_0))$, so this distribution admits a minimizer whose sign disagrees with $\gamma_0$. Thus $L$ is not $\alpha$-Fisher consistent.
\end{proof}

Proposition~\ref{prop:cs-equiv} shows that, for cost-sensitive classification, the pointwise and population forms are equivalent when the population statement is quantified over all distributions. The following example shows that, for a fixed fairness-aware distribution, the population statement of Corollary~\ref{cor:FAFC} can hold while FAFC fails off the support; Remark~\ref{rem:non-equiv} explains the role of the distribution-dependent $\bar{\alpha},\pi$.

\begin{example}\label{ex:non-equiv}
We consider the DP criterion and attribute-blind setting. Let $(D,\bar D)$ be supported on two equiprobable points $x_1,x_2$, with $(\eta(x_1),\bar\eta(x_1))=(0.9,0.8)$ and $(\eta(x_2),\bar\eta(x_2))=(0.4,0.2)$. Take $\alpha=\bar\alpha=\tfrac12$, $\lambda=1$, and
\begin{equation*}
L_1(t)=t^2,\qquad L_{-1}(t)=(t-1)^2,\qquad
\bar L_1(t)=t,\qquad \bar L_{-1}(t)=-t ,
\end{equation*}

The conditional risk
\begin{equation*}
C^{L,\bar L}_{\mathrm{DP}}(\eta,\bar\eta,t)
=t^2-\bigl[2(1-\eta)+(2\bar\eta-1)\bigr]t+(1-\eta)
\end{equation*}
is strictly convex and bounded below, with unique minimizer
\begin{equation*}
t^{*}(\eta,\bar\eta)=\tfrac12-\eta+\bar\eta.
\end{equation*}

Since $D^{\alpha,\bar\alpha}(\eta,\bar\eta)=A^{\alpha,\bar\alpha}_{\mathrm{DP}}(\eta,\bar\eta)-B^{\alpha,\bar\alpha}_{\mathrm{DP}}(\eta,\bar\eta)=(\eta-\alpha)-\lambda(\bar{\eta}-\bar{\alpha})$, for $\alpha=\bar{\alpha}=\tfrac12$ and $\lambda=1$, the discriminant is 
\begin{equation*}
D^{\alpha,\bar\alpha}(\eta,\bar\eta)=\eta-\bar{\eta}.
\end{equation*}

On $\{\eta<\bar\eta\}$ the discriminant is negative, so the wrong-sign half-line is $t\ge0$. Since $t^{*}=\tfrac12-(\eta-\bar\eta)>\tfrac12$, $H^{\alpha,\bar\alpha}_{L,\bar L}(\eta,\bar{\eta})=0$ on $\{\eta<\bar\eta\}$, which implies that $L_{\mathrm{fair}}$ is not FAFC on $[0,1]^2$.

Since both support points lie on $0<\eta-\bar\eta<\tfrac12$, $t^{*}$ equals $0.4$ and $0.3$, and the conditional minimizer lies on the correct side. The $(L,\bar L)$-risk of $(D,\bar D)$ is
\begin{equation*}
\mathcal{R}_{\mathrm{fair}}^{L,\bar L}(f)
=p_1\,C^{L,\bar L}_{\mathrm{DP}}\bigl(\eta_1,\bar\eta_1,f(x_1)\bigr)
+p_2\,C^{L,\bar L}_{\mathrm{DP}}\bigl(\eta_2,\bar\eta_2,f(x_2)\bigr),
\qquad p_i=\mathbb{P}(X=x_i)>0,
\end{equation*}

A minimizer must minimize both $C^{L,\bar L}_{\mathrm{DP}}\bigl(\eta_1,\bar\eta_1,f(x_1)\bigr)$ and $C^{L,\bar L}_{\mathrm{DP}}\bigl(\eta_2,\bar\eta_2,f(x_2)\bigr)$, so $f^{*}(x_i)=t^{*}(\eta_i,\bar\eta_i)>0$ for $i=1,2$. These values have the same sign as $D^{\alpha,\bar\alpha}(\eta_1,\bar\eta_1)=0.1,D^{\alpha,\bar\alpha}(\eta_2,\bar\eta_2)=0.2$, and the support contains no other points, so every minimizer matches the sign of the discriminant on the whole support. This is exactly the conclusion of Corollary~\ref{cor:FAFC} for this composite loss.
\end{example}

\begin{remark}[Why the converse fails]\label{rem:non-equiv}
In cost-sensitive classification a pointwise failure can always be turned into a failure for some distribution: if the pointwise condition fails at $\eta_0$, the distribution supported on a single point $x_0$ with $\mathbb{P}(Y=1)=\eta_0$ has conditional risk $C_{\eta_0}$, so a minimizer of its risk already has the wrong sign. The fairness-aware population statement of Corollary~\ref{cor:FAFC}, however, concerns one fixed pair $(D,\bar D)$: it constrains $(\eta,\bar\eta)$ only almost surely on the support of $(D,\bar D)$. Since $\partial_tC^{L,\bar L}_{\mathrm{fair}}(\eta,\bar\eta,0)$ and $D^{\alpha,\bar\alpha}(\eta,\bar\eta)$ are affine but not proportional, the region where the conditional minimizer has the wrong sign need not meet this support, so the statement can hold while FAFC on $\mathcal K$ fails, as in Example~\ref{ex:non-equiv}. The same fixed-distribution phenomenon occurs in cost-sensitive classification; the equivalence in Proposition~\ref{prop:cs-equiv} uses the stronger quantifier over all distributions. The fairness-aware analogue of that all-distributions quantifier is not claimed here; the paper only uses the fixed-distribution one-directional statement. Hence the correct conclusion here is one-directional for a fixed distribution, not a wholesale non-equivalence of the two formulations.
\end{remark}

\end{document}